\documentclass[reqno]{amsart}
\usepackage{amsmath}
\usepackage{amsopn}
\usepackage{amssymb}
\usepackage{amsfonts}
\usepackage{graphicx}
\usepackage{boldline}
\usepackage{graphicx,booktabs}
\usepackage[table]{xcolor}
\usepackage{subfigure}
\usepackage{fancyhdr}
\usepackage[section]{placeins}
\usepackage[autostyle, english = american]{csquotes}
\usepackage{multicol}
\usepackage{booktabs,dcolumn,caption}
\usepackage{url}
\usepackage[bookmarks=false,colorlinks,allcolors=blue]{hyperref}
\numberwithin{table}{section}
\numberwithin{figure}{section}
\usepackage{subfigure}
\usepackage{makecell}
\usepackage{lscape} 
\usepackage{multirow}
\usepackage{longtable}
\usepackage{rotating}
\usepackage{slashbox}
\usepackage{algorithm}
\usepackage[noend]{algpseudocode}
\usepackage{hhline}

\algrenewcommand\algorithmicrequire{\textbf{Input:}} 
\algrenewcommand\algorithmicensure{\textbf{Output:}}

\newcommand{\tn}{\tabularnewline}

\newcommand{\rc}{\centering}
\newcolumntype{R}[1]{>{\raggedleft\let\newline\\\arraybackslash\hspace{0pt}}m{#1}}
\newcolumntype{L}[1]{>{\raggedright\let\newline\\\arraybackslash\hspace{0pt}}m{#1}}
\newcolumntype{C}[1]{>{\centering\let\newline\\\arraybackslash\hspace{0pt}}m{#1}}

\usepackage[foot]{amsaddr}

\definecolor{maroon}{cmyk}{0,0.87,0.68,0.32}

\begin{document}

\title{\textbf{Principal Components of the Meaning} }
	
\author{Neslihan Suzen}
\author{Alexander Gorban}
\author{Jeremy Levesley}
\author{Evgeny Mirkes}

\address{$^{1}$School of Mathematics and Actuarial Science, University of Leicester, Leicester LE1 7RH, UK}
\address{$^{2}$Lobachevsky University, Nizhni Novgorod, Russia}

\email{ns433@leicester.ac.uk (N. Suzen)}
\email{ag153@leicester.ac.uk (A.N. Gorban)}
\email{jl1@leicester.ac.uk (J. Levesley)}
\email{em322@leicester.ac.uk (E.M. Mirkes)}

\maketitle

\begin{abstract}
In this paper we argue that (lexical) meaning in science can be represented in a 13 dimension Meaning Space. This space is constructed using principal component analysis (singular decomposition) on the matrix of word-category relative information gains, where the categories are those used by the Web of Science, and the words are taken from a reduced word set from texts in the Web of Science. We show that this reduced word set plausibly represents all texts in the corpus, so that the principal component analysis has some objective meaning with respect to the corpus. We argue that 13 dimensions is adequate to describe the meaning of scientific texts, and hypothesise about the qualitative meaning of the principal components.

\vspace{5mm}


\end{abstract}

\tableofcontents

\section{Introduction}\label{intro}
The purpose of this paper is to extract the meaning (lexical meaning) of scientific texts by careful analysis of how the words used in documents specify the category the document belongs to \cite{our2}. The scientific category that a text belongs to is used to evaluate the meaning of the text. Stated differently, the research areas behind the text are characterised by these scientific categories. In this holistic approach, the meaning of a word is defined as the Relative Information Gain (RIG) for subject categories that the text belongs to, and measures the extent to which the word is informative in placing the text in the given category. Therefore, each word is represented by an $n$-dimensional vector of RIGs, where $n$ is the number of categories defined for the corpus that is being used. The \textit{Meaning Space} (MS) is a vector space, in which coordinates correspond to the subject categories and the meaning of a word is represented by a vector of RIGs related to the categories.   

In \cite{our2}, the meaning space is created for the Leicester Scientific Corpus (LSC) with 1,673,350 texts and the Leicester Scientific Dictionary-Core (LScDC) of 103,998 words with 252 Web of Science (WoS) categories \cite{our1,LSCn,LScDCn}. Each word in the LScDC is represented by 252-dimensional vector of RIGs. We have constructed the Word-Category RIG Matrix, where each entry corresponds to a pair (word,category) and its value shows the RIG for a text to belong to a category by observing this word in this text \cite{wordcat}. Row vectors of the matrix indicate the words' meaning in the scientific texts. A thesaurus of science was created by selecting the most informative words from the LScDC. The informativeness here was measured by the sum of RIGs in categories. We introduced the Leicester Scientific Thesaurus (LScT) where the most informative 5,000 words from the LScDC were included in \cite{wordcat}. Later we will use the LScT in the study of the representation of the meaning of texts.    

The characterisation of word meaning in a metric space was the first stage in the quantification of meaning in texts. We have built a metric system to allocate meaning to words based on their importance in scientific categories, i.e, to represent each word as a vector in the MS. 

Given 252 subject categories, it is unreasonable to expect that every one of these categories is uncorrelated to all others (or distinct from each other). For instance we might expect the categories {\sl Literature} and {\sl Literary Theory \& Criticism} to represent words in a very similar way in the meaning space. Indeed, subcategories are likely to occur in the data and they are expected to have close values of RIGs for words. Such attributes will measure related information, and so the original 252 dimensional data contain measurements for redundant categories. Although the MS underlying the representation of word meaning has 252 dimensions, we expect that we will be able to represent words with significantly fewer dimensions of the meaning space. 

An efficient way to represent words would be to map vectors onto a space that is constructed based on combination of original features. Mathematically speaking, we look at a linear transformation from the original set of categories to a new space composed by new components. These new components are called \textit{Components of the Meaning}. Two precise questions to be asked are: how many components of meaning are there and how are these components constructed? Thus, analysis of components (new attributes) based on the original attributes is crucial in understanding the MS. For instance, it is very important to understand which categories contribute the most and which the least to the new dimensions. Also, it is instructive to see if the new dimensions have some real semantic meaning, for instance, in distinguishing between natural and social sciences or experimental and theoretical research.      

We narrowed the Word-Category RIG Matrix to words from the LScT; therefore, there are 5,000 rows and 252 columns in the matrix. Here we hypothesize that 252 is not the actual dimension of the meaning space. In order to identify the most significant dimensions and construct a new space, we perform Principal Component Analysis (PCA) \cite{wold}. PCA ensures that words which are represented by similar sets of categories will be nearby points in the lower dimensional space. Through PCA on vectors of RIGs, we map each vector onto a vector with a reduced number of entries. 

The PCA of word vectors provides a series of 252 principal components (PC), called as \textit{Principal Components of the Meaning}. 
Of the 252 PCs produced by PCA, only the first $m$ will resemble the true underlying MS, while the remainder will be mainly represent noise in the data. The most significant $m$ principal components are used to construct new orthogonal axes that span an $m$-dimensional vector space. Using these $m$ components, every word with 252 dimensions can be mapped onto a word vector with $m$ components. The optimal value of the $m$ can be estimated using several methods \cite{jackson,king,peres}.

It is noteworthy that this analysis can be performed under the following assumptions: (1) the thesaurus is representative of the corpus LSC; (2) each category is represented by reasonable number of words of the LScT, and (3) each word can be represented by the WoS categories. To understand the relationships between the thesaurus and the corpus, and categories and words, we evaluated the LScT in three different ways. 

Firstly, we focused on showing how well the LScT represents the texts of the LSC. If there is a text including none of words from the LScT, then this text can not be represented by the LScT. We counted the number of the LScT words in each text and came to the conclusion that all texts of the LSC contains at least 1 word of the LScT, at most 194 words, with an average of 62 words. Indeed, one may think that representing a text with only one word is not the best in quantification of text meaning. However, we will deal with this problem in later research on text representation. We now focus on showing that the LScT is a reasonable selection as a scientific thesaurus.

Secondly, we used a procedure to test the existence of the LScT words which are informative for categories. As we restrict words in the Word-Category Frequency Matrix to the LScT words, it is possible to have column vectors with 0 in all entries. This means that none of words from the LScT is present in the texts of a particular category. For such a category, the LScT would not be an appropriate set of words to represent texts of this category. We can further infer that these words are not a  representative set of words for the LSC. This is because we assume that the texts of the LSC selected from the range of 252 subject categories are representative of the population of scientific texts as a whole. Thus none of these categories can be ignored and texts in all categories have to be represented by the selected words. The analysis of each category shows that the LScT is a reasonable selection of words, as the minimum number of words that a category includes in its texts is 733. This means that each category is presented by at least around 15\% of the LScT. 

Thirdly, we examined how many categories determine a word in the LScT. We expect that some words appear in all or the majority of categories, while some words are present in only a few categories. Even if we do not take the values of RIGs into account, we can still gain an insight into different types of vocabulary: \textit{scientific content words} and \textit{generalised service words of science} \cite{our2}. One might expect that words appearing in text(s) of all categories are not topic-specific words and words appearing in text(s) of a few categories are likely to be specific for the subjects. Therefore, we looked at the distribution of the number of words that are informative for categories, and found that more than 4,500 words appear in text (s) of at least 100 categories. Approximately 200 words appear in text(s) of all categories and only 4 words appear in text(s) of only 10 categories with a minimum number of 6 categories for a word.     

Previous results confirm that the selection of words for the LScT was reasonable and further study of quantification of the meaning of text can be done with the LScT. Thus we can use the RIG matrix based on 5000 LScT words to create a meaning space based on PCA. Firstly, we applied the Double Kaiser rule to attempt to select a subset of the original attributes by ranking them in their importance determined based on some criteria (to be explained later). Using this rule, we aim only to retain attributes which explain the data in some significant way, and to discard `trivial' attributes. Having discarded some attributes we can repeat the process, and perhaps discard some more. As it turned out, all attributes (categories) were found to be informative by our selection criterion at the first iteration. Therefore, all 252 categories are retained for further use. This means that there is no trivial attribute.

We applied PCA to the data in the 252-dimensional MS and produced 252 PCs. To understand the structure of the space, we visualised the space in two ways. 

Firstly, to describe each PC in terms of categories we created charts in which category contributions to the PC (252 coordinates of PC) are shown. These charts are used to evaluate the contributions and identify which categories have the largest influence on each PC. Using this approach we can see the main attributes for a PC and the attributes that do not contribute to that PC. Categories contributing greatly (either positively or negatively) to a PC are used to interpret that dimension. By `an attribute that is contributed greatly' we mean those attributes having coefficients larger than $1/2\sqrt{252}$ in size (positive or negative). Categories with little influence get scores near zero. Such categories can be interpreted as being unrelated attributes to the PC and this information might be useful. Therefore, all attributes are meaningful in some sense and should be interpreted appropriately.

Secondly, we have projected words onto the space defined by the PCs. This allows us to see a map of how words relate to each other with respect to the principal axes. The words used for similar topics are expected to be located near each other in these plots. 

The first PC explains 12.58\% of all variation in the data. Coordinates of the vector of the first PC appear  to measure the extent to which a category is well-defined by the words in texts in the category. Categories with more precise language have relatively higher entries in this PC, and categories with more nebulous language score lower. We did not observe any explicit distinguishing between the branches of categories. Every word is represented by  the vector of  RIGs. Projection of this vector on the frist PC can be considered as a general {\em informational value} of the word. This informational value is the first coordinate of the word in the PC basis.

The second PC (PC2) appears to primarily distinguish between categories of two main branches of science. Categories that are related to natural science and engineering have negative associations with the PC2, while those that are related to social/human science have positive associations.
 
 The third, fourth and fifth PCs reflect some of sub-branches of science with the greatest influence on the PCs. Biological sciences, computer science and engineering related categories have a strong positive correlation with PC3, while categories that are related to psychology, medicine-health and applied physics have large negative correlations. Branches of social science, economics, managements, psychology, ethics, education and multidisciplinary social sciences, appear to have strong positive correlations PC4. Literature, medicine-health science have large negative scores. For the fifth PC, categories related to ecological, environmental sciences and geosciences have large correlation. 

Next, the words are plotted in the PCA space. PCA scores for words are calculated to determine their location on each PC and the data are shown on plots on planes defined by the first three PCs (PC1-PC2, PC1-PC3, PC2-PC3) and also the volume spanned by the first three PCs together (PC1-PC2-PC3). Words that are represented by similar sets of categories will be located as nearby points in the space. In other words, PCA ensures that words that are close together in PCA space have similarity of meaning. From these plots, we have concluded that some more closely grouped words are similar in that they are used in related academic disciplines. For instance, the words `argu' and `polit' are close in the PC1-PC2 plot and they both are expected to be predominantly used in texts assigned to social science categories. Similarly, the words `health', `care', and `particip' are close together in the PC1-PC3 plot, and they are all likely to be used in medicine related texts. Thus, we conclude that in the PC space we are able to cluster words based on their meaning. However, more meaningful clusters can be obtained by using more PCs, as this will allow more separation as of groups of other meaning. With the dimension of the PC space appropriately selected, we expect that vectors of words with similar meanings will form clusters.    

Finally, we pick up the first $m$ PCs, which are in descending order of amount of information content in the PC. Three approaches to determining the appropriate number of components to retain were performed; the Kaiser rule, the Broken stick rule and an empirical method based on multicollinearity control (called PCA-CN) were applied to the data \cite{zwick,fukunaga, gorban,mirkes}. We estimated the optimal number of PCs at $m=61$, $m=16$ and $m=13$ by these methods respectively. Too large an $m$ might lead to the isolation of each word vector in the PC space, leading to the possibility of over-fitting related to the curse of dimensionality, especially for a small number of words. Therefore, we can use the 13-dimensional Meaning Space delivered by PCA-CN as the actual dimensions of meaning in future research. Full lists of categories receiving large positive and negative scores on the first 13 PCs are presented in Appendix \ref{appndx}.

This paper is organised as follows. In Section \ref{stats}, we test the LScT in a variety of ways to demonstrate that the selection of LScT words is appropriate for the representation of texts. In Section \ref{dim} we perform a principal component analysis on the meaning space. To determine the optimal number of PCs, three methods are applied and the results are discussed. The principal components of meaning are explicitly presented and analysed in some detail. Finally, in Section \ref{conc} we summarise the results of the paper, discuss their significance, and outline future research directions. 

\section{Basic Statistics in the LScT} \label{stats}
In this section we investigate the word-text and word-category relations for the LScT. We ask how many texts in the corpus LSC contain a certain number of words from the LScT (is the LScT representative of the LSC), what proportion of the thesaurus each category is present in, and how many categories each word appears in. 

\begin{figure}[tb]
  \centering
  \includegraphics[height=.35\textheight, width=0.6\textwidth]{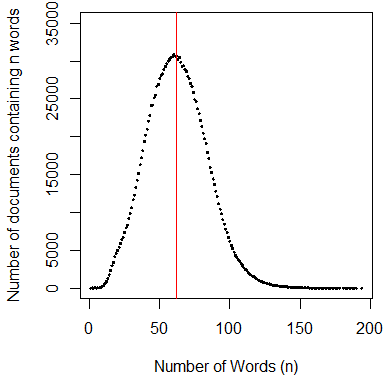}
  \caption{The number of words $(n)$ from the LScT against the number of documents from the LSC containing $n$ words. The minimum and maximum numbers of words contained in a text are 1 and 194 respectively, with an average of 62. The red line shows the average number of words.}
  \label{fig:word_doc}
\end{figure}

The LScT was created using the LSC with the aim of using a significantly reduced set of words to be representative of scientific texts in the LSC. Representativeness here refers to the extent to which words of the LScT are included in texts of the corpus LSC. Our criterion is that each text from the LSC must contain at least one word from the LScT to make the contents of the thesaurus a reasonable selection. If there is a text that contains no word from the thesaurus, the LScT is not representative for this text. Figure \ref{fig:word_doc} shows the number of texts containing a specified number of words from the LScT. All texts of the LSC include at least 1 word from the LScT, and no text contains more than 194 words, with an average of 62 words. The numbers of texts containing a small number of words are located near to 0 on the $x$-axis and the graph indicates that very few texts contain a small number of words from the LScT. 

In \cite{our2}, we discussed the relationship between the 252 Web of Science categories and words from the LScDC, and the subsequent word-category relative information gain (RIG) matrix \cite{wordcat}. For each of the  252 categories, the most informative 100 words from the LScDC were listed in descending order of their RIGs. The lists of the most informative words in each category were given in \cite{our2,wordclouds}. Clearly, the LScT is a sub-list of the LScDC; however, the two lists the most informative words from the LScDC and LScT from a category may differ from each. Table \ref{table:matex} gives such an example for the category `Mathematics'. We see that some very topic specific words (e.g. isomorph, cohomolog) in the LScDC are not present in the LScT. This is not surprising, as the LScT is representative of the average meaning of words (lexical meaning) in categories. More specific words are not needed if all we wish to do is distinguish categories. For each category, the list of the most informative 100 words from the LScT are given in Appendix \ref{InfCat}.

\begin{table}[bth]
	\centering
	\caption{Uncommon words from the most informative 100 words from the LScT and LScDC for the category `Mathematics'}
	\renewcommand\arraystretch{1.3}
\begin{tabular}{|c|c|}
\hline
\textbf{Extra Words in the LScT} &\textbf{Extra Words in the LScDC} \\\hline
affect & isomorph    \\\hline
higher & denot    \\\hline
cause & cohomolog    \\\hline
research & automorph    \\\hline
examine & semigroup    \\\hline
protein & abelian    \\\hline

\end{tabular}
\label{table:matex}
   \end{table}

For the word-category relation task, our first focus is to see how well each category is represented by the LScT. Table \ref{catwordtab} gives the number of words from the LScT for each category. The number given is the number of words appearing in the texts assigned to the given category. To gain a better insight, we provide Table \ref{table:catword1} where we list the number of categories containing numbers words from the LScT in a particular range. As we see from the table, 219 categories contains at least 3,000 words of the LScT in their texts. The maximum and the minimum numbers of words that a category includes in its texts are 4,956 and 733 respectively. The top and the bottom five categories, when categories are sorted in descending order by the number of words, are presented in Table  \ref{table:catword2}. The bottom five categories are also the five categories with the fewest number of texts. The top five categories in the Table \ref{table:catword2} appear in the list of top 30 categories containing the most texts \cite{our2}. 

\begin{table}[bth]
	\centering
	\caption{Intervals of word numbers and the number of categories for each interval. The number of words for each category is calculated by counting the number of words appearing in the texts of the category at least once.}
	\renewcommand\arraystretch{1.3}
\begin{tabular}{| L{3cm}|R{4cm}|}
\hline
\rc \textbf{Number of Words $(n)$ from the LScT} &\rc \textbf{Number of Categories Containing $n$ Words in Their Texts} \tn \hline
$4,000<n \leq 5,000$& 113    \\\hline
$3,000<n \leq 4,000$ & 106    \\\hline
$2,000<n \leq 3,000$ & 24    \\\hline
$1,000<n \leq 2,000$ & 6   \\\hline
$n \leq 1,000$ & 3    \\\hline

\end{tabular}
\label{table:catword1}
   \end{table}

\begin{table}[bth]
	\centering
	\caption{The top and the bottom five categories where categories are sorted in descending order by the number of words}
	\renewcommand\arraystretch{1.3}
\begin{tabular}{| L{4cm}|R{1.8cm}| L{4cm}|R{1.8cm}|}
\hline
\rc \textbf{Top 5 categories} &\rc \textbf{Number of Words}& \rc \textbf{Bottom 5 categories} &\rc \textbf{Number of Words} \tn \hline
Multidisciplinary Sciences	&	4956	&	Literature, Slavic	&	733	\\\hline
Environmental Sciences	&	4865	&	Literary Reviews	&	953	\\\hline
Engineering, Electrical \& Electronic	&	4864	&	Poetry	&	982	\\\hline
Computer Science, Interdisciplinary Applications	&	4834	&	Literature, African, Australian, Canadian	&	1240	\\\hline
Public, Environmental \& Occupational Health	&	4805	&	Dance	&	1295	\\\hline

\end{tabular}
\label{table:catword2}
   \end{table}

Another aspect of evaluating word-category relations is to understand the number of categories with texts containing a particular word. If the number of categories is small then this indicates the word is more specific for these categories and maybe interpreted as a \textit{scientific content word}. We suggest that words appearing in all categories are not topic-specific and may be interpreted as \textit{generalised service words of science} \cite{our2}. However, further research is necessary to identify these two groups of words.

\begin{figure}[tb]
  \centering
  \includegraphics[height=.35\textheight, width=0.6\textwidth]{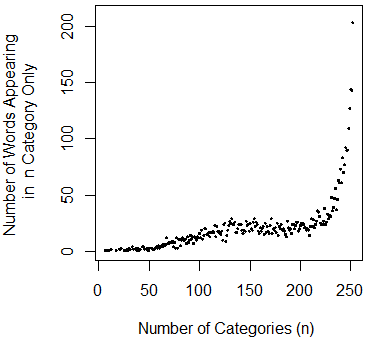}
  \caption{The number of categories $(n)$ against the number of words appearing in $n$ category. The minimum and maximum numbers of categories where a word appears in are 6 and 252.}
  \label{fig:cat_word}
\end{figure}

Figure \ref{fig:cat_word} displays the number of categories ($n$) for which texts in that category contains a word a certain number of times. So a value of 1 against $n=6$ indicates that there is one word that appears in only 6 categories. Hence, each word from the LScT appears in at least 6 and at most 252 categories. From the figure, we can see that there are approximately 200 words appearing in all categories. 2,570 words appear in more than 200 categories and approximately 2,000 words appear in between 100 and 200 categories. Very few words are located at the very beginning of the graph, and only 4 words appear less than 10 categories. Scores for the numbers of categories that some of the words appear in can be found in Table \ref{table:catword3}. One can see that words appearing in all categories are also the most frequent words in the LSC. Rare words have to be specific to a small number of categories.

 \begin{table}[bth]
	\centering
	\caption{Some words of the LScT with the number of categories where each word appears in}
	\renewcommand\arraystretch{1.3}
\begin{tabular}{| L{1.8cm}|R{1.8cm}| L{1.8cm}|R{1.8cm}|L{1.8cm}|R{1.8cm}|}
\hline
\rc \textbf{Word} &\rc \textbf{Category Number}& \rc \textbf{Word} &\rc \textbf{Category Number}& \rc \textbf{Word} &\rc \textbf{Category Number} \tn \hline
use	&	252	&	august	&	234	&	antebellum	&	12	\\ \hline
result	&	252	&	photographi	&	200	&	folklorist	&	10	\\\hline
studi	&	252	&	gravit	&	166	&	chaucer	&	9	\\\hline
show	&	252	&	paraffin	&	145	&	epigram	&	8	\\\hline
effect	&	252	&	ionospher	&	56	&	panegyr	&	6	\\\hline

\end{tabular}
\label{table:catword3}
   \end{table}

\section{Dimension of the Meaning Space}\label{dim}
In this section, our focus is on answering the question: is 252 the actual number of dimensions in the Meaning Space (MS) for words of the Leicester Scientific Thesaurus (LScT)? We explore the Meaning Space by using Principal Component Analysis (PCA). We discuss the Double Kaiser Rule for the selection of important attributes, that is a subset of original attributes (categories).

The importance of words in each category was organised in the Word-Category RIG Matrix. With 5,000 words and 252 categories, it is difficult to see what information is present in the data. A way to understand the Meaning Space and visualise words' meaning (lexical meaning) in this space is to plot the words in individual categories. We expect that trends in some of plots will be very similar for a certain word, a group of words or all words. This may suggest that groups of words appear to be very close in specific categories. For instance, subcategories are very likely to occur in the corpus and related information will be measured by such attributes. This leads to some redundant attributes that measure the same information in multiple locations in the data. 

Words can be similarly represented in two or more categories. If two categories are correlated in the MS, it is possible to represent words in a reduced dimension by using a suitable linear combination of these original attributes. More specifically, if two categories are completely correlated, we would use the sum of two categories as one new attribute. The new attribute can be considered as a representative of the two original attributes. PCA provides a solution to this problem. Linear combination weights (coefficients) are provided by PCA to create the new attribute, which we term a principal component (PC), with the aim of preserving as much variability as possible (the maximum variation in the data). The level of the effectiveness of PCA in explaining the data varies differently with the different sets of PCs. Therefore, in this we investigate the effectiveness of PCA as a technique for determining the actual dimension of the data. Our goal is also to empirically investigate the effectiveness of the RIG based word representation technique using PCs instead of the original attributes. 

PCA is a statistical technique that transforms the data into a reduced-dimension represented by linearly uncorrelated attributes (PCs), where PCs are a linear combination of the original attributes \cite{dunteman,pearson}. The Kaiser Rule is one of the methods developed to select the optimal number of components \cite{guttman,kaiser}. Eigenvalues of the covariance matrix are used to determine the appropriate number by taking components with eigenvalues greater than 1; only components explaining greater data variance than the original attributes should be kept \cite{yeomans}.   

The Double Kaiser Rule is a method for selecting a subset of the original attributes based on their importance in representing the data. It evaluates PCs and the original attributes can be ranked in importance to the PC by the size (be it positive or negative) of their coordinate. The aim is to retain only a subset of the original attributes for further use. The advantage of employing the Double Kaiser Rule is that any remaining, so-called `trivial', attributes can be discarded iteratively. The iterative algorithm for the Double Kaiser Rule is shown in Algorithm \ref{doublekaiser} (the trace of a square matrix is the sum of its diagonal elements, and also the sum of its eigenvalues).

\makeatletter
\def\algbackskip{\hskip-\ALG@thistlm}
\makeatother

\begin{algorithm}
\caption{Double Kaiser Rule}\label{doublekaiser}

\begin{algorithmic}[1]

 \Require The Word-Category RIG Matrix and n (number of categories)
 \Ensure Set of informative categories
 \Repeat
\State Calculate the covariance Matrix $ \textbf{M} $ (standardised).
\State Calculate the Kaiser threshold ($\alpha$) using $\dfrac{{\rm trace}(M)}{n}$. 
\State $\alpha$ is 1 for correlation matrix. 
\State Select informative PCs by the Kaiser rule: Components with eigenvalues above $\alpha$ are informative.
\State Determine the importance of an attribute ($\beta$) as the maximum of the absolute values of coordinates in informative PCs for the attribute.
\State Select the informative attributes: (a) Calculate the threshold of importance for an attribute as the root mean squared of values in the coordinate. The threshold of importance for an attribute is $\dfrac{1}{\sqrt{n}}$ for unit vectors. (b) Select attributes which hold $\beta\geq\dfrac{1}{\sqrt{n}}$. 
\State Otherwise, the attribute is trivial. 
\State Drop the most trivial attribute from the Word-Category RIG Matrix if any.
\State $n=n-1$
\Until{There is no trivial attribute.}
\State \textbf{return} The set of informative attributes (categories).
\end{algorithmic}
\end{algorithm}

\begin{figure}[htpb]
    \centering
    \subfigure[]{{\includegraphics[width=7.5cm]{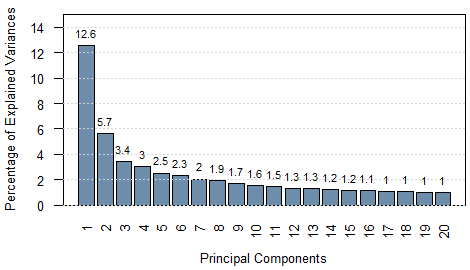} }}
     \hspace{2mm}
    \subfigure[]{{\includegraphics[width=4.5cm]{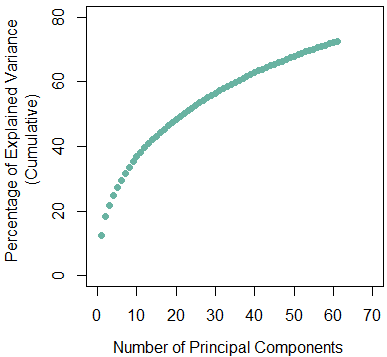} }}
    \caption{(a) Fraction of variance explained as a function of PCs retained for categories; (b) Cumulative fraction of variance explained as a function of PCs retained for categories (61 PCs)}
    \label{fig:FVE}
\end{figure}

PCs were assessed sequentially from the largest eigenvalue to the smallest. All PCs having eigenvalue less than 1 were considered to be trivial (non-significant) by the Kaiser rule. Hence 61 PCs are included as non-trivial, that is, 61 axes summarize the meaningful variation in the entire dataset. These non-trivial PCs are retained as informative at the first stage. The cumulative percentage of variance explained is displayed in Figure \ref{fig:FVE}. The cumulative percentage is approximately 73\%, indicating the variance accounted for by the first 61 components. They explain nearly 73\% of the variability in the original 252 attributes, so we can reduce the complexity of the data four times approximately, with only a 27\% loss of information.

Following the data reduction via PCA we then restricted the analysis of the informative categories to the non-trivial PCs; these are used to list informative attributes (categories). The importance of an attribute is determined as the maximum of the absolute values in coordinates of informative PCs for this attribute. The threshold $1/\sqrt{252} $ (threshold of importance) is used in the selection of informative attributes. None of the attributes was dropped after the first iteration of the Double Kaiser selection, so all 252 categories were retained.

To interpret each component, the coefficients (influence) of the linear combination of the original attributes for the first five principal components are examined (see Figures \ref{fig:PC1}, \ref{fig:PC2}, \ref{fig:PC3}, \ref{fig:PC4}, \ref{fig:PC5}). The coordinates of the attribute divided by the square root of the eigenvalue gives the unit eigenvector, whose components give the cosine of the angle of rotation of the category to the PC. Furthermore, positive values indicate a positive correlation between an attribute and a PC and negative values indicate a negative correlation. Both the magnitude and direction of coefficients for the original attributes are taken into account. The larger the absolute value of the coefficient, the more important the corresponding attribute is in calculating the PC. Positive and negative scores in PCs push the overall score of a word in the meaning space to the right or left on the PC axis. 

To examine the original attributes in the PCs, we introduce a threshold for categories having near zero values. The threshold used was $1/2\sqrt{252} $, which is half of the threshold of importance in selection of informative attributes. All values between $-1/2\sqrt{252} $ and $1/2\sqrt{252} $ are considered to be negligible so are in the \textit{zero interval}. Hence, the initial attributes are considered as belonging to three groups: (1) positive, (2) negative, and (3) zero. We interpret the categories belonging to positive and negative groups as the main coordinates of the dimension as these categories contribute significantly to that direction. Categories belonging to the `zero' group are seemed to be unrelated attributes to the PC. However, this information could be also useful. Hence, all categories in the three groups are meaningful and should be interpreted.  

\begin{figure}[!htb]
\begin{tabular}{cc}
    \begin{minipage}{0.6\textwidth} \includegraphics[width=1\textwidth]{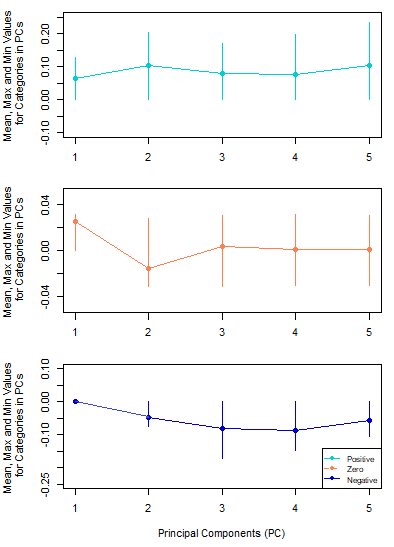} \end{minipage}& \begin{minipage}{0.5\textwidth} \includegraphics[width=0.82\textwidth]{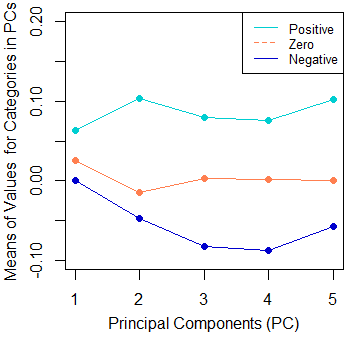} \\ \includegraphics[width=0.82\textwidth]{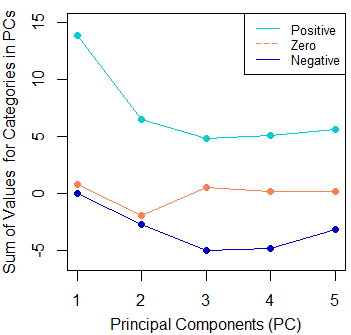} \end{minipage}
    \end{tabular}
         \caption{(Left) The average, maximum, and minimum values (coefficients of the linear combination of the original attributes) in five principal components for the groups of positive, negative and zero. (Top right) The average values in the groups of positive, negative and zero. (Bottom right) The sum of values in the groups of positive, negative and zero.}
    \label{fig:MaxMeanMin}
    \end{figure}

Categories in the three groups for each PC can be seen in Figures \ref{fig:PC1}, \ref{fig:PC2}, \ref{fig:PC3}, \ref{fig:PC4}, \ref{fig:PC5}. The zero interval is shown by a line in the figures. To understand the trend in the first five PCs, the average, maximum, and minimum values for positive, zero, and negative groups are displayed in Figure \ref{fig:MaxMeanMin}. Similarly, the sum of values in each group is plotted (see Figure \ref{fig:MaxMeanMin}). The number of categories in each group is presented in Table \ref{table:posneg}. 

Mean values in positive and negative groups are close in all PCs with a slight change for maximum and minimum values. We can see that for the second principal component, the difference between means of positive and zero groups is higher than all others. For the sum of values, the highest value belongs to the first PC as there is no negative value for this component. The full list of categories in positive, negative and zero groups for each of five PCs can be found in Appendix \ref{appndx}.

\begin{table}[bth]
	\centering
	\caption{Number of categories in the groups of positive, zero and negative for the first five principal components}
	\renewcommand\arraystretch{1.3}
\begin{tabular}{|c|c|c|c|c|c|}
\hline
&\textbf{PC1} &\textbf{PC2 }& \textbf{PC3} &\textbf{PC4 }& \textbf{PC5}  \\\hline
\textbf{Positive} & 221  & 63  &60 &67&55   \\\hline
\textbf{Zero} & 31  & 131  &131&129&142   \\\hline
\textbf{Negative} & 0 & 58  &61 &56&55   \\\hline
\end{tabular}
\label{table:posneg}
   \end{table}

The list of categories appearing in the zero interval in  the first five PCs are presented in Table \ref{table:allZero}. One may expect that as these categories contribute very small positive or negative values to the overall component score, they will not have a big effect in moving words with high RIG for that category, plotted in the Meaning Space, towards the associated end of the principal axis.

\begin{table}[bth]
	\centering
	\caption{Categories (initial attributes) appearing in zero interval in  the first five components with the number of texts assigned to the category}
	\renewcommand\arraystretch{1.3}
\begin{tabular}{|l|r|}
\hline
\textbf{Categories} & \textbf{Number of Texts}  \\\hline
Agriculture, Dairy \& Animal Science  &6,163  \\\hline
Andrology  & 391\\\hline
Astronomy \& Astrophysics & 22,825   \\\hline
Materials Science, Paper \& Wood & 1,963  \\\hline
Medicine, Legal & 1,711 \\\hline
\end{tabular}
\label{table:allZero}
   \end{table}

We can see that there are no negative values for the first principal component. The first component primarily measures the magnitude of the contribution of categories to the PC. It is a weighted average of of all initial attributes. The most prominent categories are 'Engineering, Multidisciplinary' and 'Engineering, Electrical\& Electronic', that is, they strongly influence the component. This component explains 12.58 \% of all the variation in the data (see Figure \ref{fig:FVE} (a) and Figure \ref{fig:PC1}). This means that more than 85\% of of the variation still retained in the other PCs. 

The second component has positive associations with categories related to social sciences and humanities, and negative associations with categories related to engineering and natural sciences (see Figure \ref{fig:PC2}). The plot shows that they are completely oppositely correlated. Hence, this component primarily measures the separation of two main branches of science. The most prominent category in the component is 'Cultural Studies'. This contributes a large positive value to overall component score, that is, it pushes the scores of words with high RIG for 'Cultural Studies' to the right on the axis. The largest negative contribution to the component score is from the category 'Engineering, Multidisciplinary', which is approximately 2.5 times smaller than the contribution of 'Cultural Studies'. In the zero interval, extremely low values are present for attributes such as 'Psychology, Developmental', 'Ergonomics' and 'Medicine, Legal'. Such attributes do not influence the movement of words to the extreme ends of this PC.

The largest positive values on the third component can be interpreted as contrasting the biological science, computer science and engineering related areas with medicine, social care and some other disciplines (Figure \ref{fig:PC3}). We may expect words that are used in biological science, computer science and engineering will go toward the positive side of the axis on the third principal coordinate. The largest negative values suggest a strong effect of psychology, medicine-health and physics related areas.

The other two principal components can be interpreted in the same manner (See Figure \ref{fig:PC4} and Figure \ref{fig:PC5}). In the fourth component, the most prominent categories with positive values are some of social science branches such as economics, managements, psychology, ethics, education and multidisciplinary social science. It can be seen large negative values for categories related to literature and medicine-health science. The fifth component has large positive associations with ecological, environmental sciences and geosciences. 
   
\begin{landscape}\centering
\thispagestyle{plain} 
\begin{figure}[htpb]
  \centering
  \includegraphics[height=.82\textheight, width=1.8\textwidth]{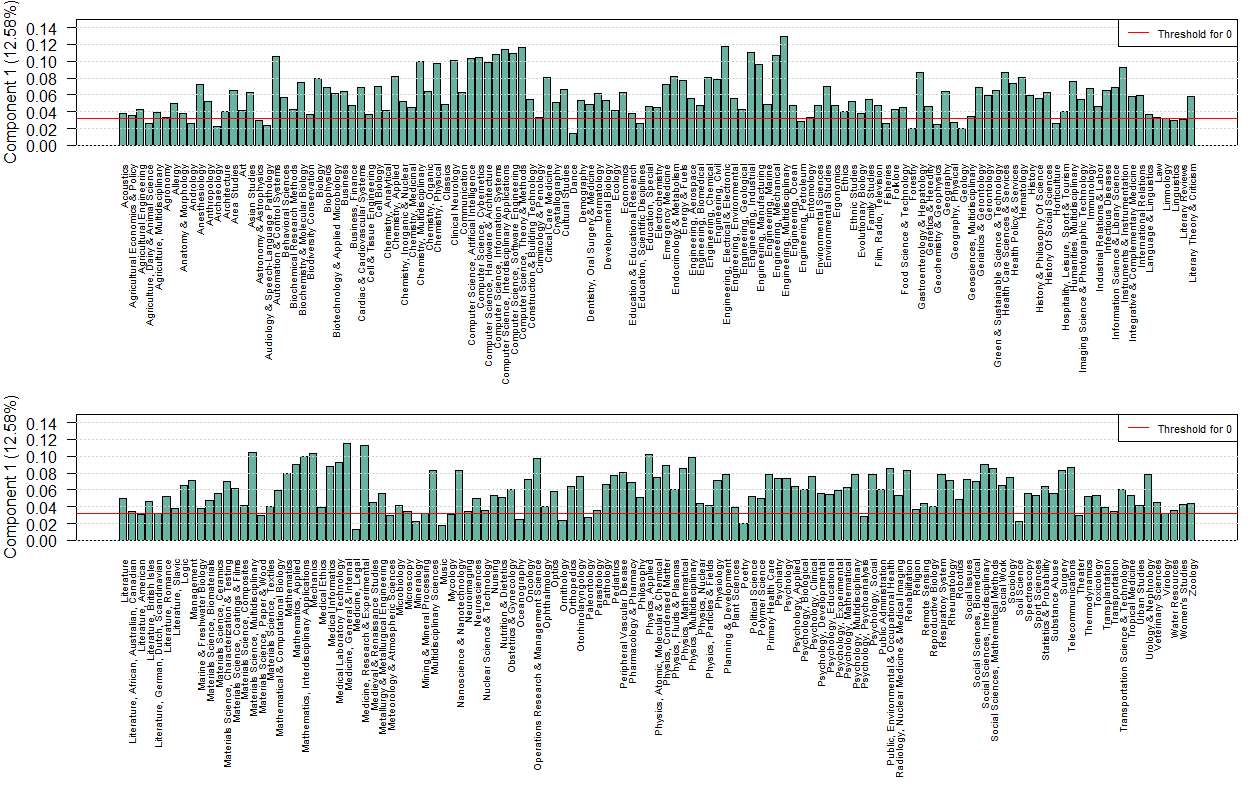}
  \caption{The first principal component of the LScT. The plot shows the contributions of original attributes (categories) on the first principal component.}
  \label{fig:PC1}
\end{figure}
\end{landscape}

\begin{landscape}\centering
\thispagestyle{plain} 
\begin{figure}[htpb]
  \centering
  \includegraphics[height=.82\textheight, width=1.8\textwidth]{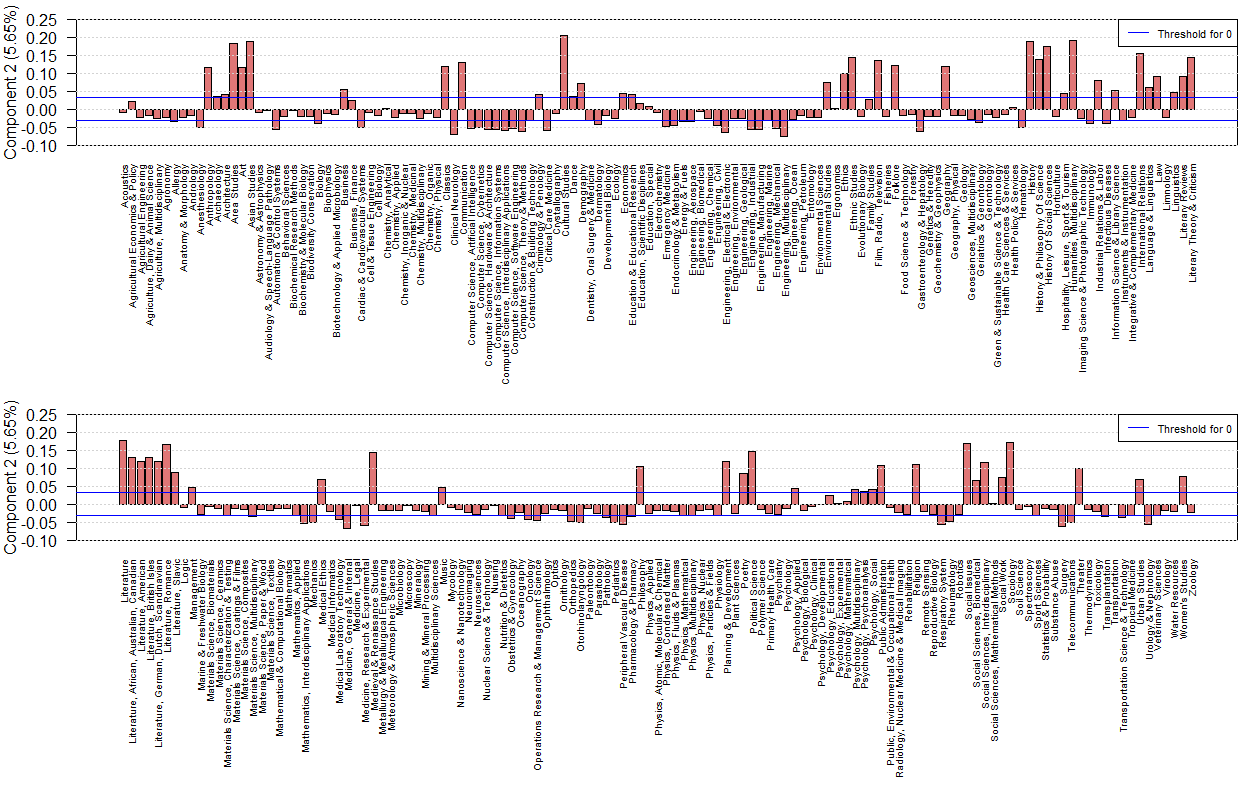}
  \caption{The second principal component of the LScT. The plot shows the contributions of original attributes (categories) on the second principal component.}
  \label{fig:PC2}
\end{figure}
\end{landscape}

\begin{landscape}\centering
\thispagestyle{plain} 
\begin{figure}[htpb]
  \centering
  \includegraphics[height=.82\textheight, width=1.8\textwidth]{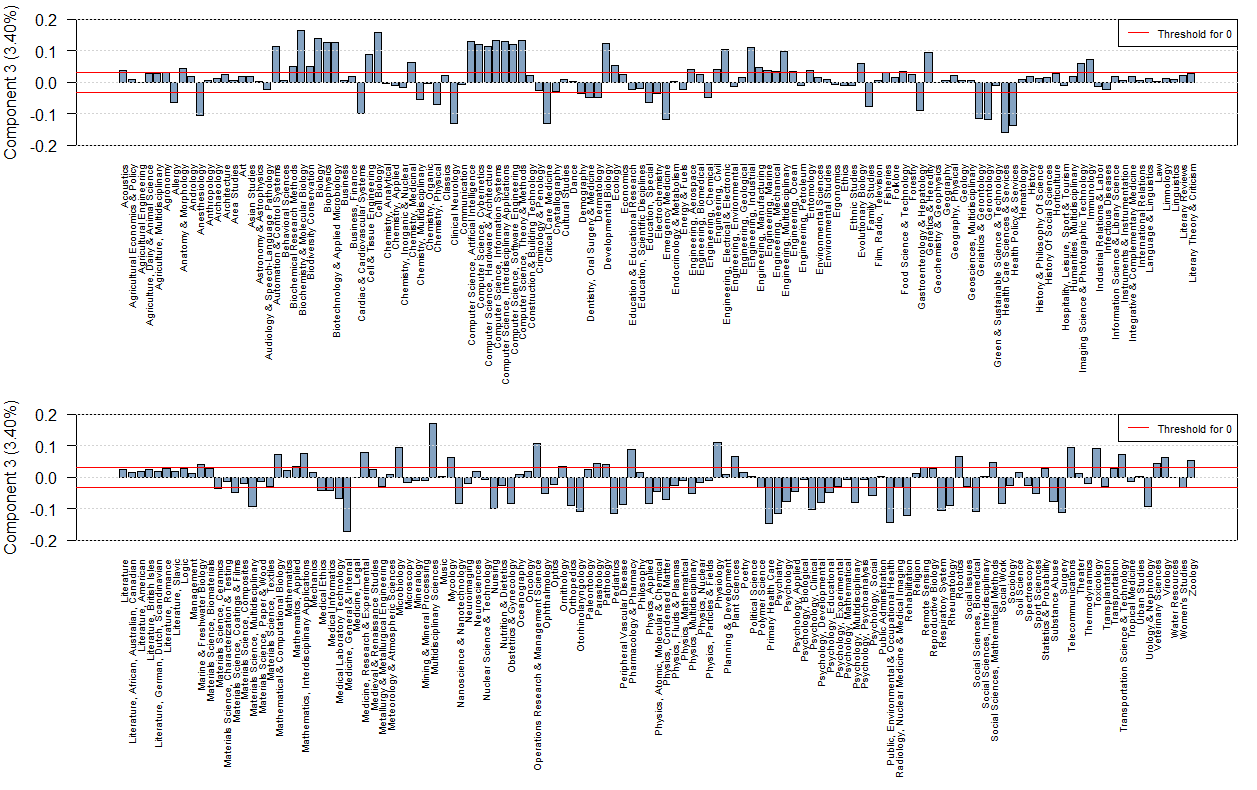}
  \caption{The third principal component of the LScT. The plot shows the contributions of original attributes (categories) on the third principal component.}
  \label{fig:PC3}
\end{figure}
\end{landscape}

\begin{landscape}\centering
\thispagestyle{plain} 
\begin{figure}[htpb]
  \centering
  \includegraphics[height=.82\textheight, width=1.8\textwidth]{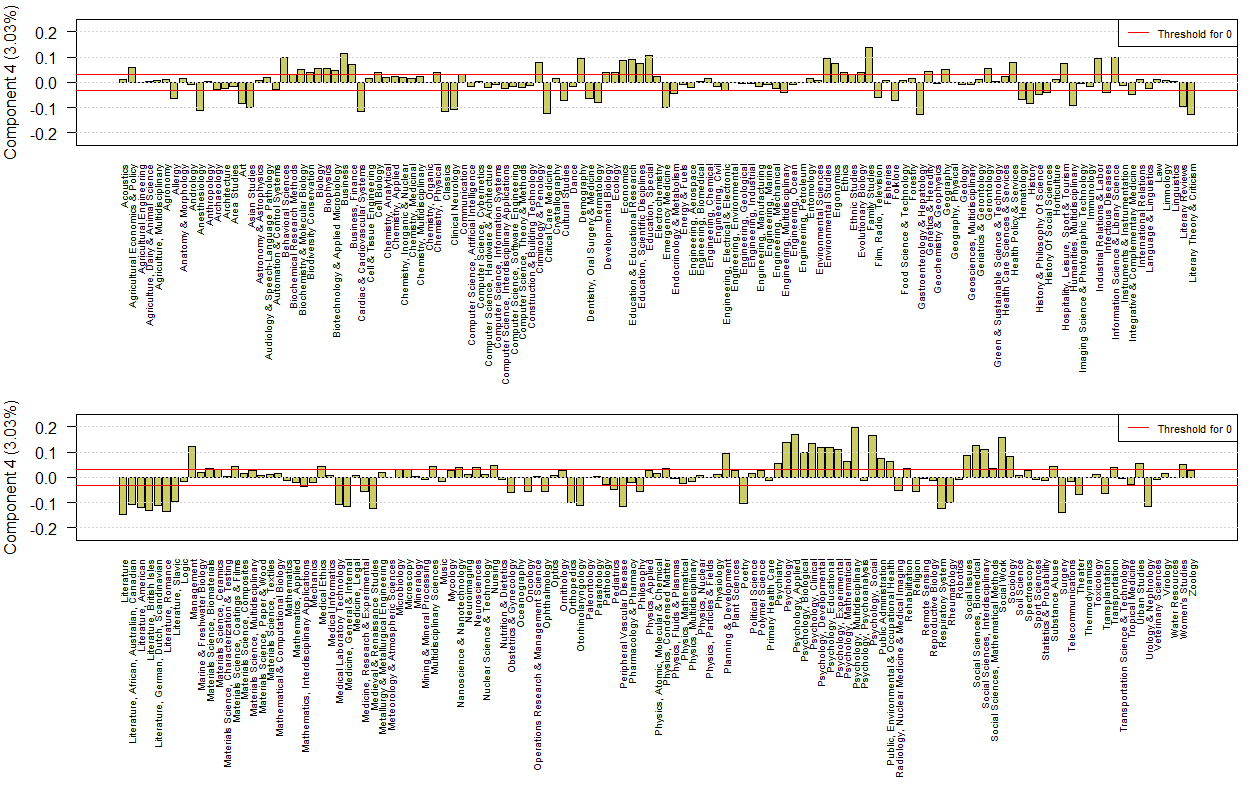}
  \caption{The fourth principal component of the LScT. The plot shows the contributions of original attributes (categories) on the fourth principal component.}
  \label{fig:PC4}
\end{figure}
\end{landscape}

\begin{landscape}\centering
\thispagestyle{plain} 
\begin{figure}[htpb]
  \centering
  \includegraphics[height=.82\textheight, width=1.8\textwidth]{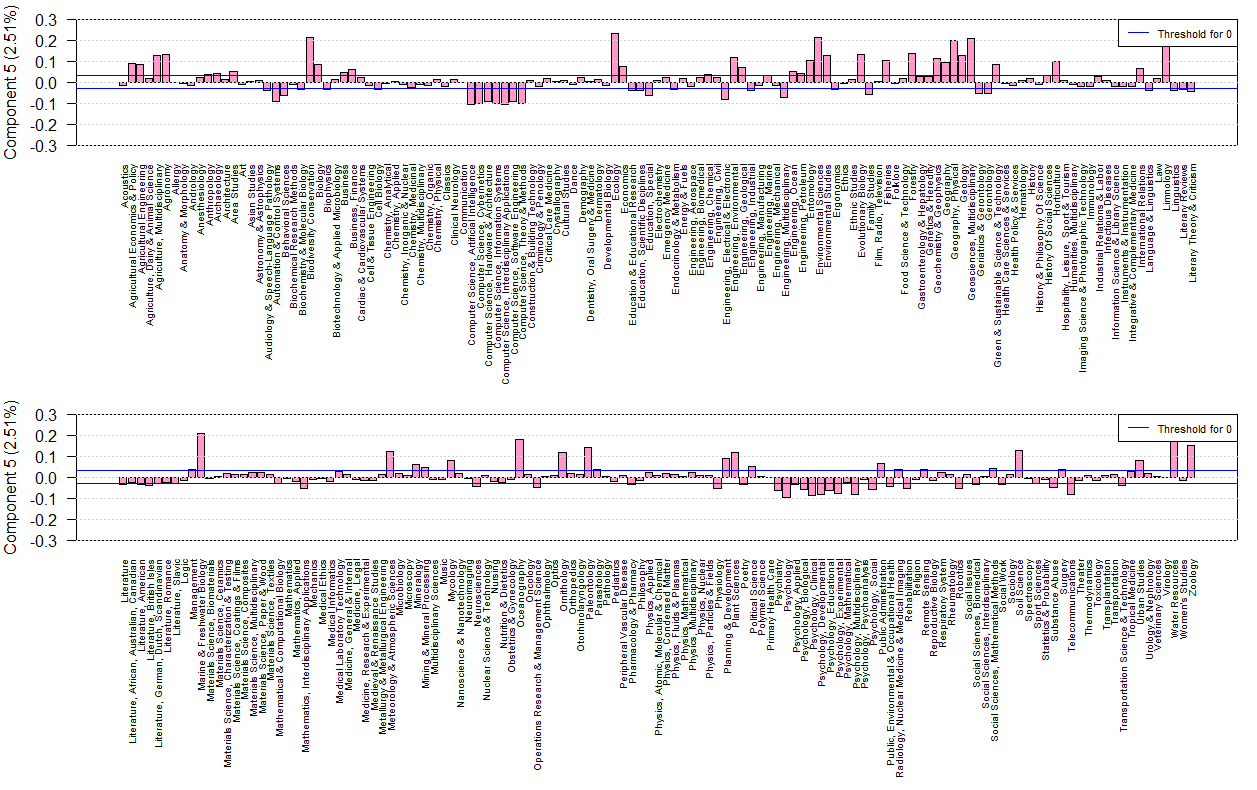}
  \caption{The fifth principal component of the LScT. The plot shows the contributions of original attributes (categories) on the fifth principal component.}
  \label{fig:PC5}
\end{figure}
\end{landscape}

\begin{figure}[htpb]
    \centering
    \subfigure[]{{\includegraphics[width=6cm]{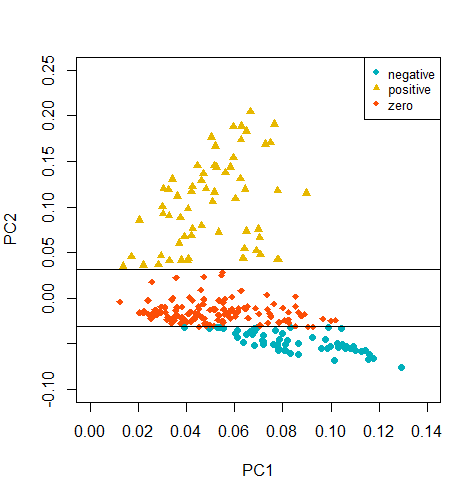} }}
     \hspace{1mm}
    \subfigure[]{{\includegraphics[width=6cm]{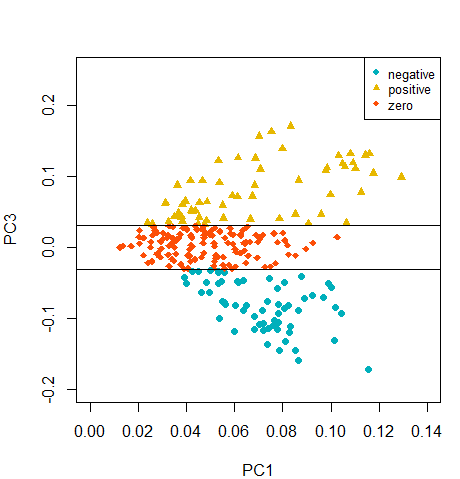} }}
    \caption{(a) The first principal component versus the second principal component of the original attributes (categories) (b) The first principal component versus the third principal component of the original attributes.}
    \label{fig:2DCAT1}
\end{figure}

We have seen that individual components are not fully explanatory of variations in the data. For instance, the first component explains only 12.58\% of the variation, and it is much less for subsequent PCs. The first two components explain about 18.3\% of the variation in the data. To gain insight about correlated attributes in a certain portion of the variation, it is reasonable to plot two PC axes. The Figure \ref{fig:2DCAT1} (a) shows the results for the first two components. The influence on the variance of the PC axis is low for those categories near to the intercept of zero interval.

The category 'Engineering, Multidisciplinary' has a large positive value to the extreme right on PC1. As we mentioned before, this component measures the magnitude of contribution of the original attributes to the component. For PC2, categories in positive, negative and zero groups are shown in different colors. This indicates that some attributes are correlated in approximately 18\% of the variation represented by the first two PCs.

It can be seen that categories in the positive group on PC2 are not at the extreme right on the PC1 axis. Also, it is apparent that attributes in the positive area are not as well correlated (dense) as attributes in the negative and zero areas. The attribute 'Medicine, Legal' appears close to zero in both components, which indicates that these components are not indicative of variations for this attribute.

Figure \ref{fig:2DCAT1} (b) is a comparison of PC1 with PC3. We note particularly that the density of categories in the positive and negative areas are similar, having almost equal distribution for negative and positive values in the zero interval. Attributes in the zero interval seem to be much more dense denser so more correlated than attributes in the other two groups. However, these two components do not reflect the variation for the zero interval.

\begin{figure}[htpb]
    \centering
    \subfigure[]{{\includegraphics[width=6cm]{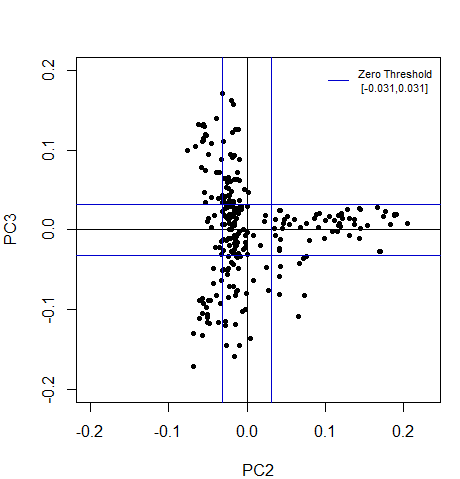} }}
     \hspace{1mm}
    \subfigure[]{{\includegraphics[width=6cm]{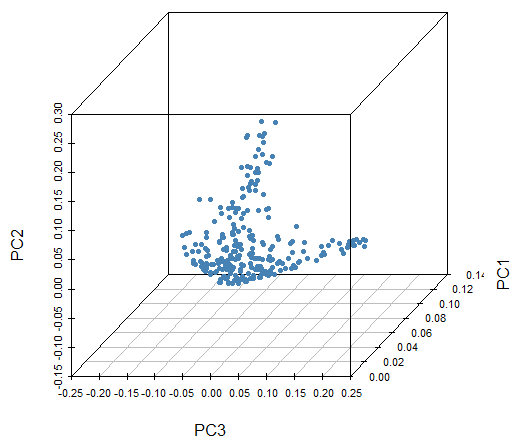} }}
    \caption{(a) The second principal component versus the third principal component of the original attributes (categories) (b) The first three principal components of the original attributes.}
    \label{fig:3DCAT}
\end{figure}

The second and the third principal components, which account for 9.1\% of the variance in the data, are plotted in Figure \ref{fig:3DCAT} (a). If attributes are inversely correlated, they are positioned on opposite sides of the origin in this plot. These two components seem to have correlated attributes (high density) mainly in the zero-zero area. It is noticeable in the plot that the majority of categories have negative and positive values in one of components when they are in zero interval for the other component. This implies that certain  attributes are at the extremes of a spread in one PC, and not in the other. Also, there are attributes in the area of negative-positive, negative-negative and positive-negative, but not in positive-positive. This indicates that in a qualitative sense, categories are associated with PCS in very different ways. This result is consistent with the observations from the Figure \ref{fig:3DCAT} (b), a three dimensional picture in which we can observe categories stretching out along the PC axes, like the tail and wings of a bird.

\subsection{Analysis of Extreme Topic Groups at Opposite Ends of the PCs\nopunct}\hspace*{\fill} \\\\
As mentioned above, we applied the Double Kaiser rule to evaluate PCs and select a subset of original attributes based on their importance for the PCs. The importance of attributes was defined by a 'threshold of importance', which was $ 1/\sqrt{252} $. All attributes appeared as non-trivial, so we retained all 252 categories. Then, we used the half of the value $ 1/\sqrt{252} $  to define a threshold for categories having near zero values under the assumption that as those categories having very low contribution to the PC do not have a strong influence on the PC, they are not correlated to the corresponding PC and so not the main coordinates of the dimension. All values of component between $ -1/2\sqrt{252} $ and $ 1/2\sqrt{252} $  were considered into zero interval. So, three groups of categories were defined based on this threshold: (1) positive (2) negative (3) zero. We then analysed PCs based on categories in each groups.

In this section, we analyse the topic groups at opposite ends of the PCs (positive and negative ends) in order to describe the PCs based on extremely influential categories at both ends. As such categories have high contributions in the PC, they are the parts of the trends in PCs and so explain the general trends of PCs. This implies that we consider positive and negative groups introduced before, select the top $n$ categories with the highest component coefficients in each group and describe the grouping of categories in a way that categories at extreme ends can be distinguished from each other somehow and meaningfully described by a classification of research fields in science.

A simple way to explain a PC by groups of categories at the both ends of the PC is by looking at the scientific categorisation of the categories and words used in the texts of grouped categories. In particular, as there exist two sets of categories that are located at opposite ends of the PC, it would be interesting to extract the most influential common words used in each of this extreme topic groups. It is important to note that as common words appear in all selected categories as informative words, they help to describe categories and then PCs.

We implemented a heuristic technique. This approach starts with a search for a set of 10 categories with maximum coefficients at the two ends of the PC. The most informative 150 words are extracted in each of 10 categories, and the common words are listed. Words are analysed by human inspection to understand the meaning behind the opposite ends of the PC. The procedure is repeated for the PC2, PC3, PC4 and PC5. For the first PC1, the sign of coefficients are positive for  all categories. High numbers for categories in this PC indicate that that category is well-described by words in the LScT.

Both the top 10 categories and the most influential words in the set of categories are used to extract information for describing the PCs. The description is now more focussed on the big picture of science rather than description based on branches/sub-branches.

The second PC seems to correspond a separation between discourse studies and experimental studies when we consider both the categories and words. For example, it is seen that three of the most informative common words are 'argu', 'polit' and 'discours' for the groups of categories in the positive side and  three of the most informative common words are 'clinic', 'treatment' and 'therapi' for the groups of categories in the negative side in the PC. This is the {\bf Nature of Science} dimension.

The third PC reflects two opposite types of research in terms of the requirement of microscopic and macroscopic instruments (see Table \ref{table:PC3cats}). At the positive end, scientific research mostly required detailed tools to work with the objects. Such tools can be instruments such as the microscope as well as programming tools used in coding. On the negative end, we are talking about human and  population scale objects, but still related to humans. So this is the {\bf Human Scale} dimension.

\begin{table}[bth]
	\centering
	\caption{Categories at opposite ends of the PC3}
	\renewcommand\arraystretch{1.3}
\begin{tabular}{|L{6cm}|L{6cm}|}
\hline
\textbf{Categories in the positive side of the PC3} & \textbf{Categories in the negative side of the PC3}  \\\hline
Multidisciplinary Sciences    & Medicine, General \& Internal   \\\hline                                   
Biochemistry \& Molecular Biology  &           Health Care Sciences \& Services   \\\hline                
Cell Biology                        &           Primary Health Care           \\\hline                  
Biology                              &            Public, Environmental \& Occupational Health \\\hline
Computer Science, Information Systems &           Health Policy \& Services \\\hline                     
Computer Science, Theory \& Methods    &           Critical Care Medicine    \\\hline                  
Computer Science, Interdisciplinary Applications & Clinical Neurology       \\\hline                     
Computer Science, Artificial Intelligence        & Rehabilitation  \\\hline
Biotechnology \& Applied Microbiology             & Gerontology \\\hline
Biophysics &  Emergency Medicine  \\\hline
\end{tabular}
\label{table:PC3cats}
   \end{table}  

The fourth component appears as to describe two classes of science: science of understanding the human condition through experiments and science of understanding the human condition through critical discourse studies. For instance, literary studies in the negative side are prominent and many texts from the literature are literary criticism of works.   This is the {\bf Human Condition} dimension.

Finally, the fifth component can be interpreted as contrasting the natural science and the intelligence. Categories related to natural science researches are grouped in the positive extreme side and categories of understanding intelligence are located in the negative extreme side in this PC. 'Intelligence' can be both human intelligence and machine intelligence. For example, the categories 'Computer Science, Artificial Intelligence' and 'Psychology' are two of the top 10 categories.  This is the {\bf Inner World/Outer World} dimension.

\subsection{Visualisation of Words in the Space of PCs\nopunct}
\begin{figure}[htpb]
    \centering
    \subfigure[]{{\includegraphics[width=6cm]{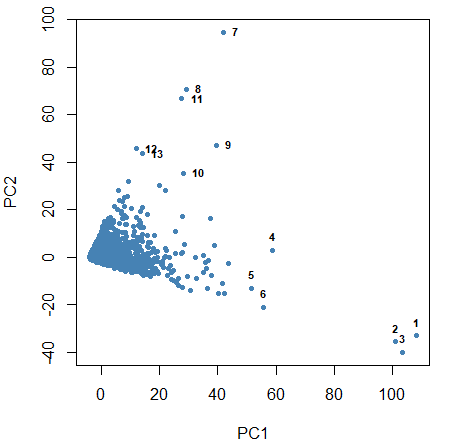}}}
     \hspace{1mm}
    \subfigure[]{{\includegraphics[width=6cm]{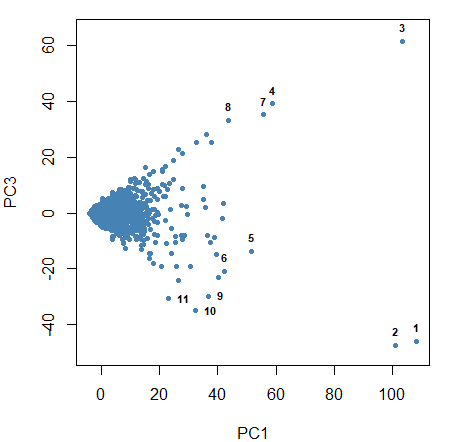}}}
    \caption{The PCA score plot of words. (a) First and second PC axes with principal component scores (b) First and third PC axes with principal component scores.}
    \label{fig:2dword1}
\end{figure}

\begin{figure}[htpb]
    \centering
    \subfigure[]{{\includegraphics[width=6cm]{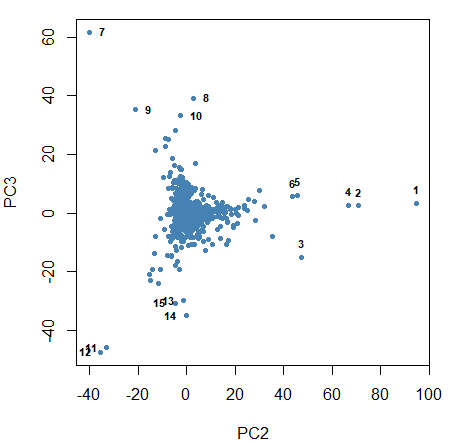}}}
     \hspace{1mm}
    \subfigure[]{{\includegraphics[width=6cm]{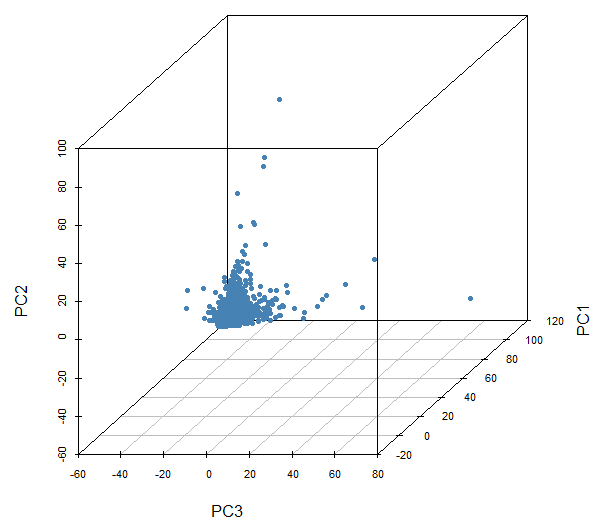}}}
    \caption{The PCA score plot of words. (a) Second and third PC axes with principal component scores (b) The first three PC axes with principal component scores.}
    \label{fig:2dword2}
\end{figure}

\hspace*{\fill} \\\\
We now move onto visualizing words on the principal axes. Figure \ref{fig:2dword1} and \ref{fig:2dword2} are visualisations of words in the space of any of the first three components and the first three PCs. PC scores are calculated for each word to determine its location on the PCs. Those words that have negative scores when projected onto a PC will represent categories at that end of the PC. The majority of words are located around the origin in the figures, showing that relatively few words are strong indicators for any category. However, some words seem to be distinctly different from others and grouped together in the plots. This suggests that even a small variation in the data (such as 12.58\% and 5.65\%) helps to distinguish certain groupings of words.

Words further apart on the PC map are more different from each other than words grouped together. Labelled words in are presented in the Table \ref{table:PCwords}. We can see that the words 'patient', 'paper', 'conclus', 'articl' and are located far from other words. It is worth mentioning that these words were the most informative words in the LScT. There are other interesting features of these plots. Words grouped closer together and far from the origin have similarity with respect to the academic disciplines they appear in.  For example, the two words 'argu' and 'polit' are far from the origin and close together in Figure \ref{fig:2dword1} (a). They both are expected to be included in texts for social science. Similarly, the words 'essay' and 'centuri' are close in the plot.

\begin{table}[bth]
	\centering
	\caption{Words labelled in Figure \ref{fig:2dword1} and Figure \ref{fig:2dword2}.}
	\renewcommand\arraystretch{1.3}
\begin{tabular}{|c|c|c|c|}
\hline
\textbf{Label}& \textbf{PC1-PC2}&\textbf{PC1-PC3}&\textbf{PC2-PC3}  \\\hline
1& patient & patient&articl   \\\hline
2& conclus &paper&argu\\\hline
3& paper &conclus&social   \\\hline
4& cell & cell&polit  \\\hline
5& clinic &propos&essay \\\hline
6& propos & clinic&centuri \\\hline
7& articl &protein&paper \\\hline
8& argu & background&cell \\\hline
9& social &care&propos \\\hline
10& result &health& protein \\\hline
11& polit &particip&patient \\\hline
12& essay & &conclus \\\hline
13& centuri && particip \\\hline
14& &&health \\\hline
15&  && care\\\hline
\end{tabular}
\label{table:PCwords}
   \end{table}

In Figure \ref{fig:2dword1} (b), the words 'cell' and 'protein' are near to each other. A group of words 'health', 'care' and 'particip' are also co-located, which is to be expected as they are likely to be used together in medical texts. The same behaviour can be observed in Figure \ref{fig:2dword2} (a). Words that are likely to be used in social science related articles go towards the positive end of the PC2 axis, and words that are likely to be used in medical texts go towards negative end. Words that are likely to be used in biological science related articles are towards the positive end of the PC3 axis. Words in these three regions are given in Table \ref{table:PCwords2}. We list words where the scores on
\begin{itemize}
\item PC2 are greater than 20 and less than -10;
\item PC3 are greater than 20 and less than -20.
\end{itemize}
We observe a difference of topic-specific words in medicine, social science and  biological science related articles. As we expected, words that are used in biological, computer and engineering science related areas are towards the positive end of PC3; see Figure \ref{fig:PC3}.

\begin{table}[h]
	\centering
	\caption{Words when the scores on PC2 are greater than 20 and less than -10,  words when the scores on PC3 are greater than 20 and less than -20 in Figure \ref{fig:2dword1} and Figure \ref{fig:2dword2}.}
	\renewcommand\arraystretch{1.3}
\begin{tabular}{|c|c|c|c|}
\hline
\textbf{PC2$>$20}& \textbf{PC2$<$-10}&\textbf{PC3$>$20}&\textbf{PC3$<$-20}  \\\hline
article& patient & speci&patient\\\hline
argu& conclus &paper&conclus\\\hline
social& paper &cell&age   \\\hline
result& background& protein&background \\\hline
polit& clinic &propos&particip\\\hline
essay& propos & gene&health \\\hline
cultur& age &algorithm&object \\\hline
centuri&  diseas& express&care \\\hline
polici& associ &problem& \\\hline
text& year&&  \\\hline
public& object&& \\\hline
literari& outcome & &\\\hline
discours& problem &&  \\\hline
histori& && \\\hline
contemporari&  && \\\hline
govern&  && \\\hline
draw&  && \\\hline
scholar&  && \\\hline
war&  && \\\hline
narrat&  && \\\hline

\end{tabular}
\label{table:PCwords2}
   \end{table}
   
\subsection{Visualisation of Categories into the Word Space \nopunct}\hspace*{\fill} \\\\
The data can be represented as a set of points in two different spaces: category space and word space. In the category space every point (vector of dimension 252) is a word, represented by its RIGs in 252 categories. In the word space every point (vector of dimension 5,000) is a category, represented by RIGs of 5,000 words. Results on the visualisations in the category space were provided in the previous section. Here we provide results on the visualisation of the data in the word space. The analysis in this section answers the question of how successfully categories are distinguished into PC space constructed in the word space.

\begin{figure}[h]
  \centering
  \includegraphics[height=.19\textheight, width=1\textwidth]{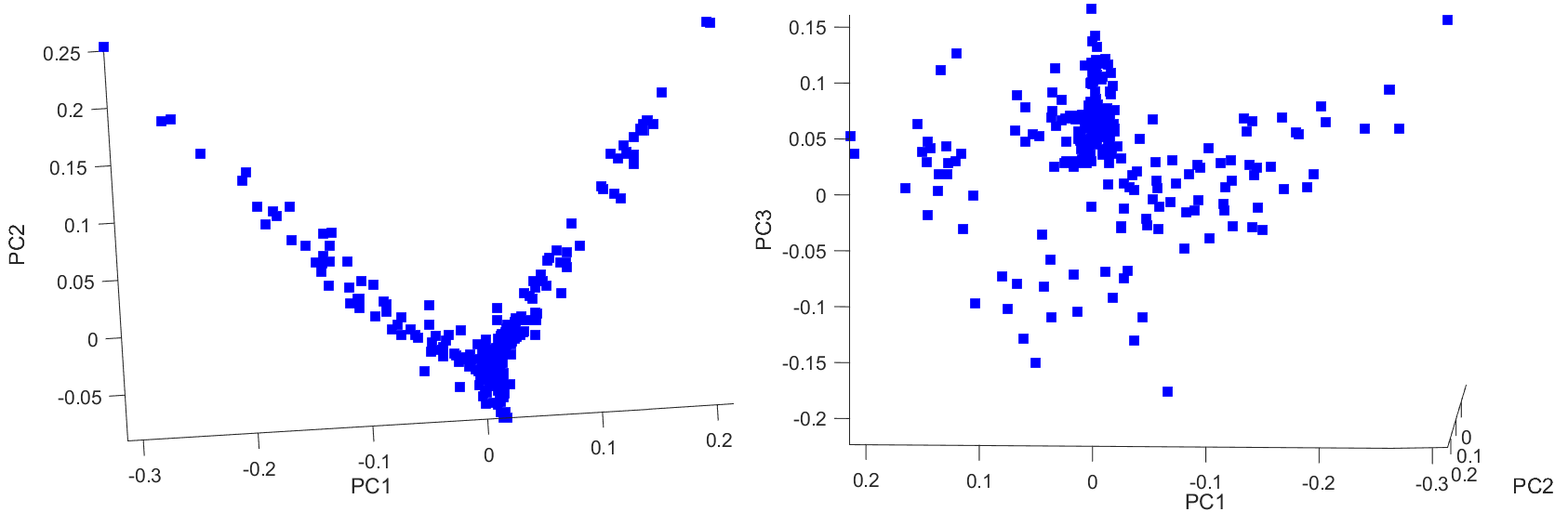}
  \caption{Visualisation of categories in the first two and three PCs}
  \label{fig:2PCsCat}
\end{figure}

\begin{figure}[h]
  \centering
  \includegraphics[height=.3\textheight, width=1\textwidth]{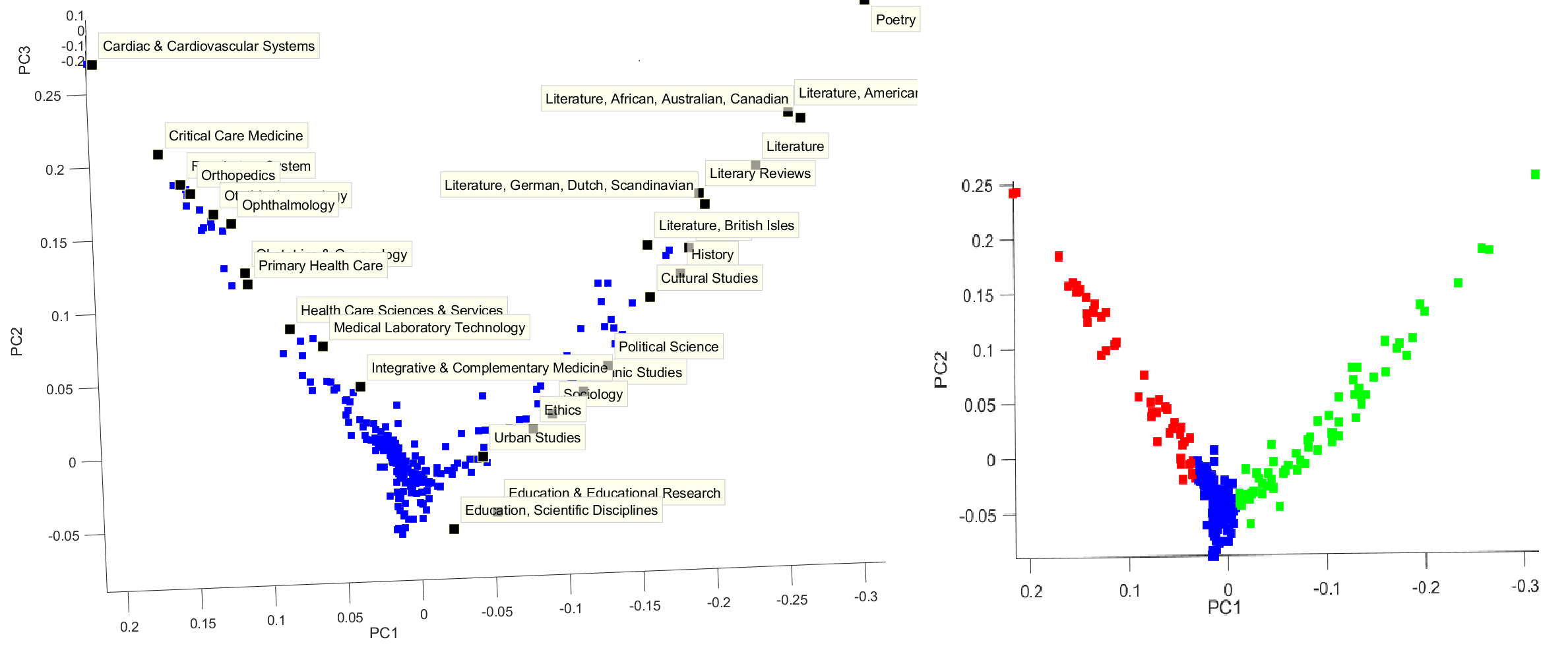}
  \caption{Visualisation of the annotate version of categories and colouring of the bird-like shape}
  \label{fig:3PCsCat}
\end{figure}

Categories are initially visualised in the PC axes (see Figure \ref{fig:2PCsCat}). The figure demonstrates the data in the first two and three PCs and the obtained bird-like shape. Annotated version of categories in the space of the first three PCs is presented in Figure \ref{fig:3PCsCat}. One can see the bird-like shape of the graphs has a meaningful separation of categories on two wings in terms of the branches of science. The left wing carries medicine related categories, while the right wing carries the literature and social science categories. At the right figure, the bird is coloured: red and green wings, and blue body. Lists of categories for each part of the bird with the projection on the first PC are presented in Table \ref{left}. Colouring indicates the categories of social science (green), categories of medicine (red) and all other categories (blue).  They all support the general findings. 

\subsection{Identifying Groups of Categories in PCs by Approximation of Vectors\nopunct}\hspace*{\fill} \\\\
We now introduce another approach for determining the groups of categories which are influential for PCs. We start with introducing the method for determining two such groups of categories in a specific PC.
 
Suppose we have a given $ n $-dimensional vector of component coefficients in an arbitrary PC, defined as
$$\textbf{u}=(u_1,u_2,\cdots,u_n ),$$
with categories ordered so that $u_1>u_2>\cdots>u_n$ (the chance of two being equal is negligible). Our assumption is that two classes of categories can be separated at some point $u_k$, where categories are correlated to each other within each class. That is, we aim at finding the optimal $k$ in 
$$\textbf{u}=(u_1,u_2,\cdots,u_k,\cdots,u_n ),$$
where those categories with coefficients $u_1,u_2,\cdots,u_k$ will belong to the first class and those categories with coefficients $u_{k+1},u_{k+2},\cdots,u_n$ will belong to the second class. To find the optimal $k$, we use a method based on the approximation of vectors. 

Let us consider the binarized vector of two classes
$$\textbf{e}=(1,1,\cdots,1,0,\cdots,0),$$
in which elements of the first class are transformed to 1 and elements of the second class are transformed to 0. We assume that the best approximate to the given vector $\textbf{u}$ is a vector aligned in the direction of the vector $\textbf{e}$, that is, the approximate vector
$$\alpha \textbf{e}=(\alpha,\alpha,\cdots,\alpha,0,\cdots,0)$$
for some $\alpha\in\mathbb{R}$, where the number of elements with the value $\alpha$ is $k$. The vector $\textbf{e}$ is a basis vector in the space. The approximation of the vector implies that the Euclidean distance between vectors $\textbf{u}$ and $\alpha\textbf{e}$ should be minimum. Therefore, there is an $\alpha$  that minimizes the length of the difference between the approximation $\alpha\textbf{e}$ and the given $\textbf{u}$. Our aim is now to find the constant $\alpha$ such that minimizes
$$W_{1}=\parallel\textbf{u}-\alpha\textbf{e}\parallel$$
is minimised.

We can simplify the algebra by minimizing the square of the norm
$$W_{1}^2=\parallel\textbf{u}-\alpha\textbf{e}\parallel^2.$$
Minimizing $W_{1}^2$ implies finding $\alpha$ such that
\begin{eqnarray*}
\dfrac{\partial W_{1}^2}{\partial \alpha} &=&\dfrac{\partial}{\partial \alpha}\left(\sum\limits_{i=1}^k (u_{i}-\alpha)^2+\sum\limits_{i=1+k}^n u_{i}^2\right)=\dfrac{\partial}{\partial \alpha}\sum\limits_{i=1}^k (u_{i}-\alpha)^2\\
& = & -2\sum\limits_{i=1}^k (u_{i}-\alpha)=0.
\end{eqnarray*}

Solving this for $\alpha$ gives

$$\alpha=\dfrac{1}{k}\sum\limits_{i=1}^k u_{i}.$$

This means that $\alpha$ is the average of values in the first class and the approximate vector is

$$\alpha\textbf{e}=(\underbrace{\alpha,\cdots,\alpha}_{\text{k times}},\underbrace{0,\cdots,0}_{\text{n-k times}}).$$

To determine the optimal $k$, we search for the best approximation to the vector of component coefficients among those created for $k=1,\cdots,n-1$. Once the optimal $k$ is found, we will be able to classify categories into two classes. 

To identify three groupings of categories in a PC, we first consider the approximate vector
$${\bf v}(\alpha,\beta,k,r)=(\alpha,\cdots,\alpha,0,\cdots0,\beta,\cdots\beta),$$
where the number of elements with the value $\alpha$ is $k$, the number of elements with the value $\beta$ is $r$ and $k+r<n$. Our aim to find the optimal $k$ and $r$ that minimize $W_2=\parallel\textbf{u}-{\bf v}(\alpha,\beta,k,r)\parallel$. This implies

\begin{equation}
\begin{split}
\dfrac{\partial W_{2}^2}{\partial \alpha}=\dfrac{\partial}{\partial \alpha}\left(\sum\limits_{i=1}^k (u_{i}-\alpha)^2+\sum\limits_{i=1+k}^n-r u_{i}^2+\sum\limits_{i=n-r+1}^n (u_{i}-\beta)^2\right) \\
=\dfrac{\partial}{\partial \alpha}\sum\limits_{i=1}^k (u_{i}-\alpha)^2=-2\sum\limits_{i=1}^k (u_{i}-\alpha)=0
\end{split}
\end{equation}
and 
\begin{equation}
\begin{split}
\dfrac{\partial W_{2}^2}{\partial \beta}=\dfrac{\partial}{\partial \beta}\left(\sum\limits_{i=1}^k (u_{i}-\alpha)^2+\sum\limits_{i=1+k}^n-r u_{i}^2+\sum\limits_{i=n-r+1}^n (u_{i}-\beta)^2\right)\\
=\dfrac{\partial}{\partial \beta}\sum\limits_{i=n-r+1}^n (u_{i}-\beta)^2=-2\sum\limits_{i=n-r+1}^n (u_{i}-\beta)=0.
\end{split}
\end{equation}
We note that finding $\alpha$ and $\beta$ are independent of each other. Thus we have
$$\alpha=\dfrac{1}{k}\sum\limits_{i=1}^k u_{i},  \; {\rm and} \; \beta=\dfrac{1}{r}\sum\limits_{i=n-r+1}^n u_{i}.$$

If we want to identify three groupings of categories as above, we will search for the optimal $k$ and $r$ independently. We apply the procedure for two classes to the vector $\textbf{u}$ in two steps: 1) finding the optimal $k$ in the vector $\textbf{u}$ where elements $u_i$ are sorted in descending order; 2) finding the optimal $r$ in the vector $\textbf{u}$ where elements $u_i$ are sorted in ascending order. 

\subsubsection{Results\nopunct}\hspace*{\fill}
The grouping of categories into PCs is done by using the vector approximation described in the previous section. The numbers of categories in the positive, negative and zero groups are given in  Table \ref{table:groups123}. The full lists of categories in each group for PC2, PC3, PC4 and PC5 can be found in Table \ref{appndx2}.

\begin{table}[bth]
	\centering
	\caption{Number of categories in the positive, zero and negative groups identified by vector approximation based approach for the first five principal components}
	\renewcommand\arraystretch{1.3}
\begin{tabular}{|c|c|c|c|c|c|}
\hline
&\textbf{PC1} &\textbf{PC2 }& \textbf{PC3} &\textbf{PC4 }& \textbf{PC5}  \\\hline
\textbf{Positive} & 221  & 46  &41 &44&37  \\\hline
\textbf{Zero} & 31  & 91  &159&160&158   \\\hline
\textbf{Negative} & 0 & 115 &52 &48&57   \\\hline
\end{tabular}
\label{table:groups123}
   \end{table}

The results using the vector approximation based grouping method turn out to show similar patterns to those using the threshold  $ 1/2\sqrt{252} $. In general, there is no clear change of trends in PCs excluding PC2, where psychology related areas were split between the positive and zero group previously. The vector approximation method reduces their importance, placing them in the zero group. This does not conflict with our hypothesis that this PC is related to the difference between discourse and experiment in what we suggest is the Nature of Science dimension. It is certainly not unreasonable to exclude psychology from the group of discourse studies. More categories, where experimental methods of research are used, now appear in the negative category; For instance, the categories 'Developmental Biology' and 'Biochemistry \& Molecular Biology' are now included.  

Another interesting finding is that the positive group for PC5 contains only environmental science related areas, and excludes economics and public administration which appear in the tail of the list. Some social science categories such as 'International Relations', Business, Finance' and 'Political Science' were in the positive group, but are categorised into the zero group zero by the vector approximation based grouping method. Again, this does not conflict with the idea that PC5 is the Inner World/Outer World dimension.

\subsection{Deciding the Dimension of the Meaning Space \nopunct} \label{dimspace} 
\begin{figure}[htb]
  \centering
  \includegraphics[height=.4\textheight, width=.6\textwidth]{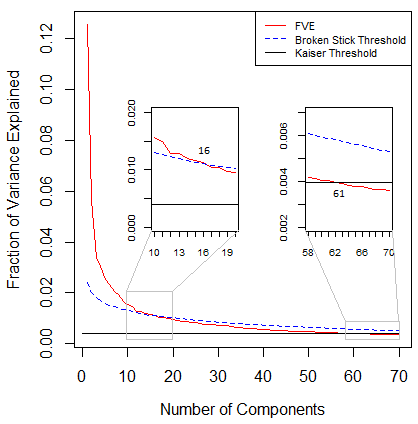}
  \caption{The number of principal components determined based on the Kaiser rule and the Broken Stick rule}
  \label{fig:Broken}
\end{figure}

\hspace*{\fill} \\\\
The number of principal components determined by the Kaiser rule was 61. However, the Kaiser rule can underestimate or overestimate the number of PCs to be retained \cite{zwick}. So, we also tested the Broken-Stick rule to determine the number of PCs \cite{jackson,king,peres,cangelous}. Figure \ref{fig:Broken} demonstrates the optimal number of components determined by the Broken Stick and the Kaiser rules. The Broken Stick rule suggests that the reduction to only 16 PCs is reasonable.

Finally, we compared these two criteria of PC selection with the criterion: the ratio of the maximal and minimal retained eigenvalues ($\lambda_{max}/\lambda_{min}$) should not exceed the selected number (the condition number) \cite{fukunaga,gorban,mirkes}. It is described as the \textit{multicollinearity control}. In order to avoid the effects of multicollinearity, the conditional number of the covariance matrix after deleting the minor components should not be too large. That is, $k$ is the number of components to be retained if $k$ is the largest number for which $\lambda_{1}/\lambda_{k}<C$, where $C$ is the conditional number. This method is called PCA-CN \cite{mirkes}. In our work, modest collinearity is defined using collinearity with $C<10$ as in \cite{gorban}. Therefore, the number of PCs to be retained is 13 by PCA-CN. Table \ref{table:3methods} shows the number of PCs to be retained (informative PCs) found by three approaches: Kaiser rule, Broken Stick rule and PCA-CN.

\begin{table}[bt]
	\centering
	\caption{Number of PCs to be retained, found by Kaiser rule, Broken Stick rule and PCA-CN}
	\renewcommand\arraystretch{1.3}
\begin{tabular}{|c|c|}
\hline
\textbf{Method} &\textbf{Number of PCs } \\\hline
\textbf{Kaiser} &61 \\\hline
\textbf{Broken Stick}& 16 \\\hline
\textbf{PCA-CN} & 13 \\\hline

\end{tabular}
\label{table:3methods}
   \end{table}

Increasing the number of dimensions could lead to even the closest neighbours being too far away in the Meaning Space, especially for a small number of words. This notion is closely related to the problem of curse of dimensionality. To avoid this isolation of word vectors in the space because of over-fitting, we have decided to use this new 13-dimensional meaning space in the future study of the quantification of meaning of text. The full list of categories in positive and negative groups for each of the first 13 PCs can be found in Appendix \ref{appndx}.

\section{Conclusion and Discussion}\label{conc}
In this paper we compute the dimension of meaning; our answer is 13. We also suggest qualitative meanings for the first three of these dimensions:
\begin{itemize}
\item PC1 describes how well 252 categories from the Web of Science are described by the words in texts classified as being in these categories. The first PC coordinate of a word is its general informational value for separating of categories;
\item PC2 is the Nature of Science dimension, which categorises topics as either discourse or experiment; The corresponding PC coordinate reflects the difference in the use of the word in these two groups of topics;
\item PC3 is the Human Scale dimension, which distinguishes biological science (with some attached disciplines) and medicine (also with some attached disciplines); 
\item PC4 is the Human Condition dimension, where the distinction is between understanding the human through psychology, social and behavioral sciences, or  art (with some admixture of medicine);
\item PC5 is the  Inner World/Outer World  dimension where the human experience of itself (a combination of psychology and computers) is contrasted with the experience of the external world  (environment, ecology and related topics).
\end{itemize}
We welcome fierce debate over the meaning of these dimensions, but giving a qualitative meaning to these is a crucial step to understanding the meaning of meaning.

We arrived at these conclusions by first of all arguing that the reduced word set of 5000 words in LScT reasonably represent the texts from our Web of Science corpus; see Section~\ref{stats}. Having done this it reasonable to perform a principal component analysis of the word-category matrix, whose entries are the relative information gains for the category with the given word; see Section~\ref{dim}. By exploring three different selection criteria (Double Kaiser, Broken Stick, PCA-CN, Section~\ref{dimspace}) we reduced the dimensionality of the category space to 61, 16 and 13 respectively. Meaning cannot be so complicated - so we choose the lowest dimensionality. If it turns out that we cannot explain some component of meaning at some time in the future with only 13 dimensions, we can increase the dimension. It remains a challenge to describe all 13 such dimensions in a way that makes some philosophical sense, but we hope that we have opened up this debate in this paper.

In the future we hope to explore a new categorisation of texts using the positive, negative and zero correlation of each subject category with the influential PCs. Even if we use the first five PCs (with some ternary division on PC1) this would give $3^5=243$ categories (eerily close to the 252 categories in WoS).

\section*{Appendices}
\appendix

\section{The most Informative 100 words of the LScT in categories with their RIGs} \label{InfCat}

 \clearpage 

\newpage 

\thispagestyle{plain} 

\vspace*{-1.6in}

\setlength{\LTleft}{-20cm plus -1fill}

\setlength{\LTright}{\LTleft} 

\begin{longtable}[h]{| C{.6cm}| L{5cm} |  R{1.6cm}  |  R{.3cm} |  C{.6cm}| L{5cm} |  R{1.6cm} | }

\caption{Acoustics}\\ 

\hhline{|-|-|-|~|-|-|-|}

\multicolumn{1}{|C{.6cm}|}{\textbf{No.}} & \multicolumn{1}{C{5cm}|}{\textbf{Word}} & \multicolumn{1}{C{1.6cm}|}{\textbf{RIG}} &   &\multicolumn{1}{C{.6cm}|}{\textbf{No.}} & \multicolumn{1}{C{5cm}|}{\textbf{Word}} & \multicolumn{1}{C{1.6cm}|}{\textbf{RIG}} \\ \hhline{|-|-|-|~|-|-|-|} 

\endhead 

\hhline{|-|-|-|~|-|-|-|}

\endfoot
 1 & acoust & $ 9.4\times10^{-2} $ &   & 51 & recognit & $ 5.9\times10^{-3} $ \\ \hhline{|-|-|-|~|-|-|-|} 
 
 2 & ultrasound & $ 6.2\times10^{-2} $ &   & 52 & activ & $ 5.7\times10^{-3} $ \\ \hhline{|-|-|-|~|-|-|-|} 
 
 3 & sound & $ 4.1\times10^{-2} $ &   & 53 & modal & $ 5.6\times10^{-3} $ \\ \hhline{|-|-|-|~|-|-|-|} 
 
 4 & speech & $ 4.1\times10^{-2} $ &   & 54 & gaussian & $ 5.5\times10^{-3} $ \\ \hhline{|-|-|-|~|-|-|-|} 
 
 5 & frequenc & $ 4\times10^{-2} $ &   & 55 & simul & $ 5.4\times10^{-3} $ \\ \hhline{|-|-|-|~|-|-|-|} 
 
 6 & nois & $ 3.3\times10^{-2} $ &   & 56 & accuraci & $ 5.3\times10^{-3} $ \\ \hhline{|-|-|-|~|-|-|-|} 
 
 7 & ultrason & $ 2.7\times10^{-2} $ &   & 57 & model & $ 5.2\times10^{-3} $ \\ \hhline{|-|-|-|~|-|-|-|} 
 
 8 & wave & $ 2.4\times10^{-2} $ &   & 58 & experiment & $ 5\times10^{-3} $ \\
\end{longtable}
 \newpage 
 
 \clearpage 

\newpage 

\thispagestyle{plain} 

\vspace*{-1.6in}

\setlength{\LTleft}{-20cm plus -1fill}

\setlength{\LTright}{\LTleft} 

\begin{longtable}[h]{| C{.6cm}| L{5cm} |  R{1.6cm}  |  R{.3cm} |  C{.6cm}| L{5cm} |  R{1.6cm} | }

\caption{Agricultural Economics and Policy}\\ 

\hhline{|-|-|-|~|-|-|-|}

\multicolumn{1}{|C{.6cm}|}{\textbf{No.}} & \multicolumn{1}{C{5cm}|}{\textbf{Word}} & \multicolumn{1}{C{1.6cm}|}{\textbf{RIG}} &   &\multicolumn{1}{C{.6cm}|}{\textbf{No.}} & \multicolumn{1}{C{5cm}|}{\textbf{Word}} & \multicolumn{1}{C{1.6cm}|}{\textbf{RIG}} \\ \hhline{|-|-|-|~|-|-|-|} 

\endhead 

\hhline{|-|-|-|~|-|-|-|}

\endfoot
 1 & agricultur & $ 1.1\times10^{-1} $ &   & 51 & choic & $ 9.3\times10^{-3} $ \\ \hhline{|-|-|-|~|-|-|-|} 
 
 2 & farmer & $ 8.7\times10^{-2} $ &   & 52 & competit & $ 9.3\times10^{-3} $ \\ \hhline{|-|-|-|~|-|-|-|} 
 
 3 & market & $ 8.6\times10^{-2} $ &   & 53 & domest & $ 9.1\times10^{-3} $ \\ \hhline{|-|-|-|~|-|-|-|} 
 
 4 & farm & $ 8.3\times10^{-2} $ &   & 54 & conclus & $ 8.8\times10^{-3} $ \\ \hhline{|-|-|-|~|-|-|-|} 
 
 5 & price & $ 8.1\times10^{-2} $ &   & 55 & clinic & $ 8.8\times10^{-3} $ \\ \hhline{|-|-|-|~|-|-|-|} 
 
 6 & food & $ 7.2\times10^{-2} $ &   & 56 & nutrit & $ 8.6\times10^{-3} $ \\ \hhline{|-|-|-|~|-|-|-|} 
 
 7 & polici & $ 6\times10^{-2} $ &   & 57 & veget & $ 8.6\times10^{-3} $ \\ \hhline{|-|-|-|~|-|-|-|} 
 
 8 & product & $ 4.8\times10^{-2} $ &   & 58 & african & $ 8.5\times10^{-3} $ \\
\end{longtable}
 \newpage 
 
 \clearpage 

\newpage 

\thispagestyle{plain} 

\vspace*{-1.6in}

\setlength{\LTleft}{-20cm plus -1fill}

\setlength{\LTright}{\LTleft} 

\begin{longtable}[h]{| C{.6cm}| L{5cm} |  R{1.6cm}  |  R{.3cm} |  C{.6cm}| L{5cm} |  R{1.6cm} | }

\caption{Agricultural Engineering}\\ 

\hhline{|-|-|-|~|-|-|-|}

\multicolumn{1}{|C{.6cm}|}{\textbf{No.}} & \multicolumn{1}{C{5cm}|}{\textbf{Word}} & \multicolumn{1}{C{1.6cm}|}{\textbf{RIG}} &   &\multicolumn{1}{C{.6cm}|}{\textbf{No.}} & \multicolumn{1}{C{5cm}|}{\textbf{Word}} & \multicolumn{1}{C{1.6cm}|}{\textbf{RIG}} \\ \hhline{|-|-|-|~|-|-|-|} 

\endhead 

\hhline{|-|-|-|~|-|-|-|}

\endfoot
 1 & biomass & $ 6.8\times10^{-2} $ &   & 51 & corn & $ 1.1\times10^{-2} $ \\ \hhline{|-|-|-|~|-|-|-|} 
 
 2 & product & $ 6\times10^{-2} $ &   & 52 & fuel & $ 1.1\times10^{-2} $ \\ \hhline{|-|-|-|~|-|-|-|} 
 
 3 & yield & $ 4.1\times10^{-2} $ &   & 53 & enzymat & $ 1.1\times10^{-2} $ \\ \hhline{|-|-|-|~|-|-|-|} 
 
 4 & anaerob & $ 3.1\times10^{-2} $ &   & 54 & manur & $ 1\times10^{-2} $ \\ \hhline{|-|-|-|~|-|-|-|} 
 
 5 & content & $ 2.9\times10^{-2} $ &   & 55 & wood & $ 1\times10^{-2} $ \\ \hhline{|-|-|-|~|-|-|-|} 
 
 6 & ferment & $ 2.9\times10^{-2} $ &   & 56 & extract & $ 9.8\times10^{-3} $ \\ \hhline{|-|-|-|~|-|-|-|} 
 
 7 & crop & $ 2.9\times10^{-2} $ &   & 57 & respect & $ 9.8\times10^{-3} $ \\ \hhline{|-|-|-|~|-|-|-|} 
 
 8 & sludg & $ 2.6\times10^{-2} $ &   & 58 & degre & $ 9.7\times10^{-3} $ \\
\end{longtable}
 \newpage 
 
 \clearpage 

\newpage 

\thispagestyle{plain} 

\vspace*{-1.6in}

\setlength{\LTleft}{-20cm plus -1fill}

\setlength{\LTright}{\LTleft} 

\begin{longtable}[h]{| C{.6cm}| L{5cm} |  R{1.6cm}  |  R{.3cm} |  C{.6cm}| L{5cm} |  R{1.6cm} | }

\caption{Agriculture, Dairy and Animal Science}\\ 

\hhline{|-|-|-|~|-|-|-|}

\multicolumn{1}{|C{.6cm}|}{\textbf{No.}} & \multicolumn{1}{C{5cm}|}{\textbf{Word}} & \multicolumn{1}{C{1.6cm}|}{\textbf{RIG}} &   &\multicolumn{1}{C{.6cm}|}{\textbf{No.}} & \multicolumn{1}{C{5cm}|}{\textbf{Word}} & \multicolumn{1}{C{1.6cm}|}{\textbf{RIG}} \\ \hhline{|-|-|-|~|-|-|-|} 

\endhead 

\hhline{|-|-|-|~|-|-|-|}

\endfoot
 1 & feed & $ 1.1\times10^{-1} $ &   & 51 & period & $ 2.2\times10^{-2} $ \\ \hhline{|-|-|-|~|-|-|-|} 
 
 2 & diet & $ 1.1\times10^{-1} $ &   & 52 & affect & $ 2.2\times10^{-2} $ \\ \hhline{|-|-|-|~|-|-|-|} 
 
 3 & cow & $ 1\times10^{-1} $ &   & 53 & effect & $ 2.1\times10^{-2} $ \\ \hhline{|-|-|-|~|-|-|-|} 
 
 4 & milk & $ 9.4\times10^{-2} $ &   & 54 & bodi & $ 2\times10^{-2} $ \\ \hhline{|-|-|-|~|-|-|-|} 
 
 5 & fed & $ 9.1\times10^{-2} $ &   & 55 & acid & $ 2\times10^{-2} $ \\ \hhline{|-|-|-|~|-|-|-|} 
 
 6 & dairi & $ 8.9\times10^{-2} $ &   & 56 & crude & $ 2\times10^{-2} $ \\ \hhline{|-|-|-|~|-|-|-|} 
 
 7 & anim & $ 7.1\times10^{-2} $ &   & 57 & studi & $ 2\times10^{-2} $ \\ \hhline{|-|-|-|~|-|-|-|} 
 
 8 & intak & $ 6.9\times10^{-2} $ &   & 58 & content & $ 1.9\times10^{-2} $ \\
\end{longtable}
 \newpage 
 
 \clearpage 

\newpage 

\thispagestyle{plain} 

\vspace*{-1.6in}

\setlength{\LTleft}{-20cm plus -1fill}

\setlength{\LTright}{\LTleft} 

\begin{longtable}[h]{| C{.6cm}| L{5cm} |  R{1.6cm}  |  R{.3cm} |  C{.6cm}| L{5cm} |  R{1.6cm} | }

\caption{Agriculture, Multidisciplinary}\\ 

\hhline{|-|-|-|~|-|-|-|}

\multicolumn{1}{|C{.6cm}|}{\textbf{No.}} & \multicolumn{1}{C{5cm}|}{\textbf{Word}} & \multicolumn{1}{C{1.6cm}|}{\textbf{RIG}} &   &\multicolumn{1}{C{.6cm}|}{\textbf{No.}} & \multicolumn{1}{C{5cm}|}{\textbf{Word}} & \multicolumn{1}{C{1.6cm}|}{\textbf{RIG}} \\ \hhline{|-|-|-|~|-|-|-|} 

\endhead 

\hhline{|-|-|-|~|-|-|-|}

\endfoot
 1 & crop & $ 6\times10^{-2} $ &   & 51 & chromatographi & $ 8.4\times10^{-3} $ \\ \hhline{|-|-|-|~|-|-|-|} 
 
 2 & plant & $ 4.4\times10^{-2} $ &   & 52 & agronom & $ 8.4\times10^{-3} $ \\ \hhline{|-|-|-|~|-|-|-|} 
 
 3 & agricultur & $ 4\times10^{-2} $ &   & 53 & paper & $ 8.3\times10^{-3} $ \\ \hhline{|-|-|-|~|-|-|-|} 
 
 4 & soil & $ 3.5\times10^{-2} $ &   & 54 & day & $ 8.3\times10^{-3} $ \\ \hhline{|-|-|-|~|-|-|-|} 
 
 5 & product & $ 3.1\times10^{-2} $ &   & 55 & weed & $ 8.3\times10^{-3} $ \\ \hhline{|-|-|-|~|-|-|-|} 
 
 6 & farm & $ 2.9\times10^{-2} $ &   & 56 & matter & $ 8.2\times10^{-3} $ \\ \hhline{|-|-|-|~|-|-|-|} 
 
 7 & wheat & $ 2.4\times10^{-2} $ &   & 57 & diet & $ 8.2\times10^{-3} $ \\ \hhline{|-|-|-|~|-|-|-|} 
 
 8 & cultivar & $ 2.3\times10^{-2} $ &   & 58 & digest & $ 8.2\times10^{-3} $ \\
\end{longtable}
 \newpage 
 
 \clearpage 

\newpage 

\thispagestyle{plain} 

\vspace*{-1.6in}

\setlength{\LTleft}{-20cm plus -1fill}

\setlength{\LTright}{\LTleft} 

\begin{longtable}[h]{| C{.6cm}| L{5cm} |  R{1.6cm}  |  R{.3cm} |  C{.6cm}| L{5cm} |  R{1.6cm} | }

\caption{Agronomy}\\ 

\hhline{|-|-|-|~|-|-|-|}

\multicolumn{1}{|C{.6cm}|}{\textbf{No.}} & \multicolumn{1}{C{5cm}|}{\textbf{Word}} & \multicolumn{1}{C{1.6cm}|}{\textbf{RIG}} &   &\multicolumn{1}{C{.6cm}|}{\textbf{No.}} & \multicolumn{1}{C{5cm}|}{\textbf{Word}} & \multicolumn{1}{C{1.6cm}|}{\textbf{RIG}} \\ \hhline{|-|-|-|~|-|-|-|} 

\endhead 

\hhline{|-|-|-|~|-|-|-|}

\endfoot
 1 & plant & $ 1.3\times10^{-1} $ &   & 51 & winter & $ 1.4\times10^{-2} $ \\ \hhline{|-|-|-|~|-|-|-|} 
 
 2 & crop & $ 1.2\times10^{-1} $ &   & 52 & plot & $ 1.4\times10^{-2} $ \\ \hhline{|-|-|-|~|-|-|-|} 
 
 3 & cultivar & $ 9.3\times10^{-2} $ &   & 53 & total & $ 1.4\times10^{-2} $ \\ \hhline{|-|-|-|~|-|-|-|} 
 
 4 & soil & $ 7.9\times10^{-2} $ &   & 54 & farmer & $ 1.4\times10^{-2} $ \\ \hhline{|-|-|-|~|-|-|-|} 
 
 5 & yield & $ 6\times10^{-2} $ &   & 55 & pest & $ 1.4\times10^{-2} $ \\ \hhline{|-|-|-|~|-|-|-|} 
 
 6 & seed & $ 4.5\times10^{-2} $ &   & 56 & potato & $ 1.3\times10^{-2} $ \\ \hhline{|-|-|-|~|-|-|-|} 
 
 7 & leaf & $ 4.4\times10^{-2} $ &   & 57 & tree & $ 1.3\times10^{-2} $ \\ \hhline{|-|-|-|~|-|-|-|} 
 
 8 & breed & $ 4.2\times10^{-2} $ &   & 58 & germin & $ 1.3\times10^{-2} $ \\
\end{longtable}
 \newpage 
 
 \clearpage 

\newpage 

\thispagestyle{plain} 

\vspace*{-1.6in}

\setlength{\LTleft}{-20cm plus -1fill}

\setlength{\LTright}{\LTleft} 

\begin{longtable}[h]{| C{.6cm}| L{5cm} |  R{1.6cm}  |  R{.3cm} |  C{.6cm}| L{5cm} |  R{1.6cm} | }

\caption{Allergy}\\ 

\hhline{|-|-|-|~|-|-|-|}

\multicolumn{1}{|C{.6cm}|}{\textbf{No.}} & \multicolumn{1}{C{5cm}|}{\textbf{Word}} & \multicolumn{1}{C{1.6cm}|}{\textbf{RIG}} &   &\multicolumn{1}{C{.6cm}|}{\textbf{No.}} & \multicolumn{1}{C{5cm}|}{\textbf{Word}} & \multicolumn{1}{C{1.6cm}|}{\textbf{RIG}} \\ \hhline{|-|-|-|~|-|-|-|} 

\endhead 

\hhline{|-|-|-|~|-|-|-|}

\endfoot
 1 & asthma & $ 2.6\times10^{-1} $ &   & 51 & medic & $ 1.7\times10^{-2} $ \\ \hhline{|-|-|-|~|-|-|-|} 
 
 2 & allerg & $ 2.2\times10^{-1} $ &   & 52 & sensit & $ 1.7\times10^{-2} $ \\ \hhline{|-|-|-|~|-|-|-|} 
 
 3 & allergi & $ 1.7\times10^{-1} $ &   & 53 & assess & $ 1.6\times10^{-2} $ \\ \hhline{|-|-|-|~|-|-|-|} 
 
 4 & allergen & $ 1.4\times10^{-1} $ &   & 54 & control & $ 1.6\times10^{-2} $ \\ \hhline{|-|-|-|~|-|-|-|} 
 
 5 & conclus & $ 1.3\times10^{-1} $ &   & 55 & signific & $ 1.6\times10^{-2} $ \\ \hhline{|-|-|-|~|-|-|-|} 
 
 6 & background & $ 1.3\times10^{-1} $ &   & 56 & treatment & $ 1.6\times10^{-2} $ \\ \hhline{|-|-|-|~|-|-|-|} 
 
 7 & ige & $ 1.3\times10^{-1} $ &   & 57 & subject & $ 1.6\times10^{-2} $ \\ \hhline{|-|-|-|~|-|-|-|} 
 
 8 & patient & $ 8\times10^{-2} $ &   & 58 & histori & $ 1.5\times10^{-2} $ \\
\end{longtable}
 \newpage 
 
 \clearpage 

\newpage 

\thispagestyle{plain} 

\vspace*{-1.6in}

\setlength{\LTleft}{-20cm plus -1fill}

\setlength{\LTright}{\LTleft} 

\begin{longtable}[h]{| C{.6cm}| L{5cm} |  R{1.6cm}  |  R{.3cm} |  C{.6cm}| L{5cm} |  R{1.6cm} | }

\caption{Anatomy and Morphology}\\ 

\hhline{|-|-|-|~|-|-|-|}

\multicolumn{1}{|C{.6cm}|}{\textbf{No.}} & \multicolumn{1}{C{5cm}|}{\textbf{Word}} & \multicolumn{1}{C{1.6cm}|}{\textbf{RIG}} &   &\multicolumn{1}{C{.6cm}|}{\textbf{No.}} & \multicolumn{1}{C{5cm}|}{\textbf{Word}} & \multicolumn{1}{C{1.6cm}|}{\textbf{RIG}} \\ \hhline{|-|-|-|~|-|-|-|} 

\endhead 

\hhline{|-|-|-|~|-|-|-|}

\endfoot
 1 & anatom & $ 5.8\times10^{-2} $ &   & 51 & section & $ 9.8\times10^{-3} $ \\ \hhline{|-|-|-|~|-|-|-|} 
 
 2 & anatomi & $ 4.8\times10^{-2} $ &   & 52 & bilater & $ 9.6\times10^{-3} $ \\ \hhline{|-|-|-|~|-|-|-|} 
 
 3 & morpholog & $ 4\times10^{-2} $ &   & 53 & cartilag & $ 9.3\times10^{-3} $ \\ \hhline{|-|-|-|~|-|-|-|} 
 
 4 & cadav & $ 3.8\times10^{-2} $ &   & 54 & trunk & $ 9.2\times10^{-3} $ \\ \hhline{|-|-|-|~|-|-|-|} 
 
 5 & dissect & $ 3.2\times10^{-2} $ &   & 55 & human & $ 8.7\times10^{-3} $ \\ \hhline{|-|-|-|~|-|-|-|} 
 
 6 & posterior & $ 2.6\times10^{-2} $ &   & 56 & cadaver & $ 8.5\times10^{-3} $ \\ \hhline{|-|-|-|~|-|-|-|} 
 
 7 & neuron & $ 2.5\times10^{-2} $ &   & 57 & axon & $ 8.5\times10^{-3} $ \\ \hhline{|-|-|-|~|-|-|-|} 
 
 8 & anterior & $ 2.5\times10^{-2} $ &   & 58 & embryon & $ 8.5\times10^{-3} $ \\
\end{longtable}
 \newpage 
 
 \clearpage 

\newpage 

\thispagestyle{plain} 

\vspace*{-1.6in}

\setlength{\LTleft}{-20cm plus -1fill}

\setlength{\LTright}{\LTleft} 

\begin{longtable}[h]{| C{.6cm}| L{5cm} |  R{1.6cm}  |  R{.3cm} |  C{.6cm}| L{5cm} |  R{1.6cm} | }

\caption{Andrology}\\ 

\hhline{|-|-|-|~|-|-|-|}

\multicolumn{1}{|C{.6cm}|}{\textbf{No.}} & \multicolumn{1}{C{5cm}|}{\textbf{Word}} & \multicolumn{1}{C{1.6cm}|}{\textbf{RIG}} &   &\multicolumn{1}{C{.6cm}|}{\textbf{No.}} & \multicolumn{1}{C{5cm}|}{\textbf{Word}} & \multicolumn{1}{C{1.6cm}|}{\textbf{RIG}} \\ \hhline{|-|-|-|~|-|-|-|} 

\endhead 

\hhline{|-|-|-|~|-|-|-|}

\endfoot
 1 & sperm & $ 2.2\times10^{-1} $ &   & 51 & cryopreserv & $ 1.2\times10^{-2} $ \\ \hhline{|-|-|-|~|-|-|-|} 
 
 2 & infertil & $ 1.2\times10^{-1} $ &   & 52 & evalu & $ 1.2\times10^{-2} $ \\ \hhline{|-|-|-|~|-|-|-|} 
 
 3 & testicular & $ 1.2\times10^{-1} $ &   & 53 & associ & $ 1.2\times10^{-2} $ \\ \hhline{|-|-|-|~|-|-|-|} 
 
 4 & semen & $ 1.1\times10^{-1} $ &   & 54 & inject & $ 1.2\times10^{-2} $ \\ \hhline{|-|-|-|~|-|-|-|} 
 
 5 & spermatozoa & $ 1.1\times10^{-1} $ &   & 55 & malondialdehyd & $ 1.1\times10^{-2} $ \\ \hhline{|-|-|-|~|-|-|-|} 
 
 6 & male & $ 9.9\times10^{-2} $ &   & 56 & control & $ 1.1\times10^{-2} $ \\ \hhline{|-|-|-|~|-|-|-|} 
 
 7 & testi & $ 9.7\times10^{-2} $ &   & 57 & wistar & $ 1\times10^{-2} $ \\ \hhline{|-|-|-|~|-|-|-|} 
 
 8 & motil & $ 8\times10^{-2} $ &   & 58 & glutathion & $ 1\times10^{-2} $ \\
\end{longtable}
 \newpage 
 
 \clearpage 

\newpage 

\thispagestyle{plain} 

\vspace*{-1.6in}

\setlength{\LTleft}{-20cm plus -1fill}

\setlength{\LTright}{\LTleft} 

\begin{longtable}[h]{| C{.6cm}| L{5cm} |  R{1.6cm}  |  R{.3cm} |  C{.6cm}| L{5cm} |  R{1.6cm} | }

\caption{Anesthesiology}\\ 

\hhline{|-|-|-|~|-|-|-|}

\multicolumn{1}{|C{.6cm}|}{\textbf{No.}} & \multicolumn{1}{C{5cm}|}{\textbf{Word}} & \multicolumn{1}{C{1.6cm}|}{\textbf{RIG}} &   &\multicolumn{1}{C{.6cm}|}{\textbf{No.}} & \multicolumn{1}{C{5cm}|}{\textbf{Word}} & \multicolumn{1}{C{1.6cm}|}{\textbf{RIG}} \\ \hhline{|-|-|-|~|-|-|-|} 

\endhead 

\hhline{|-|-|-|~|-|-|-|}

\endfoot
 1 & pain & $ 1.4\times10^{-1} $ &   & 51 & hour & $ 1.8\times10^{-2} $ \\ \hhline{|-|-|-|~|-|-|-|} 
 
 2 & patient & $ 1.2\times10^{-1} $ &   & 52 & trial & $ 1.7\times10^{-2} $ \\ \hhline{|-|-|-|~|-|-|-|} 
 
 3 & anesthesia & $ 9.7\times10^{-2} $ &   & 53 & intraven & $ 1.7\times10^{-2} $ \\ \hhline{|-|-|-|~|-|-|-|} 
 
 4 & conclus & $ 9\times10^{-2} $ &   & 54 & incid & $ 1.7\times10^{-2} $ \\ \hhline{|-|-|-|~|-|-|-|} 
 
 5 & surgeri & $ 8.4\times10^{-2} $ &   & 55 & median & $ 1.7\times10^{-2} $ \\ \hhline{|-|-|-|~|-|-|-|} 
 
 6 & background & $ 8\times10^{-2} $ &   & 56 & blood & $ 1.7\times10^{-2} $ \\ \hhline{|-|-|-|~|-|-|-|} 
 
 7 & anesthet & $ 6\times10^{-2} $ &   & 57 & dose & $ 1.7\times10^{-2} $ \\ \hhline{|-|-|-|~|-|-|-|} 
 
 8 & postop & $ 5.6\times10^{-2} $ &   & 58 & complic & $ 1.7\times10^{-2} $ \\
\end{longtable}
 \newpage 
 
 \clearpage 

\newpage 

\thispagestyle{plain} 

\vspace*{-1.6in}

\setlength{\LTleft}{-20cm plus -1fill}

\setlength{\LTright}{\LTleft} 

\begin{longtable}[h]{| C{.6cm}| L{5cm} |  R{1.6cm}  |  R{.3cm} |  C{.6cm}| L{5cm} |  R{1.6cm} | }

\caption{Anthropology}\\ 

\hhline{|-|-|-|~|-|-|-|}

\multicolumn{1}{|C{.6cm}|}{\textbf{No.}} & \multicolumn{1}{C{5cm}|}{\textbf{Word}} & \multicolumn{1}{C{1.6cm}|}{\textbf{RIG}} &   &\multicolumn{1}{C{.6cm}|}{\textbf{No.}} & \multicolumn{1}{C{5cm}|}{\textbf{Word}} & \multicolumn{1}{C{1.6cm}|}{\textbf{RIG}} \\ \hhline{|-|-|-|~|-|-|-|} 

\endhead 

\hhline{|-|-|-|~|-|-|-|}

\endfoot
 1 & archaeolog & $ 8.7\times10^{-2} $ &   & 51 & chronolog & $ 9.6\times10^{-3} $ \\ \hhline{|-|-|-|~|-|-|-|} 
 
 2 & anthropolog & $ 4.7\times10^{-2} $ &   & 52 & individu & $ 9.5\times10^{-3} $ \\ \hhline{|-|-|-|~|-|-|-|} 
 
 3 & ethnograph & $ 4.5\times10^{-2} $ &   & 53 & world & $ 9.1\times10^{-3} $ \\ \hhline{|-|-|-|~|-|-|-|} 
 
 4 & social & $ 4.3\times10^{-2} $ &   & 54 & earli & $ 9.1\times10^{-3} $ \\ \hhline{|-|-|-|~|-|-|-|} 
 
 5 & polit & $ 3.1\times10^{-2} $ &   & 55 & debat & $ 9\times10^{-3} $ \\ \hhline{|-|-|-|~|-|-|-|} 
 
 6 & cultur & $ 3.1\times10^{-2} $ &   & 56 & effect & $ 9\times10^{-3} $ \\ \hhline{|-|-|-|~|-|-|-|} 
 
 7 & argu & $ 2.7\times10^{-2} $ &   & 57 & interpret & $ 9\times10^{-3} $ \\ \hhline{|-|-|-|~|-|-|-|} 
 
 8 & articl & $ 2.4\times10^{-2} $ &   & 58 & pleistocen & $ 9\times10^{-3} $ \\
\end{longtable}
 \newpage 
 
 \clearpage 

\newpage 

\thispagestyle{plain} 

\vspace*{-1.6in}

\setlength{\LTleft}{-20cm plus -1fill}

\setlength{\LTright}{\LTleft} 

\begin{longtable}[h]{| C{.6cm}| L{5cm} |  R{1.6cm}  |  R{.3cm} |  C{.6cm}| L{5cm} |  R{1.6cm} | }

\caption{Archaeology}\\ 

\hhline{|-|-|-|~|-|-|-|}

\multicolumn{1}{|C{.6cm}|}{\textbf{No.}} & \multicolumn{1}{C{5cm}|}{\textbf{Word}} & \multicolumn{1}{C{1.6cm}|}{\textbf{RIG}} &   &\multicolumn{1}{C{.6cm}|}{\textbf{No.}} & \multicolumn{1}{C{5cm}|}{\textbf{Word}} & \multicolumn{1}{C{1.6cm}|}{\textbf{RIG}} \\ \hhline{|-|-|-|~|-|-|-|} 

\endhead 

\hhline{|-|-|-|~|-|-|-|}

\endfoot
 1 & archaeolog & $ 2.9\times10^{-1} $ &   & 51 & eastern & $ 1.3\times10^{-2} $ \\ \hhline{|-|-|-|~|-|-|-|} 
 
 2 & excav & $ 8.9\times10^{-2} $ &   & 52 & cave & $ 1.3\times10^{-2} $ \\ \hhline{|-|-|-|~|-|-|-|} 
 
 3 & site & $ 7.6\times10^{-2} $ &   & 53 & paint & $ 1.3\times10^{-2} $ \\ \hhline{|-|-|-|~|-|-|-|} 
 
 4 & centuri & $ 7.2\times10^{-2} $ &   & 54 & materi & $ 1.2\times10^{-2} $ \\ \hhline{|-|-|-|~|-|-|-|} 
 
 5 & neolith & $ 5.1\times10^{-2} $ &   & 55 & past & $ 1.2\times10^{-2} $ \\ \hhline{|-|-|-|~|-|-|-|} 
 
 6 & archaeologist & $ 4.9\times10^{-2} $ &   & 56 & argu & $ 1.2\times10^{-2} $ \\ \hhline{|-|-|-|~|-|-|-|} 
 
 7 & date & $ 4.9\times10^{-2} $ &   & 57 & domest & $ 1.2\times10^{-2} $ \\ \hhline{|-|-|-|~|-|-|-|} 
 
 8 & bronz & $ 4.8\times10^{-2} $ &   & 58 & ceram & $ 1.2\times10^{-2} $ \\
\end{longtable}
 \newpage 
 
 \clearpage 

\newpage 

\thispagestyle{plain} 

\vspace*{-1.6in}

\setlength{\LTleft}{-20cm plus -1fill}

\setlength{\LTright}{\LTleft} 

\begin{longtable}[h]{| C{.6cm}| L{5cm} |  R{1.6cm}  |  R{.3cm} |  C{.6cm}| L{5cm} |  R{1.6cm} | }

\caption{Architecture}\\ 

\hhline{|-|-|-|~|-|-|-|}

\multicolumn{1}{|C{.6cm}|}{\textbf{No.}} & \multicolumn{1}{C{5cm}|}{\textbf{Word}} & \multicolumn{1}{C{1.6cm}|}{\textbf{RIG}} &   &\multicolumn{1}{C{.6cm}|}{\textbf{No.}} & \multicolumn{1}{C{5cm}|}{\textbf{Word}} & \multicolumn{1}{C{1.6cm}|}{\textbf{RIG}} \\ \hhline{|-|-|-|~|-|-|-|} 

\endhead 

\hhline{|-|-|-|~|-|-|-|}

\endfoot
 1 & architectur & $ 1.3\times10^{-1} $ &   & 51 & church & $ 8.5\times10^{-3} $ \\ \hhline{|-|-|-|~|-|-|-|} 
 
 2 & build & $ 9.6\times10^{-2} $ &   & 52 & studio & $ 8.3\times10^{-3} $ \\ \hhline{|-|-|-|~|-|-|-|} 
 
 3 & architect & $ 6.5\times10^{-2} $ &   & 53 & protein & $ 8.3\times10^{-3} $ \\ \hhline{|-|-|-|~|-|-|-|} 
 
 4 & heritag & $ 6.2\times10^{-2} $ &   & 54 & context & $ 8.2\times10^{-3} $ \\ \hhline{|-|-|-|~|-|-|-|} 
 
 5 & urban & $ 4.8\times10^{-2} $ &   & 55 & place & $ 8.2\times10^{-3} $ \\ \hhline{|-|-|-|~|-|-|-|} 
 
 6 & vernacular & $ 4.5\times10^{-2} $ &   & 56 & draw & $ 8.2\times10^{-3} $ \\ \hhline{|-|-|-|~|-|-|-|} 
 
 7 & hous & $ 4.4\times10^{-2} $ &   & 57 & focus & $ 8\times10^{-3} $ \\ \hhline{|-|-|-|~|-|-|-|} 
 
 8 & centuri & $ 4.2\times10^{-2} $ &   & 58 & develop & $ 8\times10^{-3} $ \\
\end{longtable}
 \newpage 
 
 \clearpage 

\newpage 

\thispagestyle{plain} 

\vspace*{-1.6in}

\setlength{\LTleft}{-20cm plus -1fill}

\setlength{\LTright}{\LTleft} 

\begin{longtable}[h]{| C{.6cm}| L{5cm} |  R{1.6cm}  |  R{.3cm} |  C{.6cm}| L{5cm} |  R{1.6cm} | }

\caption{Area Studies}\\ 

\hhline{|-|-|-|~|-|-|-|}

\multicolumn{1}{|C{.6cm}|}{\textbf{No.}} & \multicolumn{1}{C{5cm}|}{\textbf{Word}} & \multicolumn{1}{C{1.6cm}|}{\textbf{RIG}} &   &\multicolumn{1}{C{.6cm}|}{\textbf{No.}} & \multicolumn{1}{C{5cm}|}{\textbf{Word}} & \multicolumn{1}{C{1.6cm}|}{\textbf{RIG}} \\ \hhline{|-|-|-|~|-|-|-|} 

\endhead 

\hhline{|-|-|-|~|-|-|-|}

\endfoot
 1 & polit & $ 1.5\times10^{-1} $ &   & 51 & elect & $ 1.5\times10^{-2} $ \\ \hhline{|-|-|-|~|-|-|-|} 
 
 2 & articl & $ 1.1\times10^{-1} $ &   & 52 & peopl & $ 1.5\times10^{-2} $ \\ \hhline{|-|-|-|~|-|-|-|} 
 
 3 & argu & $ 6.5\times10^{-2} $ &   & 53 & patient & $ 1.4\times10^{-2} $ \\ \hhline{|-|-|-|~|-|-|-|} 
 
 4 & govern & $ 5.1\times10^{-2} $ &   & 54 & militari & $ 1.4\times10^{-2} $ \\ \hhline{|-|-|-|~|-|-|-|} 
 
 5 & polici & $ 4.4\times10^{-2} $ &   & 55 & south & $ 1.4\times10^{-2} $ \\ \hhline{|-|-|-|~|-|-|-|} 
 
 6 & result & $ 3.7\times10^{-2} $ &   & 56 & protest & $ 1.4\times10^{-2} $ \\ \hhline{|-|-|-|~|-|-|-|} 
 
 7 & social & $ 3.7\times10^{-2} $ &   & 57 & religi & $ 1.4\times10^{-2} $ \\ \hhline{|-|-|-|~|-|-|-|} 
 
 8 & nation & $ 3.5\times10^{-2} $ &   & 58 & union & $ 1.4\times10^{-2} $ \\
\end{longtable}
 \newpage 
 
 \clearpage 

\newpage 

\thispagestyle{plain} 

\vspace*{-1.6in}

\setlength{\LTleft}{-20cm plus -1fill}

\setlength{\LTright}{\LTleft} 

\begin{longtable}[h]{| C{.6cm}| L{5cm} |  R{1.6cm}  |  R{.3cm} |  C{.6cm}| L{5cm} |  R{1.6cm} | }

\caption{Art}\\ 

\hhline{|-|-|-|~|-|-|-|}

\multicolumn{1}{|C{.6cm}|}{\textbf{No.}} & \multicolumn{1}{C{5cm}|}{\textbf{Word}} & \multicolumn{1}{C{1.6cm}|}{\textbf{RIG}} &   &\multicolumn{1}{C{.6cm}|}{\textbf{No.}} & \multicolumn{1}{C{5cm}|}{\textbf{Word}} & \multicolumn{1}{C{1.6cm}|}{\textbf{RIG}} \\ \hhline{|-|-|-|~|-|-|-|} 

\endhead 

\hhline{|-|-|-|~|-|-|-|}

\endfoot
 1 & artist & $ 1.2\times10^{-1} $ &   & 51 & creat & $ 1\times10^{-2} $ \\ \hhline{|-|-|-|~|-|-|-|} 
 
 2 & art & $ 1.1\times10^{-1} $ &   & 52 & interpret & $ 9.4\times10^{-3} $ \\ \hhline{|-|-|-|~|-|-|-|} 
 
 3 & centuri & $ 6.1\times10^{-2} $ &   & 53 & question & $ 9.4\times10^{-3} $ \\ \hhline{|-|-|-|~|-|-|-|} 
 
 4 & paint & $ 6\times10^{-2} $ &   & 54 & control & $ 9.3\times10^{-3} $ \\ \hhline{|-|-|-|~|-|-|-|} 
 
 5 & articl & $ 5.4\times10^{-2} $ &   & 55 & rate & $ 9.3\times10^{-3} $ \\ \hhline{|-|-|-|~|-|-|-|} 
 
 6 & aesthet & $ 4.4\times10^{-2} $ &   & 56 & decreas & $ 8.8\times10^{-3} $ \\ \hhline{|-|-|-|~|-|-|-|} 
 
 7 & artwork & $ 4.3\times10^{-2} $ &   & 57 & embodi & $ 8.5\times10^{-3} $ \\ \hhline{|-|-|-|~|-|-|-|} 
 
 8 & cultur & $ 3.4\times10^{-2} $ &   & 58 & modernist & $ 8.4\times10^{-3} $ \\
\end{longtable}
 \newpage 
 
 \clearpage 

\newpage 

\thispagestyle{plain} 

\vspace*{-1.6in}

\setlength{\LTleft}{-20cm plus -1fill}

\setlength{\LTright}{\LTleft} 

\begin{longtable}[h]{| C{.6cm}| L{5cm} |  R{1.6cm}  |  R{.3cm} |  C{.6cm}| L{5cm} |  R{1.6cm} | }

\caption{Asian Studies}\\ 

\hhline{|-|-|-|~|-|-|-|}

\multicolumn{1}{|C{.6cm}|}{\textbf{No.}} & \multicolumn{1}{C{5cm}|}{\textbf{Word}} & \multicolumn{1}{C{1.6cm}|}{\textbf{RIG}} &   &\multicolumn{1}{C{.6cm}|}{\textbf{No.}} & \multicolumn{1}{C{5cm}|}{\textbf{Word}} & \multicolumn{1}{C{1.6cm}|}{\textbf{RIG}} \\ \hhline{|-|-|-|~|-|-|-|} 

\endhead 

\hhline{|-|-|-|~|-|-|-|}

\endfoot
 1 & articl & $ 9.8\times10^{-2} $ &   & 51 & examin & $ 1.4\times10^{-2} $ \\ \hhline{|-|-|-|~|-|-|-|} 
 
 2 & argu & $ 6.5\times10^{-2} $ &   & 52 & china & $ 1.4\times10^{-2} $ \\ \hhline{|-|-|-|~|-|-|-|} 
 
 3 & polit & $ 6.3\times10^{-2} $ &   & 53 & india & $ 1.4\times10^{-2} $ \\ \hhline{|-|-|-|~|-|-|-|} 
 
 4 & centuri & $ 6.2\times10^{-2} $ &   & 54 & offici & $ 1.4\times10^{-2} $ \\ \hhline{|-|-|-|~|-|-|-|} 
 
 5 & buddhist & $ 5.2\times10^{-2} $ &   & 55 & christian & $ 1.4\times10^{-2} $ \\ \hhline{|-|-|-|~|-|-|-|} 
 
 6 & result & $ 4.5\times10^{-2} $ &   & 56 & cell & $ 1.3\times10^{-2} $ \\ \hhline{|-|-|-|~|-|-|-|} 
 
 7 & religi & $ 4\times10^{-2} $ &   & 57 & patient & $ 1.3\times10^{-2} $ \\ \hhline{|-|-|-|~|-|-|-|} 
 
 8 & text & $ 3.7\times10^{-2} $ &   & 58 & test & $ 1.3\times10^{-2} $ \\
\end{longtable}
 \newpage 
 
 \clearpage 

\newpage 

\thispagestyle{plain} 

\vspace*{-1.6in}

\setlength{\LTleft}{-20cm plus -1fill}

\setlength{\LTright}{\LTleft} 

\begin{longtable}[h]{| C{.6cm}| L{5cm} |  R{1.6cm}  |  R{.3cm} |  C{.6cm}| L{5cm} |  R{1.6cm} | }

\caption{Astronomy and Astrophysics}\\ 

\hhline{|-|-|-|~|-|-|-|}

\multicolumn{1}{|C{.6cm}|}{\textbf{No.}} & \multicolumn{1}{C{5cm}|}{\textbf{Word}} & \multicolumn{1}{C{1.6cm}|}{\textbf{RIG}} &   &\multicolumn{1}{C{.6cm}|}{\textbf{No.}} & \multicolumn{1}{C{5cm}|}{\textbf{Word}} & \multicolumn{1}{C{1.6cm}|}{\textbf{RIG}} \\ \hhline{|-|-|-|~|-|-|-|} 

\endhead 

\hhline{|-|-|-|~|-|-|-|}

\endfoot
 1 & star & $ 1.5\times10^{-1} $ &   & 51 & dot & $ 1.6\times10^{-2} $ \\ \hhline{|-|-|-|~|-|-|-|} 
 
 2 & galaxi & $ 1.1\times10^{-1} $ &   & 52 & find & $ 1.6\times10^{-2} $ \\ \hhline{|-|-|-|~|-|-|-|} 
 
 3 & telescop & $ 9.9\times10^{-2} $ &   & 53 & perturb & $ 1.6\times10^{-2} $ \\ \hhline{|-|-|-|~|-|-|-|} 
 
 4 & stellar & $ 9.8\times10^{-2} $ &   & 54 & radiat & $ 1.6\times10^{-2} $ \\ \hhline{|-|-|-|~|-|-|-|} 
 
 5 & galact & $ 6.6\times10^{-2} $ &   & 55 & spacetim & $ 1.6\times10^{-2} $ \\ \hhline{|-|-|-|~|-|-|-|} 
 
 6 & mass & $ 5.9\times10^{-2} $ &   & 56 & magnet & $ 1.5\times10^{-2} $ \\ \hhline{|-|-|-|~|-|-|-|} 
 
 7 & cosmolog & $ 5.4\times10^{-2} $ &   & 57 & densiti & $ 1.5\times10^{-2} $ \\ \hhline{|-|-|-|~|-|-|-|} 
 
 8 & luminos & $ 5.1\times10^{-2} $ &   & 58 & radial & $ 1.5\times10^{-2} $ \\
\end{longtable}
 \newpage 
 
 \clearpage 

\newpage 

\thispagestyle{plain} 

\vspace*{-1.6in}

\setlength{\LTleft}{-20cm plus -1fill}

\setlength{\LTright}{\LTleft} 

\begin{longtable}[h]{| C{.6cm}| L{5cm} |  R{1.6cm}  |  R{.3cm} |  C{.6cm}| L{5cm} |  R{1.6cm} | }

\caption{Audiology and Speech-Language Pathology}\\ 

\hhline{|-|-|-|~|-|-|-|}

\multicolumn{1}{|C{.6cm}|}{\textbf{No.}} & \multicolumn{1}{C{5cm}|}{\textbf{Word}} & \multicolumn{1}{C{1.6cm}|}{\textbf{RIG}} &   &\multicolumn{1}{C{.6cm}|}{\textbf{No.}} & \multicolumn{1}{C{5cm}|}{\textbf{Word}} & \multicolumn{1}{C{1.6cm}|}{\textbf{RIG}} \\ \hhline{|-|-|-|~|-|-|-|} 

\endhead 

\hhline{|-|-|-|~|-|-|-|}

\endfoot
 1 & speech & $ 1.4\times10^{-1} $ &   & 51 & stimulus & $ 9.5\times10^{-3} $ \\ \hhline{|-|-|-|~|-|-|-|} 
 
 2 & hear & $ 1.3\times10^{-1} $ &   & 52 & pitch & $ 9.2\times10^{-3} $ \\ \hhline{|-|-|-|~|-|-|-|} 
 
 3 & acoust & $ 8.9\times10^{-2} $ &   & 53 & intellig & $ 9.2\times10^{-3} $ \\ \hhline{|-|-|-|~|-|-|-|} 
 
 4 & listen & $ 7.9\times10^{-2} $ &   & 54 & result & $ 9\times10^{-3} $ \\ \hhline{|-|-|-|~|-|-|-|} 
 
 5 & sound & $ 7.2\times10^{-2} $ &   & 55 & skill & $ 8.9\times10^{-3} $ \\ \hhline{|-|-|-|~|-|-|-|} 
 
 6 & auditori & $ 6\times10^{-2} $ &   & 56 & record & $ 8.9\times10^{-3} $ \\ \hhline{|-|-|-|~|-|-|-|} 
 
 7 & languag & $ 5.8\times10^{-2} $ &   & 57 & examin & $ 8.9\times10^{-3} $ \\ \hhline{|-|-|-|~|-|-|-|} 
 
 8 & cochlear & $ 5.2\times10^{-2} $ &   & 58 & tempor & $ 8.8\times10^{-3} $ \\
\end{longtable}
 \newpage 
 
 \clearpage 

\newpage 

\thispagestyle{plain} 

\vspace*{-1.6in}

\setlength{\LTleft}{-20cm plus -1fill}

\setlength{\LTright}{\LTleft} 

\begin{longtable}[h]{| C{.6cm}| L{5cm} |  R{1.6cm}  |  R{.3cm} |  C{.6cm}| L{5cm} |  R{1.6cm} | }

\caption{Automation and Control Systems}\\ 

\hhline{|-|-|-|~|-|-|-|}

\multicolumn{1}{|C{.6cm}|}{\textbf{No.}} & \multicolumn{1}{C{5cm}|}{\textbf{Word}} & \multicolumn{1}{C{1.6cm}|}{\textbf{RIG}} &   &\multicolumn{1}{C{.6cm}|}{\textbf{No.}} & \multicolumn{1}{C{5cm}|}{\textbf{Word}} & \multicolumn{1}{C{1.6cm}|}{\textbf{RIG}} \\ \hhline{|-|-|-|~|-|-|-|} 

\endhead 

\hhline{|-|-|-|~|-|-|-|}

\endfoot
 1 & paper & $ 9.7\times10^{-2} $ &   & 51 & scheme & $ 1\times10^{-2} $ \\ \hhline{|-|-|-|~|-|-|-|} 
 
 2 & propos & $ 7\times10^{-2} $ &   & 52 & examin & $ 1\times10^{-2} $ \\ \hhline{|-|-|-|~|-|-|-|} 
 
 3 & system & $ 5.7\times10^{-2} $ &   & 53 & trajectori & $ 9.9\times10^{-3} $ \\ \hhline{|-|-|-|~|-|-|-|} 
 
 4 & control & $ 4.4\times10^{-2} $ &   & 54 & year & $ 9.9\times10^{-3} $ \\ \hhline{|-|-|-|~|-|-|-|} 
 
 5 & robot & $ 3.9\times10^{-2} $ &   & 55 & constraint & $ 9.5\times10^{-3} $ \\ \hhline{|-|-|-|~|-|-|-|} 
 
 6 & problem & $ 3.8\times10^{-2} $ &   & 56 & background & $ 9.4\times10^{-3} $ \\ \hhline{|-|-|-|~|-|-|-|} 
 
 7 & algorithm & $ 3.7\times10^{-2} $ &   & 57 & converg & $ 9.3\times10^{-3} $ \\ \hhline{|-|-|-|~|-|-|-|} 
 
 8 & simul & $ 3.6\times10^{-2} $ &   & 58 & motion & $ 9.2\times10^{-3} $ \\
\end{longtable}
 \newpage 
 
 \clearpage 

\newpage 

\thispagestyle{plain} 

\vspace*{-1.6in}

\setlength{\LTleft}{-20cm plus -1fill}

\setlength{\LTright}{\LTleft} 

\begin{longtable}[h]{| C{.6cm}| L{5cm} |  R{1.6cm}  |  R{.3cm} |  C{.6cm}| L{5cm} |  R{1.6cm} | }

\caption{Behavioral Sciences}\\ 

\hhline{|-|-|-|~|-|-|-|}

\multicolumn{1}{|C{.6cm}|}{\textbf{No.}} & \multicolumn{1}{C{5cm}|}{\textbf{Word}} & \multicolumn{1}{C{1.6cm}|}{\textbf{RIG}} &   &\multicolumn{1}{C{.6cm}|}{\textbf{No.}} & \multicolumn{1}{C{5cm}|}{\textbf{Word}} & \multicolumn{1}{C{1.6cm}|}{\textbf{RIG}} \\ \hhline{|-|-|-|~|-|-|-|} 

\endhead 

\hhline{|-|-|-|~|-|-|-|}

\endfoot
 1 & behavior & $ 5.2\times10^{-2} $ &   & 51 & bird & $ 1.2\times10^{-2} $ \\ \hhline{|-|-|-|~|-|-|-|} 
 
 2 & anim & $ 4.4\times10^{-2} $ &   & 52 & eat & $ 1.2\times10^{-2} $ \\ \hhline{|-|-|-|~|-|-|-|} 
 
 3 & task & $ 4\times10^{-2} $ &   & 53 & forag & $ 1.2\times10^{-2} $ \\ \hhline{|-|-|-|~|-|-|-|} 
 
 4 & male & $ 3.7\times10^{-2} $ &   & 54 & either & $ 1.2\times10^{-2} $ \\ \hhline{|-|-|-|~|-|-|-|} 
 
 5 & suggest & $ 3.7\times10^{-2} $ &   & 55 & amygdala & $ 1.2\times10^{-2} $ \\ \hhline{|-|-|-|~|-|-|-|} 
 
 6 & cognit & $ 3.5\times10^{-2} $ &   & 56 & examin & $ 1.2\times10^{-2} $ \\ \hhline{|-|-|-|~|-|-|-|} 
 
 7 & brain & $ 3.4\times10^{-2} $ &   & 57 & affect & $ 1.1\times10^{-2} $ \\ \hhline{|-|-|-|~|-|-|-|} 
 
 8 & whether & $ 3.1\times10^{-2} $ &   & 58 & locomotor & $ 1.1\times10^{-2} $ \\
\end{longtable}
 \newpage 
 
 \clearpage 

\newpage 

\thispagestyle{plain} 

\vspace*{-1.6in}

\setlength{\LTleft}{-20cm plus -1fill}

\setlength{\LTright}{\LTleft} 

\begin{longtable}[h]{| C{.6cm}| L{5cm} |  R{1.6cm}  |  R{.3cm} |  C{.6cm}| L{5cm} |  R{1.6cm} | }

\caption{Biochemical Research Methods}\\ 

\hhline{|-|-|-|~|-|-|-|}

\multicolumn{1}{|C{.6cm}|}{\textbf{No.}} & \multicolumn{1}{C{5cm}|}{\textbf{Word}} & \multicolumn{1}{C{1.6cm}|}{\textbf{RIG}} &   &\multicolumn{1}{C{.6cm}|}{\textbf{No.}} & \multicolumn{1}{C{5cm}|}{\textbf{Word}} & \multicolumn{1}{C{1.6cm}|}{\textbf{RIG}} \\ \hhline{|-|-|-|~|-|-|-|} 

\endhead 

\hhline{|-|-|-|~|-|-|-|}

\endfoot
 1 & chromatographi & $ 5.9\times10^{-2} $ &   & 51 & drug & $ 9.4\times10^{-3} $ \\ \hhline{|-|-|-|~|-|-|-|} 
 
 2 & spectrometri & $ 5.1\times10^{-2} $ &   & 52 & analysi & $ 9.4\times10^{-3} $ \\ \hhline{|-|-|-|~|-|-|-|} 
 
 3 & protein & $ 4.2\times10^{-2} $ &   & 53 & quadrupol & $ 9.4\times10^{-3} $ \\ \hhline{|-|-|-|~|-|-|-|} 
 
 4 & proteom & $ 3.4\times10^{-2} $ &   & 54 & molecular & $ 9.2\times10^{-3} $ \\ \hhline{|-|-|-|~|-|-|-|} 
 
 5 & mass & $ 2.8\times10^{-2} $ &   & 55 & molecul & $ 9\times10^{-3} $ \\ \hhline{|-|-|-|~|-|-|-|} 
 
 6 & detect & $ 2.8\times10^{-2} $ &   & 56 & rapid & $ 8.7\times10^{-3} $ \\ \hhline{|-|-|-|~|-|-|-|} 
 
 7 & biolog & $ 2.8\times10^{-2} $ &   & 57 & limit & $ 8.7\times10^{-3} $ \\ \hhline{|-|-|-|~|-|-|-|} 
 
 8 & quantif & $ 2.7\times10^{-2} $ &   & 58 & appli & $ 8.5\times10^{-3} $ \\
\end{longtable}
 \newpage 
 
 \clearpage 

\newpage 

\thispagestyle{plain} 

\vspace*{-1.6in}

\setlength{\LTleft}{-20cm plus -1fill}

\setlength{\LTright}{\LTleft} 

\begin{longtable}[h]{| C{.6cm}| L{5cm} |  R{1.6cm}  |  R{.3cm} |  C{.6cm}| L{5cm} |  R{1.6cm} | }

\caption{Biochemistry and Molecular Biology}\\ 

\hhline{|-|-|-|~|-|-|-|}

\multicolumn{1}{|C{.6cm}|}{\textbf{No.}} & \multicolumn{1}{C{5cm}|}{\textbf{Word}} & \multicolumn{1}{C{1.6cm}|}{\textbf{RIG}} &   &\multicolumn{1}{C{.6cm}|}{\textbf{No.}} & \multicolumn{1}{C{5cm}|}{\textbf{Word}} & \multicolumn{1}{C{1.6cm}|}{\textbf{RIG}} \\ \hhline{|-|-|-|~|-|-|-|} 

\endhead 

\hhline{|-|-|-|~|-|-|-|}

\endfoot
 1 & protein & $ 1.1\times10^{-1} $ &   & 51 & involv & $ 1.3\times10^{-2} $ \\ \hhline{|-|-|-|~|-|-|-|} 
 
 2 & cell & $ 7\times10^{-2} $ &   & 52 & signal & $ 1.3\times10^{-2} $ \\ \hhline{|-|-|-|~|-|-|-|} 
 
 3 & bind & $ 6.3\times10^{-2} $ &   & 53 & apoptosi & $ 1.2\times10^{-2} $ \\ \hhline{|-|-|-|~|-|-|-|} 
 
 4 & express & $ 5.5\times10^{-2} $ &   & 54 & molecul & $ 1.2\times10^{-2} $ \\ \hhline{|-|-|-|~|-|-|-|} 
 
 5 & gene & $ 5.3\times10^{-2} $ &   & 55 & affin & $ 1.2\times10^{-2} $ \\ \hhline{|-|-|-|~|-|-|-|} 
 
 6 & regul & $ 5.2\times10^{-2} $ &   & 56 & termin & $ 1.2\times10^{-2} $ \\ \hhline{|-|-|-|~|-|-|-|} 
 
 7 & activ & $ 5.2\times10^{-2} $ &   & 57 & domain & $ 1.2\times10^{-2} $ \\ \hhline{|-|-|-|~|-|-|-|} 
 
 8 & inhibit & $ 3.8\times10^{-2} $ &   & 58 & mitochondri & $ 1.1\times10^{-2} $ \\
\end{longtable}
 \newpage 
 
 \clearpage 

\newpage 

\thispagestyle{plain} 

\vspace*{-1.6in}

\setlength{\LTleft}{-20cm plus -1fill}

\setlength{\LTright}{\LTleft} 

\begin{longtable}[h]{| C{.6cm}| L{5cm} |  R{1.6cm}  |  R{.3cm} |  C{.6cm}| L{5cm} |  R{1.6cm} | }

\caption{Biodiversity Conservation}\\ 

\hhline{|-|-|-|~|-|-|-|}

\multicolumn{1}{|C{.6cm}|}{\textbf{No.}} & \multicolumn{1}{C{5cm}|}{\textbf{Word}} & \multicolumn{1}{C{1.6cm}|}{\textbf{RIG}} &   &\multicolumn{1}{C{.6cm}|}{\textbf{No.}} & \multicolumn{1}{C{5cm}|}{\textbf{Word}} & \multicolumn{1}{C{1.6cm}|}{\textbf{RIG}} \\ \hhline{|-|-|-|~|-|-|-|} 

\endhead 

\hhline{|-|-|-|~|-|-|-|}

\endfoot
 1 & speci & $ 1.8\times10^{-1} $ &   & 51 & geograph & $ 1.4\times10^{-2} $ \\ \hhline{|-|-|-|~|-|-|-|} 
 
 2 & habitat & $ 1.1\times10^{-1} $ &   & 52 & south & $ 1.4\times10^{-2} $ \\ \hhline{|-|-|-|~|-|-|-|} 
 
 3 & conserv & $ 9.6\times10^{-2} $ &   & 53 & distribut & $ 1.4\times10^{-2} $ \\ \hhline{|-|-|-|~|-|-|-|} 
 
 4 & biodivers & $ 6.2\times10^{-2} $ &   & 54 & individu & $ 1.4\times10^{-2} $ \\ \hhline{|-|-|-|~|-|-|-|} 
 
 5 & ecolog & $ 5.9\times10^{-2} $ &   & 55 & region & $ 1.4\times10^{-2} $ \\ \hhline{|-|-|-|~|-|-|-|} 
 
 6 & popul & $ 5.7\times10^{-2} $ &   & 56 & cover & $ 1.4\times10^{-2} $ \\ \hhline{|-|-|-|~|-|-|-|} 
 
 7 & ecosystem & $ 5.5\times10^{-2} $ &   & 57 & eastern & $ 1.3\times10^{-2} $ \\ \hhline{|-|-|-|~|-|-|-|} 
 
 8 & forest & $ 5.1\times10^{-2} $ &   & 58 & park & $ 1.3\times10^{-2} $ \\
\end{longtable}
 \newpage 
 
 \clearpage 

\newpage 

\thispagestyle{plain} 

\vspace*{-1.6in}

\setlength{\LTleft}{-20cm plus -1fill}

\setlength{\LTright}{\LTleft} 

\begin{longtable}[h]{| C{.6cm}| L{5cm} |  R{1.6cm}  |  R{.3cm} |  C{.6cm}| L{5cm} |  R{1.6cm} | }

\caption{Biology}\\ 

\hhline{|-|-|-|~|-|-|-|}

\multicolumn{1}{|C{.6cm}|}{\textbf{No.}} & \multicolumn{1}{C{5cm}|}{\textbf{Word}} & \multicolumn{1}{C{1.6cm}|}{\textbf{RIG}} &   &\multicolumn{1}{C{.6cm}|}{\textbf{No.}} & \multicolumn{1}{C{5cm}|}{\textbf{Word}} & \multicolumn{1}{C{1.6cm}|}{\textbf{RIG}} \\ \hhline{|-|-|-|~|-|-|-|} 

\endhead 

\hhline{|-|-|-|~|-|-|-|}

\endfoot
 1 & speci & $ 2.3\times10^{-2} $ &   & 51 & conserv & $ 3.8\times10^{-3} $ \\ \hhline{|-|-|-|~|-|-|-|} 
 
 2 & paper & $ 1.4\times10^{-2} $ &   & 52 & pattern & $ 3.7\times10^{-3} $ \\ \hhline{|-|-|-|~|-|-|-|} 
 
 3 & anim & $ 1.2\times10^{-2} $ &   & 53 & transcript & $ 3.7\times10^{-3} $ \\ \hhline{|-|-|-|~|-|-|-|} 
 
 4 & cell & $ 1.1\times10^{-2} $ &   & 54 & patient & $ 3.6\times10^{-3} $ \\ \hhline{|-|-|-|~|-|-|-|} 
 
 5 & biolog & $ 1.1\times10^{-2} $ &   & 55 & vertebr & $ 3.6\times10^{-3} $ \\ \hhline{|-|-|-|~|-|-|-|} 
 
 6 & protein & $ 1.1\times10^{-2} $ &   & 56 & prey & $ 3.6\times10^{-3} $ \\ \hhline{|-|-|-|~|-|-|-|} 
 
 7 & gene & $ 9.9\times10^{-3} $ &   & 57 & mediat & $ 3.6\times10^{-3} $ \\ \hhline{|-|-|-|~|-|-|-|} 
 
 8 & evolutionari & $ 9\times10^{-3} $ &   & 58 & mice & $ 3.5\times10^{-3} $ \\
\end{longtable}
 \newpage 
 
 \clearpage 

\newpage 

\thispagestyle{plain} 

\vspace*{-1.6in}

\setlength{\LTleft}{-20cm plus -1fill}

\setlength{\LTright}{\LTleft} 

\begin{longtable}[h]{| C{.6cm}| L{5cm} |  R{1.6cm}  |  R{.3cm} |  C{.6cm}| L{5cm} |  R{1.6cm} | }

\caption{Biophysics}\\ 

\hhline{|-|-|-|~|-|-|-|}

\multicolumn{1}{|C{.6cm}|}{\textbf{No.}} & \multicolumn{1}{C{5cm}|}{\textbf{Word}} & \multicolumn{1}{C{1.6cm}|}{\textbf{RIG}} &   &\multicolumn{1}{C{.6cm}|}{\textbf{No.}} & \multicolumn{1}{C{5cm}|}{\textbf{Word}} & \multicolumn{1}{C{1.6cm}|}{\textbf{RIG}} \\ \hhline{|-|-|-|~|-|-|-|} 

\endhead 

\hhline{|-|-|-|~|-|-|-|}

\endfoot
 1 & protein & $ 5.6\times10^{-2} $ &   & 51 & signal & $ 6.5\times10^{-3} $ \\ \hhline{|-|-|-|~|-|-|-|} 
 
 2 & bind & $ 4.2\times10^{-2} $ &   & 52 & phosphoryl & $ 6.4\times10^{-3} $ \\ \hhline{|-|-|-|~|-|-|-|} 
 
 3 & cell & $ 3.8\times10^{-2} $ &   & 53 & target & $ 6.3\times10^{-3} $ \\ \hhline{|-|-|-|~|-|-|-|} 
 
 4 & membran & $ 2\times10^{-2} $ &   & 54 & fold & $ 6.3\times10^{-3} $ \\ \hhline{|-|-|-|~|-|-|-|} 
 
 5 & paper & $ 1.9\times10^{-2} $ &   & 55 & bilay & $ 6.2\times10^{-3} $ \\ \hhline{|-|-|-|~|-|-|-|} 
 
 6 & conform & $ 1.6\times10^{-2} $ &   & 56 & suggest & $ 5.8\times10^{-3} $ \\ \hhline{|-|-|-|~|-|-|-|} 
 
 7 & angstrom & $ 1.6\times10^{-2} $ &   & 57 & kinas & $ 5.8\times10^{-3} $ \\ \hhline{|-|-|-|~|-|-|-|} 
 
 8 & molecular & $ 1.6\times10^{-2} $ &   & 58 & play & $ 5.7\times10^{-3} $ \\
\end{longtable}
 \newpage 
 
 \clearpage 

\newpage 

\thispagestyle{plain} 

\vspace*{-1.6in}

\setlength{\LTleft}{-20cm plus -1fill}

\setlength{\LTright}{\LTleft} 

\begin{longtable}[h]{| C{.6cm}| L{5cm} |  R{1.6cm}  |  R{.3cm} |  C{.6cm}| L{5cm} |  R{1.6cm} | }

\caption{Biotechnology and Applied Microbiology}\\ 

\hhline{|-|-|-|~|-|-|-|}

\multicolumn{1}{|C{.6cm}|}{\textbf{No.}} & \multicolumn{1}{C{5cm}|}{\textbf{Word}} & \multicolumn{1}{C{1.6cm}|}{\textbf{RIG}} &   &\multicolumn{1}{C{.6cm}|}{\textbf{No.}} & \multicolumn{1}{C{5cm}|}{\textbf{Word}} & \multicolumn{1}{C{1.6cm}|}{\textbf{RIG}} \\ \hhline{|-|-|-|~|-|-|-|} 

\endhead 

\hhline{|-|-|-|~|-|-|-|}

\endfoot
 1 & gene & $ 5.4\times10^{-2} $ &   & 51 & biofuel & $ 8.2\times10^{-3} $ \\ \hhline{|-|-|-|~|-|-|-|} 
 
 2 & strain & $ 3\times10^{-2} $ &   & 52 & potenti & $ 8.1\times10^{-3} $ \\ \hhline{|-|-|-|~|-|-|-|} 
 
 3 & genom & $ 3\times10^{-2} $ &   & 53 & concentr & $ 8\times10^{-3} $ \\ \hhline{|-|-|-|~|-|-|-|} 
 
 4 & product & $ 2.9\times10^{-2} $ &   & 54 & virus & $ 7.6\times10^{-3} $ \\ \hhline{|-|-|-|~|-|-|-|} 
 
 5 & ferment & $ 2.9\times10^{-2} $ &   & 55 & fold & $ 7.6\times10^{-3} $ \\ \hhline{|-|-|-|~|-|-|-|} 
 
 6 & cell & $ 2.9\times10^{-2} $ &   & 56 & sugar & $ 7.5\times10^{-3} $ \\ \hhline{|-|-|-|~|-|-|-|} 
 
 7 & protein & $ 2.8\times10^{-2} $ &   & 57 & 16s & $ 7.3\times10^{-3} $ \\ \hhline{|-|-|-|~|-|-|-|} 
 
 8 & sequenc & $ 2.7\times10^{-2} $ &   & 58 & medium & $ 7.3\times10^{-3} $ \\
\end{longtable}
 \newpage 
 
 \clearpage 

\newpage 

\thispagestyle{plain} 

\vspace*{-1.6in}

\setlength{\LTleft}{-20cm plus -1fill}

\setlength{\LTright}{\LTleft} 

\begin{longtable}[h]{| C{.6cm}| L{5cm} |  R{1.6cm}  |  R{.3cm} |  C{.6cm}| L{5cm} |  R{1.6cm} | }

\caption{Business}\\ 

\hhline{|-|-|-|~|-|-|-|}

\multicolumn{1}{|C{.6cm}|}{\textbf{No.}} & \multicolumn{1}{C{5cm}|}{\textbf{Word}} & \multicolumn{1}{C{1.6cm}|}{\textbf{RIG}} &   &\multicolumn{1}{C{.6cm}|}{\textbf{No.}} & \multicolumn{1}{C{5cm}|}{\textbf{Word}} & \multicolumn{1}{C{1.6cm}|}{\textbf{RIG}} \\ \hhline{|-|-|-|~|-|-|-|} 

\endhead 

\hhline{|-|-|-|~|-|-|-|}

\endfoot
 1 & firm & $ 8.4\times10^{-2} $ &   & 51 & surfac & $ 1.2\times10^{-2} $ \\ \hhline{|-|-|-|~|-|-|-|} 
 
 2 & market & $ 8\times10^{-2} $ &   & 52 & advertis & $ 1.2\times10^{-2} $ \\ \hhline{|-|-|-|~|-|-|-|} 
 
 3 & busi & $ 7.8\times10^{-2} $ &   & 53 & focus & $ 1.2\times10^{-2} $ \\ \hhline{|-|-|-|~|-|-|-|} 
 
 4 & research & $ 6.2\times10^{-2} $ &   & 54 & treatment & $ 1.2\times10^{-2} $ \\ \hhline{|-|-|-|~|-|-|-|} 
 
 5 & compani & $ 5.9\times10^{-2} $ &   & 55 & influenc & $ 1.2\times10^{-2} $ \\ \hhline{|-|-|-|~|-|-|-|} 
 
 6 & manag & $ 4.6\times10^{-2} $ &   & 56 & develop & $ 1.2\times10^{-2} $ \\ \hhline{|-|-|-|~|-|-|-|} 
 
 7 & innov & $ 3.6\times10^{-2} $ &   & 57 & govern & $ 1.2\times10^{-2} $ \\ \hhline{|-|-|-|~|-|-|-|} 
 
 8 & organiz & $ 3.6\times10^{-2} $ &   & 58 & sale & $ 1.1\times10^{-2} $ \\
\end{longtable}
 \newpage 
 
 \clearpage 

\newpage 

\thispagestyle{plain} 

\vspace*{-1.6in}

\setlength{\LTleft}{-20cm plus -1fill}

\setlength{\LTright}{\LTleft} 

\begin{longtable}[h]{| C{.6cm}| L{5cm} |  R{1.6cm}  |  R{.3cm} |  C{.6cm}| L{5cm} |  R{1.6cm} | }

\caption{Business, Finance}\\ 

\hhline{|-|-|-|~|-|-|-|}

\multicolumn{1}{|C{.6cm}|}{\textbf{No.}} & \multicolumn{1}{C{5cm}|}{\textbf{Word}} & \multicolumn{1}{C{1.6cm}|}{\textbf{RIG}} &   &\multicolumn{1}{C{.6cm}|}{\textbf{No.}} & \multicolumn{1}{C{5cm}|}{\textbf{Word}} & \multicolumn{1}{C{1.6cm}|}{\textbf{RIG}} \\ \hhline{|-|-|-|~|-|-|-|} 

\endhead 

\hhline{|-|-|-|~|-|-|-|}

\endfoot
 1 & market & $ 1.1\times10^{-1} $ &   & 51 & european & $ 1.1\times10^{-2} $ \\ \hhline{|-|-|-|~|-|-|-|} 
 
 2 & financi & $ 1.1\times10^{-1} $ &   & 52 & treatment & $ 1.1\times10^{-2} $ \\ \hhline{|-|-|-|~|-|-|-|} 
 
 3 & firm & $ 7\times10^{-2} $ &   & 53 & clinic & $ 1.1\times10^{-2} $ \\ \hhline{|-|-|-|~|-|-|-|} 
 
 4 & price & $ 6.1\times10^{-2} $ &   & 54 & institut & $ 1.1\times10^{-2} $ \\ \hhline{|-|-|-|~|-|-|-|} 
 
 5 & investor & $ 5.9\times10^{-2} $ &   & 55 & incent & $ 1.1\times10^{-2} $ \\ \hhline{|-|-|-|~|-|-|-|} 
 
 6 & stock & $ 5.4\times10^{-2} $ &   & 56 & protein & $ 1\times10^{-2} $ \\ \hhline{|-|-|-|~|-|-|-|} 
 
 7 & bank & $ 4.6\times10^{-2} $ &   & 57 & conclus & $ 9.9\times10^{-3} $ \\ \hhline{|-|-|-|~|-|-|-|} 
 
 8 & asset & $ 4.5\times10^{-2} $ &   & 58 & experiment & $ 9.9\times10^{-3} $ \\
\end{longtable}
 \newpage 
 
 \clearpage 

\newpage 

\thispagestyle{plain} 

\vspace*{-1.6in}

\setlength{\LTleft}{-20cm plus -1fill}

\setlength{\LTright}{\LTleft} 

\begin{longtable}[h]{| C{.6cm}| L{5cm} |  R{1.6cm}  |  R{.3cm} |  C{.6cm}| L{5cm} |  R{1.6cm} | }

\caption{Cardiac and Cardiovascular Systems}\\ 

\hhline{|-|-|-|~|-|-|-|}

\multicolumn{1}{|C{.6cm}|}{\textbf{No.}} & \multicolumn{1}{C{5cm}|}{\textbf{Word}} & \multicolumn{1}{C{1.6cm}|}{\textbf{RIG}} &   &\multicolumn{1}{C{.6cm}|}{\textbf{No.}} & \multicolumn{1}{C{5cm}|}{\textbf{Word}} & \multicolumn{1}{C{1.6cm}|}{\textbf{RIG}} \\ \hhline{|-|-|-|~|-|-|-|} 

\endhead 

\hhline{|-|-|-|~|-|-|-|}

\endfoot
 1 & patient & $ 1.8\times10^{-1} $ &   & 51 & hospit & $ 2.8\times10^{-2} $ \\ \hhline{|-|-|-|~|-|-|-|} 
 
 2 & cardiac & $ 1.6\times10^{-1} $ &   & 52 & median & $ 2.7\times10^{-2} $ \\ \hhline{|-|-|-|~|-|-|-|} 
 
 3 & conclus & $ 1.5\times10^{-1} $ &   & 53 & cardiomyopathi & $ 2.7\times10^{-2} $ \\ \hhline{|-|-|-|~|-|-|-|} 
 
 4 & coronari & $ 1.4\times10^{-1} $ &   & 54 & revascular & $ 2.7\times10^{-2} $ \\ \hhline{|-|-|-|~|-|-|-|} 
 
 5 & heart & $ 1.4\times10^{-1} $ &   & 55 & multivari & $ 2.7\times10^{-2} $ \\ \hhline{|-|-|-|~|-|-|-|} 
 
 6 & ventricular & $ 1.4\times10^{-1} $ &   & 56 & confid & $ 2.7\times10^{-2} $ \\ \hhline{|-|-|-|~|-|-|-|} 
 
 7 & myocardi & $ 1.1\times10^{-1} $ &   & 57 & hypertens & $ 2.7\times10^{-2} $ \\ \hhline{|-|-|-|~|-|-|-|} 
 
 8 & arteri & $ 1.1\times10^{-1} $ &   & 58 & month & $ 2.6\times10^{-2} $ \\
\end{longtable}
 \newpage 
 
 \clearpage 

\newpage 

\thispagestyle{plain} 

\vspace*{-1.6in}

\setlength{\LTleft}{-20cm plus -1fill}

\setlength{\LTright}{\LTleft} 

\begin{longtable}[h]{| C{.6cm}| L{5cm} |  R{1.6cm}  |  R{.3cm} |  C{.6cm}| L{5cm} |  R{1.6cm} | }

\caption{Cell and Tissue Engineering}\\ 

\hhline{|-|-|-|~|-|-|-|}

\multicolumn{1}{|C{.6cm}|}{\textbf{No.}} & \multicolumn{1}{C{5cm}|}{\textbf{Word}} & \multicolumn{1}{C{1.6cm}|}{\textbf{RIG}} &   &\multicolumn{1}{C{.6cm}|}{\textbf{No.}} & \multicolumn{1}{C{5cm}|}{\textbf{Word}} & \multicolumn{1}{C{1.6cm}|}{\textbf{RIG}} \\ \hhline{|-|-|-|~|-|-|-|} 

\endhead 

\hhline{|-|-|-|~|-|-|-|}

\endfoot
 1 & stem & $ 2.4\times10^{-1} $ &   & 51 & protein & $ 1.8\times10^{-2} $ \\ \hhline{|-|-|-|~|-|-|-|} 
 
 2 & cell & $ 2\times10^{-1} $ &   & 52 & chondrocyt & $ 1.8\times10^{-2} $ \\ \hhline{|-|-|-|~|-|-|-|} 
 
 3 & mesenchym & $ 1\times10^{-1} $ &   & 53 & endotheli & $ 1.8\times10^{-2} $ \\ \hhline{|-|-|-|~|-|-|-|} 
 
 4 & pluripot & $ 9.3\times10^{-2} $ &   & 54 & paper & $ 1.7\times10^{-2} $ \\ \hhline{|-|-|-|~|-|-|-|} 
 
 5 & differenti & $ 8\times10^{-2} $ &   & 55 & adipos & $ 1.7\times10^{-2} $ \\ \hhline{|-|-|-|~|-|-|-|} 
 
 6 & tissu & $ 7.5\times10^{-2} $ &   & 56 & cellular & $ 1.7\times10^{-2} $ \\ \hhline{|-|-|-|~|-|-|-|} 
 
 7 & progenitor & $ 7.2\times10^{-2} $ &   & 57 & promot & $ 1.7\times10^{-2} $ \\ \hhline{|-|-|-|~|-|-|-|} 
 
 8 & bone & $ 6.9\times10^{-2} $ &   & 58 & phenotyp & $ 1.6\times10^{-2} $ \\
\end{longtable}
 \newpage 
 
 \clearpage 

\newpage 

\thispagestyle{plain} 

\vspace*{-1.6in}

\setlength{\LTleft}{-20cm plus -1fill}

\setlength{\LTright}{\LTleft} 

\begin{longtable}[h]{| C{.6cm}| L{5cm} |  R{1.6cm}  |  R{.3cm} |  C{.6cm}| L{5cm} |  R{1.6cm} | }

\caption{Cell Biology}\\ 

\hhline{|-|-|-|~|-|-|-|}

\multicolumn{1}{|C{.6cm}|}{\textbf{No.}} & \multicolumn{1}{C{5cm}|}{\textbf{Word}} & \multicolumn{1}{C{1.6cm}|}{\textbf{RIG}} &   &\multicolumn{1}{C{.6cm}|}{\textbf{No.}} & \multicolumn{1}{C{5cm}|}{\textbf{Word}} & \multicolumn{1}{C{1.6cm}|}{\textbf{RIG}} \\ \hhline{|-|-|-|~|-|-|-|} 

\endhead 

\hhline{|-|-|-|~|-|-|-|}

\endfoot
 1 & cell & $ 1.7\times10^{-1} $ &   & 51 & beta & $ 1.7\times10^{-2} $ \\ \hhline{|-|-|-|~|-|-|-|} 
 
 2 & express & $ 1.1\times10^{-1} $ &   & 52 & factor & $ 1.6\times10^{-2} $ \\ \hhline{|-|-|-|~|-|-|-|} 
 
 3 & protein & $ 1\times10^{-1} $ &   & 53 & extracellular & $ 1.6\times10^{-2} $ \\ \hhline{|-|-|-|~|-|-|-|} 
 
 4 & regul & $ 9.5\times10^{-2} $ &   & 54 & downregul & $ 1.6\times10^{-2} $ \\ \hhline{|-|-|-|~|-|-|-|} 
 
 5 & induc & $ 6.2\times10^{-2} $ &   & 55 & phenotyp & $ 1.6\times10^{-2} $ \\ \hhline{|-|-|-|~|-|-|-|} 
 
 6 & inhibit & $ 5.6\times10^{-2} $ &   & 56 & mutat & $ 1.5\times10^{-2} $ \\ \hhline{|-|-|-|~|-|-|-|} 
 
 7 & pathway & $ 5.3\times10^{-2} $ &   & 57 & embryon & $ 1.5\times10^{-2} $ \\ \hhline{|-|-|-|~|-|-|-|} 
 
 8 & activ & $ 5.2\times10^{-2} $ &   & 58 & involv & $ 1.5\times10^{-2} $ \\
\end{longtable}
 \newpage 
 
 \clearpage 

\newpage 

\thispagestyle{plain} 

\vspace*{-1.6in}

\setlength{\LTleft}{-20cm plus -1fill}

\setlength{\LTright}{\LTleft} 

\begin{longtable}[h]{| C{.6cm}| L{5cm} |  R{1.6cm}  |  R{.3cm} |  C{.6cm}| L{5cm} |  R{1.6cm} | }

\caption{Chemistry, Analytical}\\ 

\hhline{|-|-|-|~|-|-|-|}

\multicolumn{1}{|C{.6cm}|}{\textbf{No.}} & \multicolumn{1}{C{5cm}|}{\textbf{Word}} & \multicolumn{1}{C{1.6cm}|}{\textbf{RIG}} &   &\multicolumn{1}{C{.6cm}|}{\textbf{No.}} & \multicolumn{1}{C{5cm}|}{\textbf{Word}} & \multicolumn{1}{C{1.6cm}|}{\textbf{RIG}} \\ \hhline{|-|-|-|~|-|-|-|} 

\endhead 

\hhline{|-|-|-|~|-|-|-|}

\endfoot
 1 & detect & $ 8.4\times10^{-2} $ &   & 51 & use & $ 1.3\times10^{-2} $ \\ \hhline{|-|-|-|~|-|-|-|} 
 
 2 & chromatographi & $ 8\times10^{-2} $ &   & 52 & microfluid & $ 1.3\times10^{-2} $ \\ \hhline{|-|-|-|~|-|-|-|} 
 
 3 & spectrometri & $ 6.9\times10^{-2} $ &   & 53 & solvent & $ 1.3\times10^{-2} $ \\ \hhline{|-|-|-|~|-|-|-|} 
 
 4 & sampl & $ 5.3\times10^{-2} $ &   & 54 & methanol & $ 1.3\times10^{-2} $ \\ \hhline{|-|-|-|~|-|-|-|} 
 
 5 & analyt & $ 4.9\times10^{-2} $ &   & 55 & esi & $ 1.3\times10^{-2} $ \\ \hhline{|-|-|-|~|-|-|-|} 
 
 6 & liquid & $ 4.6\times10^{-2} $ &   & 56 & gold & $ 1.2\times10^{-2} $ \\ \hhline{|-|-|-|~|-|-|-|} 
 
 7 & sensit & $ 4\times10^{-2} $ &   & 57 & monitor & $ 1.2\times10^{-2} $ \\ \hhline{|-|-|-|~|-|-|-|} 
 
 8 & limit & $ 3.9\times10^{-2} $ &   & 58 & conclus & $ 1.2\times10^{-2} $ \\
\end{longtable}
 \newpage 
 
 \clearpage 

\newpage 

\thispagestyle{plain} 

\vspace*{-1.6in}

\setlength{\LTleft}{-20cm plus -1fill}

\setlength{\LTright}{\LTleft} 

\begin{longtable}[h]{| C{.6cm}| L{5cm} |  R{1.6cm}  |  R{.3cm} |  C{.6cm}| L{5cm} |  R{1.6cm} | }

\caption{Chemistry, Applied}\\ 

\hhline{|-|-|-|~|-|-|-|}

\multicolumn{1}{|C{.6cm}|}{\textbf{No.}} & \multicolumn{1}{C{5cm}|}{\textbf{Word}} & \multicolumn{1}{C{1.6cm}|}{\textbf{RIG}} &   &\multicolumn{1}{C{.6cm}|}{\textbf{No.}} & \multicolumn{1}{C{5cm}|}{\textbf{Word}} & \multicolumn{1}{C{1.6cm}|}{\textbf{RIG}} \\ \hhline{|-|-|-|~|-|-|-|} 

\endhead 

\hhline{|-|-|-|~|-|-|-|}

\endfoot
 1 & acid & $ 3.9\times10^{-2} $ &   & 51 & contain & $ 7.5\times10^{-3} $ \\ \hhline{|-|-|-|~|-|-|-|} 
 
 2 & prepar & $ 2.3\times10^{-2} $ &   & 52 & algorithm & $ 7.4\times10^{-3} $ \\ \hhline{|-|-|-|~|-|-|-|} 
 
 3 & compound & $ 2.3\times10^{-2} $ &   & 53 & chitosan & $ 7.3\times10^{-3} $ \\ \hhline{|-|-|-|~|-|-|-|} 
 
 4 & patient & $ 1.9\times10^{-2} $ &   & 54 & problem & $ 7.3\times10^{-3} $ \\ \hhline{|-|-|-|~|-|-|-|} 
 
 5 & catalyst & $ 1.8\times10^{-2} $ &   & 55 & zeolit & $ 7.1\times10^{-3} $ \\ \hhline{|-|-|-|~|-|-|-|} 
 
 6 & content & $ 1.8\times10^{-2} $ &   & 56 & liquid & $ 7\times10^{-3} $ \\ \hhline{|-|-|-|~|-|-|-|} 
 
 7 & oil & $ 1.8\times10^{-2} $ &   & 57 & fatti & $ 6.9\times10^{-3} $ \\ \hhline{|-|-|-|~|-|-|-|} 
 
 8 & chromatographi & $ 1.8\times10^{-2} $ &   & 58 & gas & $ 6.7\times10^{-3} $ \\
\end{longtable}
 \newpage 
 
 \input{tablecodes/RIG1_100_34.txt} 
 
 \input{tablecodes/RIG1_100_35.txt} 
 
 \input{tablecodes/RIG1_100_36.txt} 
 
 \input{tablecodes/RIG1_100_37.txt} 
 
 \input{tablecodes/RIG1_100_38.txt} 
 
 \input{tablecodes/RIG1_100_39.txt} 
 
 \input{tablecodes/RIG1_100_40.txt} 
 
 \input{tablecodes/RIG1_100_41.txt} 
 
 \input{tablecodes/RIG1_100_42.txt} 
 
 \input{tablecodes/RIG1_100_43.txt} 
 
 \input{tablecodes/RIG1_100_44.txt} 
 
 \input{tablecodes/RIG1_100_45.txt} 
 
 \input{tablecodes/RIG1_100_46.txt} 
 
 \input{tablecodes/RIG1_100_47.txt} 
 
 \input{tablecodes/RIG1_100_48.txt} 
 
 \input{tablecodes/RIG1_100_49.txt} 
 
 \input{tablecodes/RIG1_100_50.txt} 
 
 \input{tablecodes/RIG1_100_51.txt} 
 
 \input{tablecodes/RIG1_100_52.txt} 
 
 \input{tablecodes/RIG1_100_53.txt} 
 
 \input{tablecodes/RIG1_100_54.txt} 
 
 \input{tablecodes/RIG1_100_55.txt} 
 
 \input{tablecodes/RIG1_100_56.txt} 
 
 \input{tablecodes/RIG1_100_57.txt} 
 
 \input{tablecodes/RIG1_100_58.txt} 
 
 \input{tablecodes/RIG1_100_59.txt} 
 
 \input{tablecodes/RIG1_100_60.txt} 
 
 \input{tablecodes/RIG1_100_61.txt} 
 
 \input{tablecodes/RIG1_100_62.txt} 
 
 \input{tablecodes/RIG1_100_63.txt} 
 
 \input{tablecodes/RIG1_100_64.txt} 
 
 \input{tablecodes/RIG1_100_65.txt} 
 
 \input{tablecodes/RIG1_100_66.txt} 
 
 \input{tablecodes/RIG1_100_67.txt} 
 
 \input{tablecodes/RIG1_100_68.txt} 
 
 \input{tablecodes/RIG1_100_69.txt} 
 
 \input{tablecodes/RIG1_100_70.txt} 
 
 \input{tablecodes/RIG1_100_71.txt} 
 
 \input{tablecodes/RIG1_100_72.txt} 
 
 \input{tablecodes/RIG1_100_73.txt} 
 
 \input{tablecodes/RIG1_100_74.txt} 
 
 \input{tablecodes/RIG1_100_75.txt} 
 
 \input{tablecodes/RIG1_100_76.txt} 
 
 \input{tablecodes/RIG1_100_77.txt} 
 
 \input{tablecodes/RIG1_100_78.txt} 
 
 \input{tablecodes/RIG1_100_79.txt} 
 
 \input{tablecodes/RIG1_100_80.txt} 
 
 \input{tablecodes/RIG1_100_81.txt} 
 
 \input{tablecodes/RIG1_100_82.txt} 
 
 \input{tablecodes/RIG1_100_83.txt} 
 
 \input{tablecodes/RIG1_100_84.txt} 
 
 \input{tablecodes/RIG1_100_85.txt} 
 
 \input{tablecodes/RIG1_100_86.txt} 
 
 \input{tablecodes/RIG1_100_87.txt} 
 
 \input{tablecodes/RIG1_100_88.txt} 
 
 \input{tablecodes/RIG1_100_89.txt} 
 
 \input{tablecodes/RIG1_100_90.txt} 
 
 \input{tablecodes/RIG1_100_91.txt} 
 
 \input{tablecodes/RIG1_100_92.txt} 
 
 \input{tablecodes/RIG1_100_93.txt} 
 
 \input{tablecodes/RIG1_100_94.txt} 
 
 \input{tablecodes/RIG1_100_95.txt} 
 
 \input{tablecodes/RIG1_100_96.txt} 
 
 \input{tablecodes/RIG1_100_97.txt} 
 
 \input{tablecodes/RIG1_100_98.txt} 
 
 \input{tablecodes/RIG1_100_99.txt} 
 
 \input{tablecodes/RIG1_100_100.txt} 
 
 \input{tablecodes/RIG1_100_101.txt} 
 
 \input{tablecodes/RIG1_100_102.txt} 
 
 \input{tablecodes/RIG1_100_103.txt} 
 
 \input{tablecodes/RIG1_100_104.txt} 
 
 \input{tablecodes/RIG1_100_105.txt} 
 
 \input{tablecodes/RIG1_100_106.txt} 
 
 \input{tablecodes/RIG1_100_107.txt} 
 
 \input{tablecodes/RIG1_100_108.txt} 
 
 \input{tablecodes/RIG1_100_109.txt} 
 
 \input{tablecodes/RIG1_100_110.txt} 
 
 \input{tablecodes/RIG1_100_111.txt} 
 
 \input{tablecodes/RIG1_100_112.txt} 
 
 \input{tablecodes/RIG1_100_113.txt} 
 
 \input{tablecodes/RIG1_100_114.txt} 
 
 \input{tablecodes/RIG1_100_115.txt} 
 
 \input{tablecodes/RIG1_100_116.txt} 
 
 \input{tablecodes/RIG1_100_117.txt} 
 
 \input{tablecodes/RIG1_100_118.txt} 
 
 \input{tablecodes/RIG1_100_119.txt} 
 
 \input{tablecodes/RIG1_100_120.txt} 
 
 \input{tablecodes/RIG1_100_121.txt} 
 
 \input{tablecodes/RIG1_100_122.txt} 
 
 \input{tablecodes/RIG1_100_123.txt} 
 
 \input{tablecodes/RIG1_100_124.txt} 
 
 \input{tablecodes/RIG1_100_125.txt} 
 
 \input{tablecodes/RIG1_100_126.txt} 
 
 \input{tablecodes/RIG1_100_127.txt} 
 
 \input{tablecodes/RIG1_100_128.txt} 
 
 \input{tablecodes/RIG1_100_129.txt} 
 
 \input{tablecodes/RIG1_100_130.txt} 
 
 \input{tablecodes/RIG1_100_131.txt} 
 
 \input{tablecodes/RIG1_100_132.txt} 
 
 \input{tablecodes/RIG1_100_133.txt} 
 
 \input{tablecodes/RIG1_100_134.txt} 
 
 \input{tablecodes/RIG1_100_135.txt} 
 
 \input{tablecodes/RIG1_100_136.txt} 
 
 \input{tablecodes/RIG1_100_137.txt} 
 
 \input{tablecodes/RIG1_100_138.txt} 
 
 \input{tablecodes/RIG1_100_139.txt} 
 
 \input{tablecodes/RIG1_100_140.txt} 
 
 \input{tablecodes/RIG1_100_141.txt} 
 
 \input{tablecodes/RIG1_100_142.txt} 
 
 \input{tablecodes/RIG1_100_143.txt} 
 
 \input{tablecodes/RIG1_100_144.txt} 
 
 \input{tablecodes/RIG1_100_145.txt} 
 
 \input{tablecodes/RIG1_100_146.txt} 
 
 \input{tablecodes/RIG1_100_147.txt} 
 
 \input{tablecodes/RIG1_100_148.txt} 
 
 \input{tablecodes/RIG1_100_149.txt} 
 
 \input{tablecodes/RIG1_100_150.txt} 
 
 \input{tablecodes/RIG1_100_151.txt} 
 
 \input{tablecodes/RIG1_100_152.txt} 
 
 \input{tablecodes/RIG1_100_153.txt} 
 
 \input{tablecodes/RIG1_100_154.txt} 
 
 \input{tablecodes/RIG1_100_155.txt} 
 
 \input{tablecodes/RIG1_100_156.txt} 
 
 \input{tablecodes/RIG1_100_157.txt} 
 
 \input{tablecodes/RIG1_100_158.txt} 
 
 \input{tablecodes/RIG1_100_159.txt} 
 
 \input{tablecodes/RIG1_100_160.txt} 
 
 \input{tablecodes/RIG1_100_161.txt} 
 
 \input{tablecodes/RIG1_100_162.txt} 
 
 \input{tablecodes/RIG1_100_163.txt} 
 
 \input{tablecodes/RIG1_100_164.txt} 
 
 \input{tablecodes/RIG1_100_165.txt} 
 
 \input{tablecodes/RIG1_100_166.txt} 
 
 \input{tablecodes/RIG1_100_167.txt} 
 
 \input{tablecodes/RIG1_100_168.txt} 
 
 \input{tablecodes/RIG1_100_169.txt} 
 
 \input{tablecodes/RIG1_100_170.txt} 
 
 \input{tablecodes/RIG1_100_171.txt} 
 
 \input{tablecodes/RIG1_100_172.txt} 
 
 \input{tablecodes/RIG1_100_173.txt} 
 
 \input{tablecodes/RIG1_100_174.txt} 
 
 \input{tablecodes/RIG1_100_175.txt} 
 
 \input{tablecodes/RIG1_100_176.txt} 
 
 \input{tablecodes/RIG1_100_177.txt} 
 
 \input{tablecodes/RIG1_100_178.txt} 
 
 \input{tablecodes/RIG1_100_179.txt} 
 
 \input{tablecodes/RIG1_100_180.txt} 
 
 \input{tablecodes/RIG1_100_181.txt} 
 
 \input{tablecodes/RIG1_100_182.txt} 
 
 \input{tablecodes/RIG1_100_183.txt} 
 
 \input{tablecodes/RIG1_100_184.txt} 
 
 \input{tablecodes/RIG1_100_185.txt} 
 
 \input{tablecodes/RIG1_100_186.txt} 
 
 \input{tablecodes/RIG1_100_187.txt} 
 
 \input{tablecodes/RIG1_100_188.txt} 
 
 \input{tablecodes/RIG1_100_189.txt} 
 
 \input{tablecodes/RIG1_100_190.txt} 
 
 \input{tablecodes/RIG1_100_191.txt} 
 
 \input{tablecodes/RIG1_100_192.txt} 
 
 \input{tablecodes/RIG1_100_193.txt} 
 
 \input{tablecodes/RIG1_100_194.txt} 
 
 \input{tablecodes/RIG1_100_195.txt} 
 
 \input{tablecodes/RIG1_100_196.txt} 
 
 \input{tablecodes/RIG1_100_197.txt} 
 
 \input{tablecodes/RIG1_100_198.txt} 
 
 \input{tablecodes/RIG1_100_199.txt} 
 
 \input{tablecodes/RIG1_100_200.txt} 
 
 \input{tablecodes/RIG1_100_201.txt} 
 
 \input{tablecodes/RIG1_100_202.txt} 
 
 \input{tablecodes/RIG1_100_203.txt} 
 
 \input{tablecodes/RIG1_100_204.txt} 
 
 \input{tablecodes/RIG1_100_205.txt} 
 
 \input{tablecodes/RIG1_100_206.txt} 
 
 \input{tablecodes/RIG1_100_207.txt} 
 
 \input{tablecodes/RIG1_100_208.txt} 
 
 \input{tablecodes/RIG1_100_209.txt} 
 
 \input{tablecodes/RIG1_100_210.txt} 
 
 \input{tablecodes/RIG1_100_211.txt} 
 
 \input{tablecodes/RIG1_100_212.txt} 
 
 \input{tablecodes/RIG1_100_213.txt} 
 
 \input{tablecodes/RIG1_100_214.txt} 
 
 \input{tablecodes/RIG1_100_215.txt} 
 
 \input{tablecodes/RIG1_100_216.txt} 
 
 \input{tablecodes/RIG1_100_217.txt} 
 
 \input{tablecodes/RIG1_100_218.txt} 
 
 \input{tablecodes/RIG1_100_219.txt} 
 
 \input{tablecodes/RIG1_100_220.txt} 
 
 \input{tablecodes/RIG1_100_221.txt} 
 
 \input{tablecodes/RIG1_100_222.txt} 
 
 \input{tablecodes/RIG1_100_223.txt} 
 
 \input{tablecodes/RIG1_100_224.txt} 
 
 \input{tablecodes/RIG1_100_225.txt} 
 
 \input{tablecodes/RIG1_100_226.txt} 
 
 \input{tablecodes/RIG1_100_227.txt} 
 
 \input{tablecodes/RIG1_100_228.txt} 
 
 \input{tablecodes/RIG1_100_229.txt} 
 
 \input{tablecodes/RIG1_100_230.txt} 
 
 \input{tablecodes/RIG1_100_231.txt} 
 
 \input{tablecodes/RIG1_100_232.txt} 
 
 \input{tablecodes/RIG1_100_233.txt} 
 
 \input{tablecodes/RIG1_100_234.txt} 
 
 \input{tablecodes/RIG1_100_235.txt} 
 
 \input{tablecodes/RIG1_100_236.txt} 
 
 \input{tablecodes/RIG1_100_237.txt} 
 
 \input{tablecodes/RIG1_100_238.txt} 
 
 \input{tablecodes/RIG1_100_239.txt} 
 
 \input{tablecodes/RIG1_100_240.txt} 
 
 \input{tablecodes/RIG1_100_241.txt} 
 
 \input{tablecodes/RIG1_100_242.txt} 
 
 \input{tablecodes/RIG1_100_243.txt} 
 
 \input{tablecodes/RIG1_100_244.txt} 
 
 \input{tablecodes/RIG1_100_245.txt} 
 
 \input{tablecodes/RIG1_100_246.txt} 
 
 \input{tablecodes/RIG1_100_247.txt} 
 
 \input{tablecodes/RIG1_100_248.txt} 
 
 \input{tablecodes/RIG1_100_249.txt} 
 
 \input{tablecodes/RIG1_100_250.txt} 
 
 \input{tablecodes/RIG1_100_251.txt} 
 
 \input{tablecodes/RIG1_100_252.txt}

\section{Number of words for each category} \label{catwordtab}
\raggedbottom	

\begin{longtable}[h]{|C{1cm}| L{8cm} |R{3cm}| }
	
		\caption{Number of words for each category}\\ 
		\hline
  	\multicolumn{1}{|C{1cm}|}{\textbf{No.}} &\multicolumn{1}{|C{8cm}}{\textbf{Category}} &\multicolumn{1}{|C{3cm}|}{\textbf{Number of Words}}  \\ \hline

\endhead
\hline
\endfoot
1	&	Acoustics	&	4198	\\ \hline
2	&	Agricultural Economics \& Policy	&	2788	\\ \hline
3	&	Agricultural Engineering	&	3727	\\ \hline
4	&	Agriculture, Dairy \& Animal Science	&	4080	\\ \hline
5	&	Agriculture, Multidisciplinary	&	4359	\\ \hline
6	&	Agronomy	&	4180	\\ \hline
7	&	Allergy	&	3466	\\ \hline
8	&	Anatomy \& Morphology	&	3849	\\
\hline

\end{longtable}

\section{List of original attributes (categories) in groups of positive, negative and zero on the principal components.} \label{appndx}
\raggedbottom	

\begin{longtable}[h]{|C{1cm}| L{8cm} |R{3cm}| }
	
		\caption{List of original attributes (categories) in group of positive on the $1^{th}$ principal component. Categories are sorted by values of their component coefficients in descending order, that is, from the greatest contribution to lowest contribution to the component.}\\ 
		\hline
  	  \multicolumn{3}{|C{12cm}|}{\textbf{Positive (PC1)}}  \\ \hline
\multicolumn{1}{|C{1cm}|}{\textbf{No.}} &\multicolumn{1}{|C{8cm}}{\textbf{Attribute}} &\multicolumn{1}{|C{3cm}|}{\textbf{Component Coefficient}}  \\ \hline

\endhead
\hline
\endfoot

1	&	Engineering, Multidisciplinary	&	0.1294	\\ \hline
2	&	Engineering, Electrical \& Electronic	&	0.1177	\\ \hline
3	&	Computer Science, Theory \& Methods	&	0.1161	\\ \hline
4	&	Medicine, General \& Internal	&	0.1155	\\ \hline
5	&	Computer Science, Interdisciplinary Applications	&	0.1145	\\ \hline
6	&	Medicine, Research \& Experimental	&	0.1125	\\ \hline
7	&	Engineering, Industrial	&	0.1102	\\ \hline
8	&	Computer Science, Software Engineering	&	0.1092	\\
\hline

\end{longtable}

\begin{longtable}[h]{|C{1cm}| L{8cm} |R{3cm}| }
	
		\caption{List of original attributes (categories) in group of zero on the $1^{th}$ principal component. Categories are sorted by values of their component coefficients in descending order, that is, from the greatest contribution to lowest contribution to the component.}\\ 
		\hline
  	  \multicolumn{3}{|C{12cm}|}{\textbf{Zero (PC1)}}  \\ \hline
\multicolumn{1}{|C{1cm}|}{\textbf{No.}} &\multicolumn{1}{|C{8cm}}{\textbf{Attribute}} &\multicolumn{1}{|C{3cm}|}{\textbf{Component Coefficient}  } \\ \hline

\endhead
\hline
\endfoot

1	&	Virology	&	0.0314	\\ \hline
2	&	Mycology	&	0.0311	\\ \hline
3	&	Literature, American	&	0.0306	\\ \hline
4	&	Literary Reviews	&	0.0302	\\ \hline
5	&	Theater	&	0.0301	\\ \hline
6	&	Astronomy \& Astrophysics	&	0.0296	\\ \hline
7	&	Meteorology \& Atmospheric Sciences	&	0.0294	\\ \hline
8	&	Linguistics	&	0.0294	\\ \hline
9	&	Materials Science, Paper \& Wood	&	0.0291	\\ \hline
10	&	Psychology, Psychoanalysis	&	0.0282	\\ \hline
11	&	Engineering, Petroleum	&	0.0278	\\ \hline
12	&	Geography, Physical	&	0.0277	\\ \hline
13	&	Paleontology	&	0.0271	\\ \hline
14	&	Agriculture, Dairy \& Animal Science	&	0.0266	\\ \hline
15	&	Horticulture	&	0.0266	\\ \hline
16	&	Andrology	&	0.0260	\\ \hline
17	&	Fisheries	&	0.0259	\\ \hline
18	&	Education, Scientific Disciplines	&	0.0258	\\ \hline
19	&	Oceanography	&	0.0253	\\ \hline
20	&	Geochemistry \& Geophysics	&	0.0251	\\ \hline
21	&	Audiology \& Speech-Language Pathology	&	0.0242	\\ \hline
22	&	Ornithology	&	0.0239	\\ \hline
23	&	Soil Science	&	0.0225	\\ \hline
24	&	Mineralogy	&	0.0223	\\ \hline
25	&	Archaeology	&	0.0220	\\ \hline
26	&	Forestry	&	0.0206	\\ \hline
27	&	Geology	&	0.0206	\\ \hline
28	&	Poetry	&	0.0204	\\ \hline
29	&	Music	&	0.0172	\\ \hline
30	&	Dance	&	0.0136	\\ \hline
31	&	Medicine, Legal	&	0.0125	\\ \hline

\end{longtable}

\begin{longtable}[h]{|C{1cm}| L{8cm} |R{3cm}| }
	
		\caption{List of original attributes (categories) in group of positive on the $2^{th}$ principal component. Categories are sorted by values of their component coefficients in descending order, that is, from the greatest contribution to lowest contribution to the component.}\\ 
		\hline
  	  \multicolumn{3}{|C{12cm}|}{\textbf{Positive (PC2)}}  \\ \hline
\multicolumn{1}{|C{1cm}|}{\textbf{No.}} &\multicolumn{1}{|C{8cm}}{\textbf{Attribute}} &\multicolumn{1}{|C{3cm}|}{\textbf{Component Coefficient} }  \\ \hline

\endhead
\hline
\endfoot

1	&	Cultural Studies	&	0.2044	\\ \hline
2	&	Humanities, Multidisciplinary	&	0.1905	\\ \hline
3	&	Asian Studies	&	0.1887	\\ \hline
4	&	History	&	0.1877	\\ \hline
5	&	Area Studies	&	0.1831	\\ \hline
6	&	Literature	&	0.1765	\\ \hline
7	&	History Of Social Sciences	&	0.1736	\\ \hline
8	&	Sociology	&	0.1704	\\
\hline

\end{longtable}

\begin{longtable}[h]{|C{1cm}| L{8cm} |R{3cm}| }
	
		\caption{List of original attributes (categories) in group of zero on the $2^{th}$ principal component. Categories with positive component coefficients are sorted by values of their component coefficients in descending order. Categories with negative component coefficients are sorted by absolute values of their component coefficients in descending order. That is, categories in two directions are sorted from the greatest contribution to lowest contribution to the component.}\\ 
		\hline
  	  \multicolumn{3}{|C{12cm}|}{\textbf{Zero (PC2)}}  \\ \hline
\multicolumn{1}{|C{1cm}|}{\textbf{No.}} &\multicolumn{1}{|C{8cm}}{\textbf{Attribute}} &\multicolumn{1}{|C{3cm}|}{\textbf{Component Coefficient} }  \\ \hline

\endhead
\hline
\endfoot

1	&	Family Studies	&	0.0279	\\ \hline
2	&	Psychology, Educational	&	0.0255	\\ \hline
3	&	Business, Finance	&	0.0236	\\ \hline
4	&	Agricultural Economics \& Policy	&	0.0226	\\ \hline
5	&	Education, Scientific Disciplines	&	0.0175	\\ \hline
6	&	Education, Special	&	0.0087	\\ \hline
7	&	Psychology, Mathematical	&	0.0083	\\ \hline
8	&	Health Policy \& Services	&	0.0038	\\
\hline

\end{longtable}

\begin{longtable}[h]{|C{1cm}| L{8cm} |R{3cm}| }
	
		\caption{List of original attributes (categories) in group of negative on the $2^{th}$ principal component. Categories are sorted by absolute values of their component coefficients in descending order, that is, from the greatest contribution to lowest contribution to the component.}\\ 
		\hline
  	  \multicolumn{3}{|C{12cm}|}{\textbf{Negative (PC2)} } \\ \hline
\multicolumn{1}{|C{1cm}|}{\textbf{No.}} &\multicolumn{1}{|C{8cm}}{\textbf{Attribute}} &\multicolumn{1}{|C{3cm}|}{\textbf{Component Coefficient} } \\ \hline

\endhead
\hline
\endfoot

1	&	Engineering, Multidisciplinary	&	-0.0759	\\ \hline
2	&	Clinical Neurology	&	-0.0686	\\ \hline
3	&	Medicine, General \& Internal	&	-0.0677	\\ \hline
4	&	Engineering, Electrical \& Electronic	&	-0.0655	\\ \hline
5	&	Computer Science, Theory \& Methods	&	-0.0621	\\ \hline
6	&	Gastroenterology \& Hepatology	&	-0.0612	\\ \hline
7	&	Surgery	&	-0.0605	\\ \hline
8	&	Medicine, Research \& Experimental	&	-0.0579	\\
\hline

\end{longtable}

\begin{longtable}[h]{|C{1cm}| L{8cm} |R{3cm}| }
	
		\caption{List of original attributes (categories) in group of positive on the $3^{th}$ principal component. Categories are sorted by values of their component coefficients in descending order, that is, from the greatest contribution to lowest contribution to the component.}\\ 
		\hline
  	  \multicolumn{3}{|C{12cm}|}{\textbf{Positive (PC3)}}  \\ \hline
\multicolumn{1}{|C{1cm}|}{\textbf{No.}} &\multicolumn{1}{|C{8cm}}{\textbf{Attribute}} &\multicolumn{1}{|C{3cm}|}{\textbf{Component Coefficient} }  \\ \hline

\endhead
\hline
\endfoot

1	&	Multidisciplinary Sciences	&	0.1711	\\ \hline
2	&	Biochemistry \& Molecular Biology	&	0.1628	\\ \hline
3	&	Cell Biology	&	0.1570	\\ \hline
4	&	Biology	&	0.1392	\\ \hline
5	&	Computer Science, Information Systems	&	0.1324	\\ \hline
6	&	Computer Science, Theory \& Methods	&	0.1320	\\ \hline
7	&	Computer Science, Interdisciplinary Applications	&	0.1303	\\ \hline
8	&	Computer Science, Artificial Intelligence	&	0.1296	\\
\hline

\end{longtable}

\begin{longtable}[h]{|C{1cm}| L{8cm} |R{3cm}| }
	
		\caption{List of original attributes (categories) in group of zero on the $3^{th}$ principal component. Categories with positive component coefficients are sorted by values of their component coefficients in descending order. Categories with negative component coefficients are sorted by absolute values of their component coefficients in descending order. That is, categories in two directions are sorted from the greatest contribution to lowest contribution to the component.}\\ 
		\hline
  	  \multicolumn{3}{|C{12cm}|}{\textbf{Zero (PC3)}}  \\ \hline
\multicolumn{1}{|C{1cm}|}{\textbf{No.}} &\multicolumn{1}{|C{8cm}}{\textbf{Attribute}} &\multicolumn{1}{|C{3cm}|}{\textbf{Component Coefficient} }  \\ \hline

\endhead
\hline
\endfoot

1	&	Remote Sensing	&	0.0305	\\ \hline
2	&	Agronomy	&	0.0303	\\ \hline
3	&	Horticulture	&	0.0295	\\ \hline
4	&	Logic	&	0.0293	\\ \hline
5	&	Materials Science, Biomaterials	&	0.0292	\\ \hline
6	&	Transportation	&	0.0290	\\ \hline
7	&	Agriculture, Dairy \& Animal Science	&	0.0283	\\ \hline
8	&	Literature, Romance	&	0.0279	\\
\hline

\end{longtable}
\begin{longtable}[h]{|C{1cm}| L{8cm} |R{3cm}| }
	
		\caption{List of original attributes (categories) in group of negative on the $3^{th}$ principal component. Categories are sorted by absolute values of their component coefficients in descending order, that is, from the greatest contribution to lowest contribution to the component.}\\ 
		\hline
  	  \multicolumn{3}{|C{12cm}|}{\textbf{Negative (PC3)}}  \\ \hline
\multicolumn{1}{|C{1cm}|}{\textbf{No.}} &\multicolumn{1}{|C{8cm}}{\textbf{Attribute}} &\multicolumn{1}{|C{3cm}|}{\textbf{Component Coefficient} } \\ \hline

\endhead
\hline
\endfoot

1	&	Medicine, General \& Internal	&	-0.1713	\\ \hline
2	&	Health Care Sciences \& Services	&	-0.1592	\\ \hline
3	&	Primary Health Care	&	-0.1449	\\ \hline
4	&	Public, Environmental \& Occupational Health	&	-0.1448	\\ \hline
5	&	Health Policy \& Services	&	-0.1363	\\ \hline
6	&	Critical Care Medicine	&	-0.1320	\\ \hline
7	&	Clinical Neurology	&	-0.1303	\\ \hline
8	&	Rehabilitation	&	-0.1201	\\
\hline

\end{longtable}

\begin{longtable}[h]{|C{1cm}| L{8cm} |R{3cm}| }
	
		\caption{List of original attributes (categories) in group of positive on the $4^{th}$ principal component. Categories are sorted by values of their component coefficients in descending order, that is, from the greatest contribution to lowest contribution to the component.}\\ 
		\hline
  	  \multicolumn{3}{|C{12cm}|}{\textbf{Positive (PC4)}}  \\ \hline
\multicolumn{1}{|C{1cm}|}{\textbf{No.}} &\multicolumn{1}{|C{8cm}}{\textbf{Attribute}} &\multicolumn{1}{|C{3cm}|}{\textbf{Component Coefficient} } \\ \hline

\endhead
\hline
\endfoot

1	&	Psychology, Multidisciplinary	&	0.1975	\\ \hline
2	&	Psychology, Applied	&	0.1690	\\ \hline
3	&	Psychology, Social	&	0.1638	\\ \hline
4	&	Social Work	&	0.1593	\\ \hline
5	&	Psychology	&	0.1373	\\ \hline
6	&	Family Studies	&	0.1363	\\ \hline
7	&	Psychology, Clinical	&	0.1331	\\ \hline
8	&	Social Sciences, Biomedical	&	0.1265	\\
\hline

\end{longtable}
\begin{longtable}[h]{|C{1cm}| L{8cm} |R{3cm}| }
	
		\caption{List of original attributes (categories) in group of zero on the $4^{th}$ principal component. Categories with positive component coefficients are sorted by values of their component coefficients in descending order. Categories with negative component coefficients are sorted by absolute values of their component coefficients in descending order. That is, categories in two directions are sorted from the greatest contribution to lowest contribution to the component.}\\ 
		\hline
  	  \multicolumn{3}{|C{12cm}|}{\textbf{Zero (PC4)}}  \\ \hline
\multicolumn{1}{|C{1cm}|}{\textbf{No.}} &\multicolumn{1}{|C{8cm}}{\textbf{Attribute}} &\multicolumn{1}{|C{3cm}|}{\textbf{Component Coefficient} } \\ \hline

\endhead
\hline
\endfoot

1	&	Biochemical Research Methods	&	0.0313	\\ \hline
2	&	Communication	&	0.0309	\\ \hline
3	&	Ethnic Studies	&	0.0307	\\ \hline
4	&	Microscopy	&	0.0300	\\ \hline
5	&	Spectroscopy	&	0.0284	\\ \hline
6	&	Zoology	&	0.0280	\\ \hline
7	&	Polymer Science	&	0.0279	\\ \hline
8	&	Mycology	&	0.0272	\\
\hline

\end{longtable}

\begin{longtable}[h]{|C{1cm}| L{8cm} |R{3cm}| }
	
		\caption{List of original attributes (categories) in group of negative on the $4^{th}$ principal component. Categories are sorted by absolute values of their component coefficients in descending order, that is, from the greatest contribution to lowest contribution to the component.}\\ 
		\hline
  	  \multicolumn{3}{|C{12cm}|}{\textbf{Negative (PC4)}}  \\ \hline
\multicolumn{1}{|C{1cm}|}{\textbf{No.}} &\multicolumn{1}{|C{8cm}}{\textbf{Attribute}} &\multicolumn{1}{|C{3cm}|}{\textbf{Component Coefficient} } \\ \hline

\endhead
\hline
\endfoot

1	&	Literature		&	-0.1482	\\ \hline
2	&	Surgery		&	-0.1389	\\ \hline
3	&	Literature, Romance		&	-0.1364	\\ \hline
4	&	Literature, British Isles		&	-0.1318	\\ \hline
5	&	Literary Theory \& Criticism		&	-0.1295	\\ \hline
6	&	Gastroenterology \& Hepatology		&	-0.1279	\\ \hline
7	&	Medieval \& Renaissance Studies		&	-0.1242	\\ \hline
8	&	Critical Care Medicine		&	-0.1241	\\
\hline

\end{longtable}

\begin{longtable}[h]{|C{1cm}| L{8cm} |R{3cm}| }
	
		\caption{List of original attributes (categories) in group of positive on the $5^{th}$ principal component. Categories are sorted by values of their component coefficients in descending order, that is, from the greatest contribution to lowest contribution to the component.}\\ 
		\hline
  	  \multicolumn{3}{|C{12cm}|}{\textbf{Positive (PC5)}}  \\ \hline
\multicolumn{1}{|C{1cm}|}{\textbf{No.}} &\multicolumn{1}{|C{8cm}}{\textbf{Attribute}} &\multicolumn{1}{|C{3cm}|}{\textbf{Component Coefficient} }  \\ \hline

\endhead
\hline
\endfoot

1	&	Ecology	&	0.2341	\\ \hline
2	&	Biodiversity Conservation	&	0.2128	\\ \hline
3	&	Environmental Sciences	&	0.2122	\\ \hline
4	&	Marine \& Freshwater Biology	&	0.2103	\\ \hline
5	&	Geosciences, Multidisciplinary	&	0.2101	\\ \hline
6	&	Limnology	&	0.2056	\\ \hline
7	&	Geography, Physical	&	0.1976	\\ \hline
8	&	Oceanography	&	0.1783	\\
\hline

\end{longtable}

\begin{longtable}[h]{|C{1cm}| L{8cm} |R{3cm}| }
	
		\caption{List of original attributes (categories) in group of zero on the $5^{th}$ principal component. Categories with positive component coefficients are sorted by values of their component coefficients in descending order. Categories with negative component coefficients are sorted by absolute values of their component coefficients in descending order. That is, categories in two directions are sorted from the greatest contribution to lowest contribution to the component.}\\ 
		\hline
  	  \multicolumn{3}{|C{12cm}|}{\textbf{Zero (PC5)}}  \\ \hline
\multicolumn{1}{|C{1cm}|}{\textbf{No.}} &\multicolumn{1}{|C{8cm}}{\textbf{Attribute}} &\multicolumn{1}{|C{3cm}|}{\textbf{Component Coefficient} } \\ \hline

\endhead
\hline
\endfoot
1	&	Engineering, Marine	&	0.0307	\\ \hline
2	&	History Of Social Sciences	&	0.0301	\\ \hline
3	&	Genetics \& Heredity	&	0.0294	\\ \hline
4	&	Medical Laboratory Technology	&	0.0290	\\ \hline
5	&	Gastroenterology \& Hepatology	&	0.0286	\\ \hline
6	&	Industrial Relations \& Labor	&	0.0258	\\ \hline
7	&	Tropical Medicine	&	0.0257	\\ \hline
8	&	Physics, Multidisciplinary	&	0.0251	\\
\hline

\end{longtable}

\begin{longtable}[h]{|C{1cm}| L{8cm} |R{3cm}| }
	
		\caption{List of original attributes (categories) in group of negative on the $5^{th}$ principal component. Categories are sorted by absolute values of their component coefficients in descending order, that is, from the greatest contribution to lowest contribution to the component.}\\ 
		\hline
  	  \multicolumn{3}{|C{12cm}|}{\textbf{Negative (PC5)}}  \\ \hline
\multicolumn{1}{|C{1cm}|}{\textbf{No.}} &\multicolumn{1}{|C{8cm}}{\textbf{Attribute}} &\multicolumn{1}{|C{3cm}|}{\textbf{Component Coefficient} } \\ \hline

\endhead
\hline
\endfoot

1	&	Computer Science, Interdisciplinary Applications	&	-0.1071	\\ \hline
2	&	Computer Science, Artificial Intelligence	&	-0.1066	\\ \hline
3	&	Computer Science, Information Systems	&	-0.1022	\\ \hline
4	&	Computer Science, Theory \& Methods	&	-0.1016	\\ \hline
5	&	Computer Science, Cybernetics	&	-0.0991	\\ \hline
6	&	Psychology	&	-0.0978	\\ \hline
7	&	Computer Science, Software Engineering	&	-0.0934	\\ \hline
8	&	Computer Science, Hardware \& Architecture	&	-0.0922	\\
\hline

\end{longtable}

\begin{longtable}[h]{|C{1cm}| L{8cm} |R{3cm}| }
	
		\caption{List of original attributes (categories) in group of positive on the $6^{th}$ principal component. Categories are sorted by values of their component coefficients in descending order, that is, from the greatest contribution to lowest contribution to the component.}\\ 
		\hline
  	  \multicolumn{3}{|C{12cm}|}{\textbf{Positive (PC6)}}  \\ \hline
\multicolumn{1}{|C{1cm}|}{\textbf{No.}} &\multicolumn{1}{|C{8cm}}{\textbf{Attribute}} &\multicolumn{1}{|C{3cm}|}{\textbf{Component Coefficient} }  \\ \hline

\endhead
\hline
\endfoot

1	&	Psychology	&	0.1652	\\ \hline
2	&	Ecology	&	0.1549	\\ \hline
3	&	Marine \& Freshwater Biology	&	0.1460	\\ \hline
4	&	Psychology, Experimental	&	0.1365	\\ \hline
5	&	Zoology	&	0.1343	\\ \hline
6	&	Psychology, Clinical	&	0.1327	\\ \hline
7	&	Psychology, Multidisciplinary	&	0.1308	\\ \hline
8	&	Behavioral Sciences	&	0.1294	\\
\hline

\end{longtable}

\begin{longtable}[h]{|C{1cm}| L{8cm} |R{3cm}| }
	
		\caption{List of original attributes (categories) in group of negative on the $6^{th}$ principal component. Categories are sorted by absolute values of their component coefficients in descending order, that is, from the greatest contribution to lowest contribution to the component.}\\ 
		\hline
  	  \multicolumn{3}{|C{12cm}|}{\textbf{Negative (PC6)}}  \\ \hline
\multicolumn{1}{|C{1cm}|}{\textbf{No.}} &\multicolumn{1}{|C{8cm}}{\textbf{Attribute}} &\multicolumn{1}{|C{3cm}|}{\textbf{Component Coefficient} } \\ \hline

\endhead
\hline
\endfoot

1	&	Economics	&	-0.2028	\\ \hline
2	&	Planning \& Development	&	-0.1995	\\ \hline
3	&	Business, Finance	&	-0.1672	\\ \hline
4	&	Environmental Studies	&	-0.1599	\\ \hline
5	&	Public Administration	&	-0.1541	\\ \hline
6	&	Business	&	-0.1538	\\ \hline
7	&	Management	&	-0.1420	\\ \hline
8	&	Social Sciences, Mathematical Methods	&	-0.1284	\\
\hline

\end{longtable}

\begin{longtable}[h]{|C{1cm}| L{8cm} |R{3cm}| }
	
		\caption{List of original attributes (categories) in group of positive on the $7^{th}$ principal component. Categories are sorted by values of their component coefficients in descending order, that is, from the greatest contribution to lowest contribution to the component.}\\ 
		\hline
  	  \multicolumn{3}{|C{12cm}|}{\textbf{Positive (PC7)}}  \\ \hline
\multicolumn{1}{|C{1cm}|}{\textbf{No.}} &\multicolumn{1}{|C{8cm}}{\textbf{Attribute}} &\multicolumn{1}{|C{3cm}|}{\textbf{Component Coefficient} }  \\ \hline

\endhead
\hline
\endfoot

1	&	Cell Biology	&	0.2339	\\ \hline
2	&	Multidisciplinary Sciences	&	0.2313	\\ \hline
3	&	Biochemistry \& Molecular Biology	&	0.1986	\\ \hline
4	&	Developmental Biology	&	0.1826	\\ \hline
5	&	Medicine, Research \& Experimental	&	0.1784	\\ \hline
6	&	Cell \& Tissue Engineering	&	0.1582	\\ \hline
7	&	Immunology	&	0.1435	\\ \hline
8	&	Biophysics	&	0.1386	\\
\hline

\end{longtable}

\begin{longtable}[h]{|C{1cm}| L{8cm} |R{3cm}| }
	
		\caption{List of original attributes (categories) in group of negative on the $7^{th}$ principal component. Categories are sorted by absolute values of their component coefficients in descending order, that is, from the greatest contribution to lowest contribution to the component.}\\ 
		\hline
  	  \multicolumn{3}{|C{12cm}|}{\textbf{Negative (PC7)}}  \\ \hline
\multicolumn{1}{|C{1cm}|}{\textbf{No.}} &\multicolumn{1}{|C{8cm}}{\textbf{Attribute}} &\multicolumn{1}{|C{3cm}|}{\textbf{Component Coefficient} } \\ \hline

\endhead
\hline
\endfoot

1	&	Computer Science, Information Systems	&	-0.1967	\\ \hline
2	&	Computer Science, Artificial Intelligence	&	-0.1877	\\ \hline
3	&	Computer Science, Interdisciplinary Applications	&	-0.1855	\\ \hline
4	&	Computer Science, Theory \& Methods	&	-0.1831	\\ \hline
5	&	Computer Science, Hardware \& Architecture	&	-0.1729	\\ \hline
6	&	Computer Science, Software Engineering	&	-0.1705	\\ \hline
7	&	Computer Science, Cybernetics	&	-0.1646	\\ \hline
8	&	Telecommunications	&	-0.1446	\\
\hline

\end{longtable}

\begin{longtable}[h]{|C{1cm}| L{8cm} |R{3cm}| }
	
		\caption{List of original attributes (categories) in group of positive on the $8^{th}$ principal component. Categories are sorted by values of their component coefficients in descending order, that is, from the greatest contribution to lowest contribution to the component.}\\ 
		\hline
  	  \multicolumn{3}{|C{12cm}|}{\textbf{Positive (PC8)}}  \\ \hline
\multicolumn{1}{|C{1cm}|}{\textbf{No.}} &\multicolumn{1}{|C{8cm}}{\textbf{Attribute}} &\multicolumn{1}{|C{3cm}|}{\textbf{Component Coefficient} }  \\ \hline

\endhead
\hline
\endfoot

1	&	Materials Science, Coatings \& Films	&	0.2693	\\ \hline
2	&	Chemistry, Physical	&	0.2641	\\ \hline
3	&	Materials Science, Multidisciplinary	&	0.2636	\\ \hline
4	&	Nanoscience \& Nanotechnology	&	0.2627	\\ \hline
5	&	Physics, Condensed Matter	&	0.2369	\\ \hline
6	&	Materials Science, Ceramics	&	0.2214	\\ \hline
7	&	Chemistry, Multidisciplinary	&	0.1869	\\ \hline
8	&	Physics, Applied	&	0.1779	\\
\hline

\end{longtable}

\begin{longtable}[h]{|C{1cm}| L{8cm} |R{3cm}| }
	
		\caption{List of original attributes (categories) in group of negative on the $8^{th}$ principal component. Categories are sorted by absolute values of their component coefficients in descending order, that is, from the greatest contribution to lowest contribution to the component.}\\ 
		\hline
  	  \multicolumn{3}{|C{12cm}|}{\textbf{Negative (PC8)}}  \\ \hline
\multicolumn{1}{|C{1cm}|}{\textbf{No.}} &\multicolumn{1}{|C{8cm}}{\textbf{Attribute}} &\multicolumn{1}{|C{3cm}|}{\textbf{Component Coefficient} } \\ \hline

\endhead
\hline
\endfoot

1	&	Social Work	&	-0.0908	\\ \hline
2	&	Health Policy \& Services	&	-0.0897	\\ \hline
3	&	Engineering, Electrical \& Electronic	&	-0.0865	\\ \hline
4	&	Health Care Sciences \& Services	&	-0.0851	\\ \hline
5	&	Primary Health Care	&	-0.0836	\\ \hline
6	&	Engineering, Multidisciplinary	&	-0.0835	\\ \hline
7	&	Social Sciences, Biomedical	&	-0.0830	\\ \hline
8	&	Family Studies	&	-0.0761	\\
\hline

\end{longtable}

\begin{longtable}[h]{|C{1cm}| L{8cm} |R{3cm}| }
	
		\caption{List of original attributes (categories) in group of positive on the $9^{th}$ principal component. Categories are sorted by values of their component coefficients in descending order, that is, from the greatest contribution to lowest contribution to the component.}\\ 
		\hline
  	  \multicolumn{3}{|C{12cm}|}{\textbf{Positive (PC9)}}  \\ \hline
\multicolumn{1}{|C{1cm}|}{\textbf{No.}} &\multicolumn{1}{|C{8cm}}{\textbf{Attribute}} &\multicolumn{1}{|C{3cm}|}{\textbf{Component Coefficient} }  \\ \hline

\endhead
\hline
\endfoot

1	&	Geosciences, Multidisciplinary	&	0.2756	\\ \hline
2	&	Geochemistry \& Geophysics	&	0.2152	\\ \hline
3	&	Geology	&	0.2057	\\ \hline
4	&	Water Resources	&	0.1901	\\ \hline
5	&	Geography, Physical	&	0.1767	\\ \hline
6	&	Oceanography	&	0.1379	\\ \hline
7	&	Mineralogy	&	0.1360	\\ \hline
8	&	Meteorology \& Atmospheric Sciences	&	0.1345	\\
\hline

\end{longtable}

\begin{longtable}[h]{|C{1cm}| L{8cm} |R{3cm}| }
	
		\caption{List of original attributes (categories) in group of negative on the $9^{th}$ principal component. Categories are sorted by absolute values of their component coefficients in descending order, that is, from the greatest contribution to lowest contribution to the component.}\\ 
		\hline
  	  \multicolumn{3}{|C{12cm}|}{\textbf{Negative (PC9)}}  \\ \hline
\multicolumn{1}{|C{1cm}|}{\textbf{No.}} &\multicolumn{1}{|C{8cm}}{\textbf{Attribute}} &\multicolumn{1}{|C{3cm}|}{\textbf{Component Coefficient} } \\ \hline

\endhead
\hline
\endfoot

1	&	Zoology	&	-0.2252	\\ \hline
2	&	Evolutionary Biology	&	-0.2095	\\ \hline
3	&	Biology	&	-0.1865	\\ \hline
4	&	Entomology	&	-0.1826	\\ \hline
5	&	Mycology	&	-0.1805	\\ \hline
6	&	Biodiversity Conservation	&	-0.1613	\\ \hline
7	&	Parasitology	&	-0.1577	\\ \hline
8	&	Plant Sciences	&	-0.1575	\\
\hline

\end{longtable}

\begin{longtable}[h]{|C{1cm}| L{8cm} |R{3cm}| }
	
		\caption{List of original attributes (categories) in group of positive on the $10^{th}$ principal component. Categories are sorted by values of their component coefficients in descending order, that is, from the greatest contribution to lowest contribution to the component.}\\ 
		\hline
  	  \multicolumn{3}{|C{12cm}|}{\textbf{Positive (PC10)}}  \\ \hline
\multicolumn{1}{|C{1cm}|}{\textbf{No.}} &\multicolumn{1}{|C{8cm}}{\textbf{Attribute}} &\multicolumn{1}{|C{3cm}|}{\textbf{Component Coefficient} }  \\ \hline

\endhead
\hline
\endfoot

1	&	Agriculture, Multidisciplinary	&	0.3288	\\ \hline
2	&	Agronomy	&	0.2928	\\ \hline
3	&	Agricultural Engineering	&	0.2720	\\ \hline
4	&	Horticulture	&	0.2497	\\ \hline
5	&	Food Science \& Technology	&	0.2079	\\ \hline
6	&	Plant Sciences	&	0.1762	\\ \hline
7	&	Soil Science	&	0.1735	\\ \hline
8	&	Engineering, Environmental	&	0.1523	\\
\hline

\end{longtable}

\begin{longtable}[h]{|C{1cm}| L{8cm} |R{3cm}| }
	
		\caption{List of original attributes (categories) in group of negative on the $10^{th}$ principal component. Categories are sorted by absolute values of their component coefficients in descending order, that is, from the greatest contribution to lowest contribution to the component.}\\ 
		\hline
  	  \multicolumn{3}{|C{12cm}|}{\textbf{Negative (PC10)}}  \\ \hline
\multicolumn{1}{|C{1cm}|}{\textbf{No.}} &\multicolumn{1}{|C{8cm}}{\textbf{Attribute}} &\multicolumn{1}{|C{3cm}|}{\textbf{Component Coefficient} } \\ \hline

\endhead
\hline
\endfoot

1	&	Geology	&	-0.1763	\\ \hline
2	&	Paleontology	&	-0.1562	\\ \hline
3	&	Geosciences, Multidisciplinary	&	-0.1480	\\ \hline
4	&	Geochemistry \& Geophysics	&	-0.1418	\\ \hline
5	&	Geography, Physical	&	-0.1236	\\ \hline
6	&	Oceanography	&	-0.1195	\\ \hline
7	&	Marine \& Freshwater Biology	&	-0.1040	\\ \hline
8	&	Zoology	&	-0.1021	\\
\hline

\end{longtable}

\begin{longtable}[h]{|C{1cm}| L{8cm} |R{3cm}| }
	
		\caption{List of original attributes (categories) in group of positive on the $11^{th}$ principal component. Categories are sorted by values of their component coefficients in descending order, that is, from the greatest contribution to lowest contribution to the component.}\\ 
		\hline
  	  \multicolumn{3}{|C{12cm}|}{\textbf{Positive (PC11)}}  \\ \hline
\multicolumn{1}{|C{1cm}|}{\textbf{No.}} &\multicolumn{1}{|C{8cm}}{\textbf{Attribute}} &\multicolumn{1}{|C{3cm}|}{\textbf{Component Coefficient} }  \\ \hline

\endhead
\hline
\endfoot

1	&	Health Policy \& Services	&	0.2125	\\ \hline
2	&	Health Care Sciences \& Services	&	0.1977	\\ \hline
3	&	Social Sciences, Biomedical	&	0.1841	\\ \hline
4	&	Primary Health Care	&	0.1571	\\ \hline
5	&	Medical Informatics	&	0.1571	\\ \hline
6	&	Nursing	&	0.1566	\\ \hline
7	&	Public, Environmental \& Occupational Health	&	0.1558	\\ \hline
8	&	Biotechnology \& Applied Microbiology	&	0.1341	\\
\hline

\end{longtable}

\begin{longtable}[h]{|C{1cm}| L{8cm} |R{3cm}| }
	
		\caption{List of original attributes (categories) in group of negative on the $11^{th}$ principal component. Categories are sorted by absolute values of their component coefficients in descending order, that is, from the greatest contribution to lowest contribution to the component.}\\ 
		\hline
  	  \multicolumn{3}{|C{12cm}|}{\textbf{Negative (PC11)}}  \\ \hline
\multicolumn{1}{|C{1cm}|}{\textbf{No.}} &\multicolumn{1}{|C{8cm}}{\textbf{Attribute}} &\multicolumn{1}{|C{3cm}|}{\textbf{Component Coefficient} } \\ \hline

\endhead
\hline
\endfoot

1	&	Behavioral Sciences	&	-0.2509	\\ \hline
2	&	Psychology, Biological	&	-0.2311	\\ \hline
3	&	Neurosciences	&	-0.2232	\\ \hline
4	&	Neuroimaging	&	-0.2048	\\ \hline
5	&	Psychology	&	-0.1966	\\ \hline
6	&	Psychology, Experimental	&	-0.1948	\\ \hline
7	&	Clinical Neurology	&	-0.1311	\\ \hline
8	&	Economics	&	-0.1061	\\
\hline

\end{longtable}

\begin{longtable}[h]{|C{1cm}| L{8cm} |R{3cm}| }
	
		\caption{List of original attributes (categories) in group of positive on the $12^{th}$ principal component. Categories are sorted by values of their component coefficients in descending order, that is, from the greatest contribution to lowest contribution to the component.}\\ 
		\hline
  	  \multicolumn{3}{|C{12cm}|}{\textbf{Positive (PC12)}}  \\ \hline
\multicolumn{1}{|C{1cm}|}{\textbf{No.}} &\multicolumn{1}{|C{8cm}}{\textbf{Attribute}} &\multicolumn{1}{|C{3cm}|}{\textbf{Component Coefficient} }  \\ \hline

\endhead
\hline
\endfoot

1	&	Education \& Educational Research	&	0.2070	\\ \hline
2	&	Education, Scientific Disciplines	&	0.1871	\\ \hline
3	&	Psychology, Educational	&	0.1697	\\ \hline
4	&	Language \& Linguistics	&	0.1529	\\ \hline
5	&	Linguistics	&	0.1479	\\ \hline
6	&	Management	&	0.1459	\\ \hline
7	&	Business	&	0.1344	\\ \hline
8	&	Information Science \& Library Science	&	0.1208	\\
\hline

\end{longtable}

\begin{longtable}[h]{|C{1cm}| L{8cm} |R{3cm}| }
	
		\caption{List of original attributes (categories) in group of negative on the $12^{th}$ principal component. Categories are sorted by absolute values of their component coefficients in descending order, that is, from the greatest contribution to lowest contribution to the component.}\\ 
		\hline
  	  \multicolumn{3}{|C{12cm}|}{\textbf{Negative (PC12)}}  \\ \hline
\multicolumn{1}{|C{1cm}|}{\textbf{No.}} &\multicolumn{1}{|C{8cm}}{\textbf{Attribute}} &\multicolumn{1}{|C{3cm}|}{\textbf{Component Coefficient} } \\ \hline

\endhead
\hline
\endfoot

1	&	Mechanics	&	-0.2128	\\ \hline
2	&	Engineering, Mechanical	&	-0.1907	\\ \hline
3	&	Engineering, Civil	&	-0.1439	\\ \hline
4	&	Materials Science, Characterization \& Testing	&	-0.1426	\\ \hline
5	&	Physics, Fluids \& Plasmas	&	-0.1424	\\ \hline
6	&	Engineering, Ocean	&	-0.1421	\\ \hline
7	&	Engineering, Geological	&	-0.1305	\\ \hline
8	&	Construction \& Building Technology	&	-0.1292	\\
\hline

\end{longtable}

\begin{longtable}[h]{|C{1cm}| L{8cm} |R{3cm}| }
	
		\caption{List of original attributes (categories) in group of positive on the $13^{th}$ principal component. Categories are sorted by values of their component coefficients in descending order, that is, from the greatest contribution to lowest contribution to the component.}\\ 
		\hline
  	  \multicolumn{3}{|C{12cm}|}{\textbf{Positive (PC13)}}  \\ \hline
\multicolumn{1}{|C{1cm}|}{\textbf{No.}} &\multicolumn{1}{|C{8cm}}{\textbf{Attribute}} &\multicolumn{1}{|C{3cm}|}{\textbf{Component Coefficient} }  \\ \hline

\endhead
\hline
\endfoot
1	&	Parasitology	&	0.2090	\\ \hline
2	&	Infectious Diseases	&	0.2047	\\ \hline
3	&	Education \& Educational Research	&	0.1851	\\ \hline
4	&	Tropical Medicine	&	0.1850	\\ \hline
5	&	Psychology, Educational	&	0.1814	\\ \hline
6	&	Education, Scientific Disciplines	&	0.1586	\\ \hline
7	&	Microbiology	&	0.1520	\\ \hline
8	&	Virology	&	0.1503	\\
\hline

\end{longtable}

\begin{longtable}[h]{|C{1cm}| L{8cm} |R{3cm}| }
	
		\caption{List of original attributes (categories) in group of negative on the $13^{th}$ principal component. Categories are sorted by absolute values of their component coefficients in descending order, that is, from the greatest contribution to lowest contribution to the component.}\\ 
		\hline
  	  \multicolumn{3}{|C{12cm}|}{\textbf{Negative (PC13)}}  \\ \hline
\multicolumn{1}{|C{1cm}|}{\textbf{No.}} &\multicolumn{1}{|C{8cm}}{\textbf{Attribute}} &\multicolumn{1}{|C{3cm}|}{\textbf{Component Coefficient} } \\ \hline

\endhead
\hline
\endfoot

1	&	Physiology	&	-0.1588	\\ \hline
2	&	Cell Biology	&	-0.1469	\\ \hline
3	&	Developmental Biology	&	-0.1407	\\ \hline
4	&	Biochemistry \& Molecular Biology	&	-0.1145	\\ \hline
5	&	Neurosciences	&	-0.1113	\\ \hline
6	&	Cell \& Tissue Engineering	&	-0.1080	\\ \hline
7	&	Health Policy \& Services	&	-0.1061	\\ \hline
8	&	Biophysics	&	-0.1059	\\
\hline

\end{longtable}

\section{List of categories on the left wing, right wing and the body of the bird shape, obtained in the first three PCs, with projection onto the first principal component.} \label{left}
\raggedbottom	

\begin{longtable}[h]{|C{1cm}| L{8cm} |R{3cm}| }
	
		\caption{List of categories on the left wing (green) with projection onto the first principal component. Categories are sorted by absolute values of their component score in descending order.}\\ 
		\hline
  	  \multicolumn{3}{|C{12cm}|}{\textbf{Left (Green) Wing}}  \\ \hline
\multicolumn{1}{|C{1cm}|}{\textbf{No.}} &\multicolumn{1}{|C{8cm}}{\textbf{Category}} &\multicolumn{1}{|C{3cm}|}{\textbf{PC1}}  \\ \hline

\endhead
\hline
\endfoot
1	&	Poetry	&	-0.3080	\\ \hline
2	&	Literature, African, Australian, Canadian	&	-0.2620	\\ \hline
3	&	Literature, American	&	-0.2530	\\ \hline
4	&	Literature	&	-0.2300	\\ \hline
5	&	Literature, German, Dutch, Scandinavian	&	-0.1950	\\ \hline
6	&	Literary Reviews	&	-0.1910	\\ \hline
7	&	Theater	&	-0.1830	\\ \hline
8	&	History	&	-0.1770	\\
\hline

\end{longtable}

\begin{longtable}[h]{|C{1cm}| L{8cm} |R{3cm}| }
	
		\caption{List of categories on the body (blue) with projection onto the first principal component. Categories are sorted by absolute values of their component score in descending order.}\\ 
		\hline
  	  \multicolumn{3}{|C{12cm}|}{\textbf{Body}}  \\ \hline
\multicolumn{1}{|C{1cm}|}{\textbf{No.}} &\multicolumn{1}{|C{8cm}}{\textbf{Category}} &\multicolumn{1}{|C{3cm}|}{\textbf{PC1}}  \\ \hline

\endhead
\hline
\endfoot

1	&	Agricultural Economics \& Policy	&	-0.0020	\\ \hline
2	&	Mathematical \& Computational Biology	&	0.0000	\\ \hline
3	&	Biochemical Research Methods	&	0.0000	\\ \hline
4	&	Biotechnology \& Applied Microbiology	&	0.0000	\\ \hline
5	&	Family Studies	&	0.0000	\\ \hline
6	&	Biophysics	&	0.0010	\\ \hline
7	&	Psychology, Mathematical	&	0.0010	\\ \hline
8	&	Business, Finance	&	0.0020	\\
\hline

\end{longtable}

\begin{longtable}[h]{|C{1cm}| L{8cm} |R{3cm}| }
	
		\caption{List of categories on the right wing (red) with projection onto the first principal component. Categories are sorted by absolute values of their component score in descending order.}\\ 
		\hline
  	  \multicolumn{3}{|C{12cm}|}{\textbf{Right (red) Wing }}  \\ \hline
\multicolumn{1}{|C{1cm}|}{\textbf{No.}} &\multicolumn{1}{|C{8cm}}{\textbf{Category}} &\multicolumn{1}{|C{3cm}|}{\textbf{PC1}}  \\ \hline

\endhead
\hline
\endfoot
1	&	Reproductive Biology	&	0.0380	\\ \hline
2	&	Materials Science, Multidisciplinary	&	0.0400	\\ \hline
3	&	Neuroimaging	&	0.0410	\\ \hline
4	&	Telecommunications	&	0.0420	\\ \hline
5	&	Nutrition \& Dietetics	&	0.0430	\\ \hline
6	&	Integrative \& Complementary Medicine	&	0.0430	\\ \hline
7	&	Medicine, Research \& Experimental	&	0.0450	\\ \hline
8	&	Radiology, Nuclear Medicine \& Medical Imaging	&	0.0490	\\
\hline

\end{longtable}

\section{List of original attributes (categories) in positive, negative and zero groups identified by vector approximation on the principal components.} \label{appndx2}
\raggedbottom	

\begin{longtable}[h]{|C{1cm}| L{8cm} |R{3cm}| }
	
		\caption{List of original attributes (categories) in positive group identified by vector approximation on the $2^{nd}$ principal component. Categories are sorted by values of their component coefficients in descending order, that is, from the greatest contribution to lowest contribution to the component.}\\ 
		\hline
  	  \multicolumn{3}{|C{12cm}|}{\textbf{Positive (PC2)}}  \\ \hline
\multicolumn{1}{|C{1cm}|}{\textbf{No.}} &\multicolumn{1}{|C{8cm}}{\textbf{Attribute}} &\multicolumn{1}{|C{3cm}|}{\textbf{Component Coefficient}}  \\ \hline

\endhead
\hline
\endfoot

1	&	Cultural Studies	&	0.2044	\\ \hline
2	&	Humanities, Multidisciplinary	&	0.1905	\\ \hline
3	&	Asian Studies	&	0.1887	\\ \hline
4	&	History	&	0.1877	\\ \hline
5	&	Area Studies	&	0.1831	\\ \hline
6	&	Literature	&	0.1765	\\ \hline
7	&	History Of Social Sciences	&	0.1736	\\ \hline
8	&	Sociology	&	0.1704	\\
\hline

\end{longtable}

\begin{longtable}[h]{|C{1cm}| L{8cm} |R{3cm}| }

		\caption{List of original attributes (categories) in zero group identified by vector approximation on the $2^{nd}$ principal component. Categories with positive component coefficients are sorted by values of their component coefficients in descending order. Categories with negative component coefficients are sorted by absolute values of their component coefficients in descending order. That is, categories in two directions are sorted from the greatest contribution to lowest contribution to the component.}\\ 
		\hline
  	  \multicolumn{3}{|C{12cm}|}{\textbf{Zero (PC2)}}  \\ \hline
\multicolumn{1}{|C{1cm}|}{\textbf{No.}} &\multicolumn{1}{|C{8cm}}{\textbf{Attribute}} &\multicolumn{1}{|C{3cm}|}{\textbf{Component Coefficient}}  \\ \hline

\endhead
\hline
\endfoot

1	&	Language \& Linguistics	&	0.0603	\\ \hline
2	&	Business	&	0.0544	\\ \hline
3	&	Information Science \& Library Science	&	0.0521	\\ \hline
4	&	Management	&	0.0475	\\ \hline
5	&	Linguistics	&	0.0462	\\ \hline
6	&	Music	&	0.0457	\\ \hline
7	&	Economics	&	0.0433	\\ \hline
8	&	Psychology, Applied	&	0.0430	\\
\hline

\end{longtable}

\begin{longtable}[h]{|C{1cm}| L{8cm} |R{3cm}| }
	
		\caption{List of original attributes (categories) in negative group identified by vector approximation o the $2^{nd}$ principal component. Categories are sorted by values of their component coefficients in descending order, that is, from the greatest contribution to lowest contribution to the component.}\\ 
		\hline
  	  \multicolumn{3}{|C{12cm}|}{\textbf{Negative (PC2)}}  \\ \hline
\multicolumn{1}{|C{1cm}|}{\textbf{No.}} &\multicolumn{1}{|C{8cm}}{\textbf{Attribute}} &\multicolumn{1}{|C{3cm}|}{\textbf{Component Coefficient}}  \\ \hline

\endhead
\hline
\endfoot

1	&	Engineering, Multidisciplinary	&	-0.0759	\\ \hline
2	&	Clinical Neurology	&	-0.0686	\\ \hline
3	&	Medicine, General \& Internal	&	-0.0677	\\ \hline
4	&	Engineering, Electrical \& Electronic	&	-0.0655	\\ \hline
5	&	Computer Science, Theory \& Methods	&	-0.0621	\\ \hline
6	&	Gastroenterology \& Hepatology	&	-0.0612	\\ \hline
7	&	Surgery	&	-0.0605	\\ \hline
8	&	Medicine, Research \& Experimental	&	-0.0579	\\
\hline

\end{longtable}

\begin{longtable}[h]{|C{1cm}| L{8cm} |R{3cm}| }
	
		\caption{List of original attributes (categories) in positive group identified by vector approximation o the $3^{rd}$ principal component. Categories are sorted by values of their component coefficients in descending order, that is, from the greatest contribution to lowest contribution to the component.}\\ 
		\hline
  	  \multicolumn{3}{|C{12cm}|}{\textbf{Positive (PC3)}}  \\ \hline
\multicolumn{1}{|C{1cm}|}{\textbf{No.}} &\multicolumn{1}{|C{8cm}}{\textbf{Attribute}} &\multicolumn{1}{|C{3cm}|}{\textbf{Component Coefficient}}  \\ \hline

\endhead
\hline
\endfoot

1	&	Multidisciplinary Sciences	&	0.1711	\\ \hline
2	&	Biochemistry \& Molecular Biology	&	0.1628	\\ \hline
3	&	Cell Biology	&	0.1570	\\ \hline
4	&	Biology	&	0.1392	\\ \hline
5	&	Computer Science, Information Systems	&	0.1324	\\ \hline
6	&	Computer Science, Theory \& Methods	&	0.1320	\\ \hline
7	&	Computer Science, Interdisciplinary Applications	&	0.1303	\\ \hline
8	&	Computer Science, Artificial Intelligence	&	0.1296	\\
\hline

\end{longtable}

\begin{longtable}[h]{|C{1cm}| L{8cm} |R{3cm}| }
	
		\caption{List of original attributes (categories) in zero group identified by vector approximation o the $3^{rd}$ principal component. Categories with positive component coefficients are sorted by values of their component coefficients in descending order. Categories with negative component coefficients are sorted by absolute values of their component coefficients in descending order. That is, categories in two directions are sorted from the greatest contribution to lowest contribution to the component.}\\ 
		\hline
  	  \multicolumn{3}{|C{12cm}|}{\textbf{Zero (PC3)}}  \\ \hline
\multicolumn{1}{|C{1cm}|}{\textbf{No.}} &\multicolumn{1}{|C{8cm}}{\textbf{Attribute}} &\multicolumn{1}{|C{3cm}|}{\textbf{Component Coefficient}}  \\ \hline

\endhead
\hline
\endfoot

1	&	Biodiversity Conservation	&	0.0485	\\ \hline
2	&	Engineering, Manufacturing	&	0.0469	\\ \hline
3	&	Social Sciences, Mathematical Methods	&	0.0467	\\ \hline
4	&	Parasitology	&	0.0435	\\ \hline
5	&	Anatomy \& Morphology	&	0.0430	\\ \hline
6	&	Veterinary Sciences	&	0.0428	\\ \hline
7	&	Marine \& Freshwater Biology	&	0.0418	\\ \hline
8	&	Engineering, Aerospace	&	0.0412	\\
\hline

\end{longtable}

\begin{longtable}[h]{|C{1cm}| L{8cm} |R{3cm}| }
	
		\caption{List of original attributes (categories) in negative group identified by vector approximation o the $3^{rd}$ principal component. Categories are sorted by values of their component coefficients in descending order, that is, from the greatest contribution to lowest contribution to the component.}\\ 
		\hline
  	  \multicolumn{3}{|C{12cm}|}{\textbf{Negative (PC3)}}  \\ \hline
\multicolumn{1}{|C{1cm}|}{\textbf{No.}} &\multicolumn{1}{|C{8cm}}{\textbf{Attribute}} &\multicolumn{1}{|C{3cm}|}{\textbf{Component Coefficient}}  \\ \hline

\endhead
\hline
\endfoot

1	&	Medicine, General \& Internal	&	-0.1713	\\ \hline
2	&	Health Care Sciences \& Services	&	-0.1592	\\ \hline
3	&	Primary Health Care	&	-0.1449	\\ \hline
4	&	Public, Environmental \& Occupational Health	&	-0.1448	\\ \hline
5	&	Health Policy \& Services	&	-0.1363	\\ \hline
6	&	Critical Care Medicine	&	-0.1320	\\ \hline
7	&	Clinical Neurology	&	-0.1303	\\ \hline
8	&	Rehabilitation	&	-0.1201	\\
\hline

\end{longtable}

\begin{longtable}[h]{|C{1cm}| L{8cm} |R{3cm}| }
	
		\caption{List of original attributes (categories) in positive group identified by vector approximation o the $4^{th}$ principal component. Categories are sorted by values of their component coefficients in descending order, that is, from the greatest contribution to lowest contribution to the component.}\\ 
		\hline
  	  \multicolumn{3}{|C{12cm}|}{\textbf{Positive (PC4)}}  \\ \hline
\multicolumn{1}{|C{1cm}|}{\textbf{No.}} &\multicolumn{1}{|C{8cm}}{\textbf{Attribute}} &\multicolumn{1}{|C{3cm}|}{\textbf{Component Coefficient}}  \\ \hline

\endhead
\hline
\endfoot

1	&	Psychology, Multidisciplinary	&	0.1975	\\ \hline
2	&	Psychology, Applied	&	0.1690	\\ \hline
3	&	Psychology, Social	&	0.1638	\\ \hline
4	&	Social Work	&	0.1593	\\ \hline
5	&	Psychology	&	0.1373	\\ \hline
6	&	Family Studies	&	0.1363	\\ \hline
7	&	Psychology, Clinical	&	0.1331	\\ \hline
8	&	Social Sciences, Biomedical	&	0.1265	\\
\hline

\end{longtable}

\begin{longtable}[h]{|C{1cm}| L{8cm} |R{3cm}| }
	
		\caption{List of original attributes (categories) in zero group identified by vector approximation o the $4^{th}$ principal component. Categories with positive component coefficients are sorted by values of their component coefficients in descending order. Categories with negative component coefficients are sorted by absolute values of their component coefficients in descending order. That is, categories in two directions are sorted from the greatest contribution to lowest contribution to the component.}\\ 
		\hline
  	  \multicolumn{3}{|C{12cm}|}{\textbf{Zero (PC4)}}  \\ \hline
\multicolumn{1}{|C{1cm}|}{\textbf{No.}} &\multicolumn{1}{|C{8cm}}{\textbf{Attribute}} &\multicolumn{1}{|C{3cm}|}{\textbf{Component Coefficient}}  \\ \hline

\endhead
\hline
\endfoot

1	&	Nursing	&	0.04648	\\ \hline
2	&	Biotechnology \& Applied Microbiology	&	0.04589	\\ \hline
3	&	Multidisciplinary Sciences	&	0.04389	\\ \hline
4	&	Substance Abuse	&	0.04214	\\ \hline
5	&	Materials Science, Coatings \& Films	&	0.04184	\\ \hline
6	&	Medical Ethics	&	0.04176	\\ \hline
7	&	Genetics \& Heredity	&	0.04107	\\ \hline
8	&	Nanoscience \& Nanotechnology	&	0.04048	\\
\hline

\end{longtable}

\begin{longtable}[h]{|C{1cm}| L{8cm} |R{3cm}| }
	
		\caption{List of original attributes (categories) in negative group identified by vector approximation o the $4^{th}$ principal component. Categories are sorted by values of their component coefficients in descending order, that is, from the greatest contribution to lowest contribution to the component.}\\ 
		\hline
  	  \multicolumn{3}{|C{12cm}|}{\textbf{Negative (PC4)}}  \\ \hline
\multicolumn{1}{|C{1cm}|}{\textbf{No.}} &\multicolumn{1}{|C{8cm}}{\textbf{Attribute}} &\multicolumn{1}{|C{3cm}|}{\textbf{Component Coefficient}}  \\ \hline

\endhead
\hline
\endfoot

1	&	Literature	&	-0.1482	\\ \hline
2	&	Surgery	&	-0.1389	\\ \hline
3	&	Literature, Romance	&	-0.1364	\\ \hline
4	&	Literature, British Isles	&	-0.1318	\\ \hline
5	&	Literary Theory \& Criticism	&	-0.1295	\\ \hline
6	&	Gastroenterology \& Hepatology	&	-0.1279	\\ \hline
7	&	Medieval \& Renaissance Studies	&	-0.1242	\\ \hline
8	&	Critical Care Medicine	&	-0.1241	\\
\hline

\end{longtable}

\begin{longtable}[h]{|C{1cm}| L{8cm} |R{3cm}| }
	
		\caption{List of original attributes (categories) in positive group identified by vector approximation o the $5^{th}$ principal component. Categories are sorted by values of their component coefficients in descending order, that is, from the greatest contribution to lowest contribution to the component.}\\ 
		\hline
  	  \multicolumn{3}{|C{12cm}|}{\textbf{Positive (PC5)}}  \\ \hline
\multicolumn{1}{|C{1cm}|}{\textbf{No.}} &\multicolumn{1}{|C{8cm}}{\textbf{Attribute}} &\multicolumn{1}{|C{3cm}|}{\textbf{Component Coefficient}}  \\ \hline

\endhead
\hline
\endfoot

1	&	Ecology	&	0.2341	\\ \hline
2	&	Biodiversity Conservation	&	0.2128	\\ \hline
3	&	Environmental Sciences	&	0.2122	\\ \hline
4	&	Marine \& Freshwater Biology	&	0.2103	\\ \hline
5	&	Geosciences, Multidisciplinary	&	0.2101	\\ \hline
6	&	Limnology	&	0.2056	\\ \hline
7	&	Geography, Physical	&	0.1976	\\ \hline
8	&	Oceanography	&	0.1783	\\
\hline

\end{longtable}

\begin{longtable}[h]{|C{1cm}| L{8cm} |R{3cm}| }
	
		\caption{List of original attributes (categories) in zero group identified by vector approximation o the $5^{th}$ principal component. Categories with positive component coefficients are sorted by values of their component coefficients in descending order. Categories with negative component coefficients are sorted by absolute values of their component coefficients in descending order. That is, categories in two directions are sorted from the greatest contribution to lowest contribution to the component.}\\ 
		\hline
  	  \multicolumn{3}{|C{12cm}|}{\textbf{Zero (PC5)}}  \\ \hline
\multicolumn{1}{|C{1cm}|}{\textbf{No.}} &\multicolumn{1}{|C{8cm}}{\textbf{Attribute}} &\multicolumn{1}{|C{3cm}|}{\textbf{Component Coefficient}}  \\ \hline

\endhead
\hline
\endfoot

1	&	International Relations	&	0.0635	        \\ \hline
2	&	Mineralogy	&	0.0610	        \\ \hline
3	&	Business, Finance	&	0.0595	        \\ \hline
4	&	Political Science	&	0.0518	        \\ \hline
5	&	Engineering, Ocean	&	0.0501	        \\ \hline
6	&	Area Studies	&	0.0501	        \\ \hline
7	&	Mining \& Mineral Processing	&	0.0475	        \\ \hline
8	&	Business	&	0.0442	        \\
\hline

\end{longtable}

\begin{longtable}[h]{|C{1cm}| L{8cm} |R{3cm}| }
	
		\caption{List of original attributes (categories) in negative group identified by vector approximation o the $5^{th}$ principal component. Categories are sorted by values of their component coefficients in descending order, that is, from the greatest contribution to lowest contribution to the component.}\\ 
		\hline
  	  \multicolumn{3}{|C{12cm}|}{\textbf{Negative (PC5)}}  \\ \hline
\multicolumn{1}{|C{1cm}|}{\textbf{No.}} &\multicolumn{1}{|C{8cm}}{\textbf{Attribute}} &\multicolumn{1}{|C{3cm}|}{\textbf{Component Coefficient}}  \\ \hline

\endhead
\hline
\endfoot

1	&	Computer Science, Interdisciplinary Applications	&	-0.1071	\\ \hline
2	&	Computer Science, Artificial Intelligence	&	-0.1066	\\ \hline
3	&	Computer Science, Information Systems	&	-0.1022	\\ \hline
4	&	Computer Science, Theory \& Methods	&	-0.1016	\\ \hline
5	&	Computer Science, Cybernetics	&	-0.0991	\\ \hline
6	&	Psychology	&	-0.0978	\\ \hline
7	&	Computer Science, Software Engineering	&	-0.0934	\\ \hline
8	&	Computer Science, Hardware \& Architecture	&	-0.0922	\\
\hline

\end{longtable}

\end{document}